\documentclass[]{beingbeyond}
\usepackage{enumitem}
\usepackage[toc,page,header]{appendix}

\usepackage[utf8]{inputenc} 
\usepackage[T1]{fontenc}    
\usepackage{hyperref}       
\usepackage{url}            
\usepackage{array}          
\usepackage{booktabs}       
\usepackage{amsfonts}       
\usepackage{nicefrac}       
\usepackage{microtype}      
\usepackage{xcolor}         
\usepackage{xspace}
\usepackage{bm}
\usepackage{bbm}
\usepackage{bbding}
\usepackage{tabularx}
\usepackage{textcomp}
\usepackage{amssymb}
\usepackage{enumitem}
\usepackage{amsmath}
\usepackage{mathtools}
\usepackage{amsthm}
\usepackage{multirow}
\usepackage{makecell}
\usepackage{color}
\usepackage{colortbl}
\usepackage{adjustbox}
\usepackage{caption}
\usepackage{graphicx}
\usepackage{wrapfig}
\usepackage{array}
\usepackage{multicol}

\definecolor{myyellow}{RGB}{255,192,0}
\definecolor{mygreen}{RGB}{107,170,64}

\definecolor{mywrite}{RGB}{255,227,132}

\title{Being-H0: Vision-Language-Action Pretraining \\from Large-Scale  Human Videos}

\author{{\bfseries 
Hao Luo$^{1,3,*}$ \quad 
Yicheng Feng$^{1,3,*}$ \quad
Wanpeng Zhang$^{1,3,*}$ \quad
Sipeng Zheng$^{3,*}$  \\
Ye Wang$^{2,3}$ \quad
Haoqi Yuan$^{1,3}$ \quad
Jiazheng Liu$^{1}$ \quad
Chaoyi Xu$^{3}$ \quad
Qin Jin$^{2}$ \\
Zongqing Lu$^{1,3,\dagger}$
}}

\affiliation{{$^{1}$Peking University \quad $^{2}$Renmin University of China \quad $^{3}$BeingBeyond}}

\webpage{\url{https://beingbeyond.github.io/Being-H0}}

\firstfig[width=\linewidth][.98\textwidth]{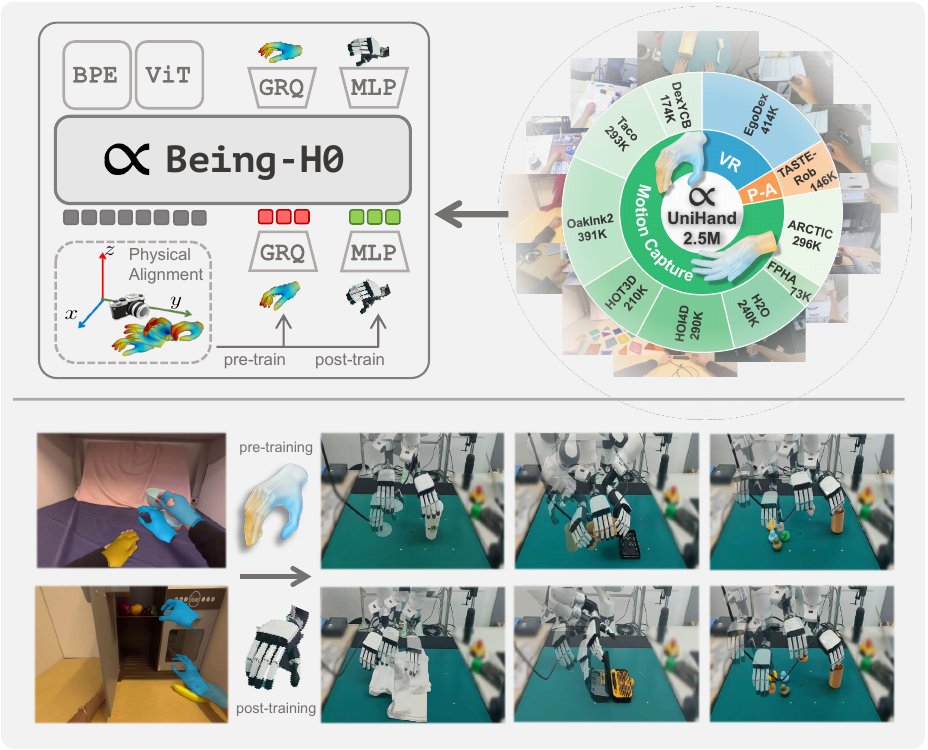}
{\textbf{\texttt{\ModelName}} acquires dexterous manipulation skills by learning from large-scale human videos in the UniHand dataset via \texttt{physical instruction tuning}. 
By explicitly modeling hand motion patterns, the resulting foundation model seamlessly transfers from human hand demonstrations to robotic manipulation.}
{fig:001}

\abstract{
We introduce \textbf{\texttt{\ModelName}}, a dexterous Vision-Language-Action model (VLA) trained on large-scale human videos.
Existing VLAs struggle with complex manipulation tasks requiring high dexterity and generalize poorly to novel scenarios and tasks, primarily due to their reliance on synthetic data with significant sim-to-real gaps or teleoperated demonstrations lacking scale and diversity.
To address this data bottleneck, we propose leveraging human hands as a ``foundation manipulator'', capitalizing on the rich dexterity and scalability present in web data.
Our approach centers on \texttt{physical instruction tuning}, a novel training paradigm that combines large-scale VLA pretraining from human videos, physical space alignment for 3D reasoning, and post-training adaptation for robotic tasks.
Additionally, we introduce a part-level motion tokenization method which achieves millimeter-level reconstruction accuracy to model precise hand trajectories for action learning. 
To support our proposed paradigm, we further develop a comprehensive data curation pipeline that integrates heterogeneous sources --- including motion capture, VR, and RGB-only videos --- into a large-scale dataset with millions of motion-based instructional instances.
We empirically show the excellence of \textbf{\texttt{\ModelName}} in hand motion generation and instruction following, and it also scales well with model and data sizes.
Importantly, we observe the expected gains of \textbf{\texttt{\ModelName}} in real-world robotic manipulation as physical instruction tuning is applied. 
}

\checkdata[Date]{July 21, 2025}

\definecolor{BlockC}{gray}{0.98}  
\definecolor{BlockA}{RGB}{191,211,230}
\definecolor{BlockB}{RGB}{199,233,192}

\newcommand{\DataName}{\texttt{UniHand}\xspace}
\newcommand{\ModelName}{\texttt{Being-H0}\xspace}

\begingroup
\footnotetext[1]{These authors contributed equally to this work.}
\footnotetext[2]{Correspondence to Zongqing Lu $<$lu@beingbeyond.com$>$.}
\endgroup

\begin{document}

\maketitle

\section{Introduction}
\label{sec:introduction}

The rise of ChatGPT and its successors has revolutionized AI research, endowing large multimodal models (LMMs) with versatile reasoning capabilities that excel in domains ranging from agentic systems to mathematics and coding. 
Yet robotics still lacks its ``ChatGPT moment'' --- a transformative leap for embodied intelligence~\cite{sampath2023review,huang2025human}. 
Recent efforts have tried to bridge this gap by adapting LMMs into Vision-Language-Action models (VLAs)~\cite{brohan2022rt, brohan2023rt, kim2024openvla, black2410pi0, ma2024survey}, leveraging their multimodal perception and human-like reasoning for robotic tasks.

However, these approaches remain constrained by their reliance on teleoperated demonstrations, limiting generalization and stability, particularly in novel environments.
The core bottleneck stems from teleoperated datasets~\cite{o2024open,khazatsky2024droid,bu2025agibot} being orders of magnitude smaller than internet-scale LMM training data, trapping embodied intelligence~\cite{sampath2023review,huang2025human} in a persistent ``data swamp''.
Models~\cite{team2024octo,kim2024openvla} trained on such lab-collected data consequently struggle with robust manipulation across diverse objects and environments.
This scarcity is particularly severe for dexterous hands due to operational complexity and hardware costs, restricting most VLAs to simple grippers that lack true dexterity~\cite{cutler2024benchmarking, an2025dexterous,guruprasad2024benchmarking, de2025scaffolding}.
With limited degrees of freedom and no fine finger control, these end-effectors cannot achieve the precise coordination or delicate force modulation needed for complex interactions.
While some attempts~\cite{xu2023unidexgrasp,wan2023unidexgrasp++,zhong2025dexgraspvla,he2025dexvlg,huang2025efficient} have leveraged simulators to obtain low-cost synthetic data, their limited diversity and unresolved sim-to-real gap continue to prevent real-world deployment of dexterous hands.

Human videos offer another promising alternative for VLA training, providing abundant real-world data with minimal reality gap. 
While prior work has employed various implicit learning approaches (e.g., contrastive learning~\cite{nair2022r3m}, masked autoencoder~\cite{radosavovic2023real}, latent action~\cite{bjorck2025gr00t}) to enhance robotic skills, the underlying learning mechanisms and transfer effectiveness of these approaches remain unclear.
Notably, these implicit methods fail to achieve the dramatic performance gains seen in LLMs/LMMs, where techniques like visual instruction tuning~\cite{liu2024improved} have demonstrated remarkable success.
We argue that this disparity stems from fundamental differences in data structure. 
In LLMs and LMMs, pretraining and downstream post-training data are isomorphic: textual reasoning aligns seamlessly with language tasks, and visual-text understanding transfers naturally to multimodal tasks.
In contrast, this alignment becomes heterogeneous in VLAs, where a significant gap exists between textual/2D visual inputs and the 3D action space with proprioceptive requirements.
Recent works have explored explicit human-centric representation~\cite{qiu2025humanoid} with promising results, but their limited training scale contradicts the original vision of leveraging web-scale data --- the very resource that enabled LLMs/LMMs' success through massive pretraining. 

In this paper, we aim to explore the following question:
\begin{center}
\begin{tcolorbox}[width=.9\textwidth, colframe=black!80, colback=gray!0, left=0.2cm, right=0.2cm, bottom=0.1cm, top=0.1cm]
Can we pretrain a dexterous VLA from large-scale human videos, analogous to GPT-3, to explicitly imitate human actions and adapt to robot hands via post-training?
\end{tcolorbox}
\end{center}

Our motivation is straightforward: human hands represent the gold standard for dexterous manipulation~\cite{kim2021integrated}, exhibiting remarkable versatility across countless tasks captured in natural settings.
Learning from their motions bridges the pretraining-downstream heterogeneity, but scaling this to massive human videos introduces new major challenges.
(1) Data Heterogeneity: 
Human video data spans varying camera systems, coordinate frames, and recording conditions, complicating model learning.  
Addressing this requires unifying disparate data sources while embedding essential 3D spatial reasoning capabilities.
(2) Hand Quantization: 
Unlike simple robotic actions, hand movements involve fine finger coordination that demands careful tokenization to preserve fine-grained control. 
Therefore, continuous hand motions are required to be discretized into language-compatible representations without sacrificing millimeter-level precision.
(3) Cross-Modal Reasoning: 
The model must learn intricate dependencies between visual observations, language instructions, and precise finger movements, a far more challenging task than traditional LMMs.
(4) Robot Control Transfer:
Human hand motions cannot be directly transferred to robots due to morphological differences, necessitating careful skill transfer to ensure learned strategies adapt effectively.
Addressing these challenges requires a comprehensive framework that systematically integrates data, motion representation, reasoning, and transfer learning. 

To achieve this, we introduce \textbf{\texttt{\ModelName}}, an advanced yet sample-efficient dexterous VLA trained on large-scale human videos, as illustrated in Figure~\ref{fig:001}.
To train this model, we propose \textbf{\texttt{Physical Instruction Tuning}}, a novel paradigm that extends visual instruction tuning~\cite{liu2023visual} to the physical domain through three key components: (1) VLA pretraining on human videos, (2) physical space alignment, and (3) post-training adaptation.
Unlike traditional visual instruction tuning, our approach addresses physical space alignment to unify heterogeneous data from diverse camera systems and recording conditions while embedding 3D spatial reasoning capabilities.
Moreover, we explicitly model hand motions as a prior to guide robotic post-training, unlike current implicit learning methods~\cite{bjorck2025gr00t, nair2022r3m} like GR00T~\cite{bjorck2025gr00t}.
\texttt{Being-H0} employs a unified autoregressive architecture with shared attention mechanisms across vision, language, and motion, enabling seamless cross-modal reasoning.
For precise motion tokenization, we introduce an effective part-level motion tokenization based on grouped residual quantization (GRQ)~\cite{lee2022autoregressive, yang2023hifi}, achieving millimeter-level reconstruction accuracy. 
To support large-scale learning, we curate \textbf{\texttt{UniHand}}, a comprehensive dataset of over 150M samples that integrates motion capture, VR recordings, and RGB-only videos across diverse manipulation tasks.
To the best of our knowledge, this is the first attempt to train dexterous VLAs based on explicit motion modeling from large-scale human videos.
The key contributions of our work can be summarized as follows:

\begin{itemize}[leftmargin=1.5em]
\item \textbf{Physical Instruction Tuning:} 
A new paradigm that establishes human hands as the foundational manipulator for robot hand transfer, bridging the gap between human videos and embodied action.
\item \textbf{Part-Level Motion Tokenization:}
A quantization method that preserves millimeter-level precision in continuous hand motions while enabling compatibility with the discrete architecture of autoregressive language models.
\item \textbf{UniHand:} 
A large-scale dataset with over one hundred fifty million instruction-following samples spanning diverse manipulation scenarios, which is collected by our scalable data pipeline unifying motion capture, VR, and RGB-only videos.
\item \textbf{Being-H0:} 
By integrating the innovations above, we present the first dexterous VLA trained on motion-based human video data at scale. 
Our model achieves robust cross-modal reasoning across vision, language, and fine-grained hand motions, with tailored adaptation strategies for downstream robotic manipulation tasks.
\end{itemize}

\section{Related Work}
\label{sec:related}

\noindent\textbf{Large Multimodal Models. }
The transformer architecture~\cite{vaswani2017attention} has revolutionized language models~\cite{radford2018improving, radford2019language, brown2020language} with powerful autoregressive text interpretation and generation ability.
This success has extended to large multimodal models (LMMs)~\cite{zhu2023minigpt,wang2024qwen2,zheng2024unicode,zhang2025beingvl0,zhang2025beingvl05}, which combine LLM reasoning~\cite{touvron2023llama,touvron2023llama2,bai2023qwen} with modal-specialized encoders~\cite{radford2021learning, zhai2023sigmoid} for unified multimodal understanding.
Pioneering works like Flamingo~\cite{alayrac2022flamingo} 
demonstrated strong few-shot VQA performance via cross-attention between visual and text inputs, while the LLaVA series~\cite{liu2024improved,liu2023visual,lu2023empirical} introduced visual instruction tuning to improve the instruction-following capabilities through carefully curated datasets.
These instruction-following datasets typically employ vision models to label an image, then use an LLM to generate QA pairs~\cite{you2023ferret,li2025otter}, or directly leverage proprietary LMMs for annotation~\cite{chen2024allava,zheng2023steve,feng2024videoorion,liu2025taking}.
While top-performing LMMs~\cite{team2023gemini,comanici2025gemini,achiam2023gpt} remain closed source, recent works have show increasing openness through released model weights~\cite{bai2025qwen2,team2024chameleon}, training details~\cite{zhu2025internvl3}, and even data recipes~\cite{deitke2024molmo}.

\noindent\textbf{Human-body Motion Generation. }
Recent advances in human-body motion modeling build upon increasing professional datasets and parametric models. 
Early motion datasets like KIT-ML~\cite{plappert2016kit} and AMASS~\cite{mahmood2019amass} have evolved into text-annotated benchmarks (e.g., HumanML3D~\cite{guo2022generating}, BABEL~\cite{punnakkal2021babel}) and large-scale collections of expressive motions (e.g., MotionX~\cite{lin2023motion}, EgoBody~\cite{zhang2022egobody}, Nymeria~\cite{ma2024nymeria}), culminating in million-sequence datasets like MotionLib~\cite{wangscaling}.
The field of human motion depends on models like SMPL~\cite{loper2023smpl} and its extension SMPL-X~\cite{pavlakos2019expressive} to provide parametric standardization for differentiable body representation.
In general, motion generation follows two dominant paradigms: diffusion-based models that produce high-fidelity motions~\cite{tevet2022human,zhang2024motiondiffuse,chen2023executing,yuan2023physdiff,zhang2023remodiffuse,lou2023diversemotion,zhang2024large}, and autoregressive approaches that excel at long-term modeling and textual reasoning.
For autoregressive models, they treat motion as a sequence of discrete tokens by using advanced tokenizers including VQ-VAE~\cite{van2017neural,zhang2023generating}, Residual Quantization (RQ)~\cite{lee2022autoregressive, guo2024momask}, Hierarchical Quantization (H2VQ)~\cite{you2022locally,lu2023humantomato}, and more recently, popular lookup-free quantization methods~\cite{mentzer2023finite, wangscaling,yu2023language}.
These models typically present advanced performance, especially on interpreting human intention by fine-tuning LLMs for motion-related tasks~\cite{jiang2023motiongpt, wang2024motiongpt,chen2024motionllm, zhou2024avatargpt,wangscaling,jiang2024motionchain,li2024lamp,li2025human}.
However, a key challenge exists that their generated motions lack physical realism, leading to artifacts like foot-sliding~\cite{tevet2022human}.
Considering this, recent studies have explored utilizing reinforcement learning (RL) and physical feedback to generate physically plausible motions for humanoid control~\cite{yue2025rl,luo2023perpetual, ji2024exbody2,han2025reindiffuse}.

\noindent\textbf{Hand Motion Generation. }
Unlike human body, hand motion research primarily focuses on hand-object interaction~\cite{grauman2024ego,jiang2021hand,liu2022joint} and fine-grained action precision.
While existing benchmarks~\cite{banerjee2025hot3d,liu2022hoi4d,zhan2024oakink2} capture interactions, their reliance on mocap systems or multi-camera setups limits diversity to tabletop scenarios, hindering generalization.
Instead, egocentric videos~\cite{grauman2024ego,grauman2022ego4d} from head-mounted cameras offer environmental diversity but often lack precise hand annotations. 
With progress~\cite{jiang2021hand,hasson2019learning} achieving accurate monocular reconstruction, advances in 3D hand modeling~\cite{pavlakos2024reconstructing,dong2024hamba,fan2024hold} now enable pseudo-label extraction from these data sources.
However, their weak perspective assumption restricts motion to camera frustums, making them incompatible with shifted-perspective datasets~\cite{damen2018scaling,grauman2022ego4d}. 
Dyn-HaMR~\cite{yu2025dyn} addresses this by integrating SLAM for camera tracking and occlusion-robust refinement.
Beyond modeling hands in isolation, recent efforts~\cite{liu2022joint} jointly focus on hand-object interaction by predicting interaction hotspots, future trajectories, and affordance.

Hand-object interaction (HOI) originated in 2D visual recognition~\cite{gkioxari2015contextual,mallya2016learning} and detection tasks~\cite{zheng2023open,gkioxari2018detecting,qi2018learning} but has progressively shifted to 3D learning for hand motion generation.
Early approaches employ multi-stage pipelines~\cite{christen2024diffh2o,cha2024text2hoi} to generate motions from action labels~\cite{ghosh2023imos,brahmbhatt2019contactdb}, while later methods adopt diffusion models~\cite{ho2020denoising} or autoregressive architectures, similar to human-body motion generation, with some leveraging LLMs as unified backbones for long-term temporal consistency~\cite{huang2025hoigpt}. 
Despite these advances, most methods overlook visual inputs until MEgoHand~\cite{zhou2025megohand}.
In this work, we equip our \texttt{Being-H0} with generation capabilities pretrained from large-scale human video data, providing hand motion priors for downstream manipulation tasks.
Note that we do not incorporate the interactive object modeling (e.g., 6D pose) in this version, leaving it as the future exploration.

\noindent\textbf{Learning VLAs from Human Videos. }
The progress of LMMs has enabled visual-language-action models (VLAs) to map multimodal inputs to physical actions. 
Early approaches like RT-1/2~\cite{brohan2022rt, brohan2023rt} quantize 7-DoF actions for autoregressive prediction.
Recently, OpenVLA~\cite{kim2024openvla} and subsequent works like $\pi_0$~\cite{black2410pi0,intelligence2025pi_} and GR00T-N1.5~\cite{bjorck2025gr00t} significantly expand capabilities through large-scale pretraining on datasets like Open X-Embodiment~\cite{o2024open} and Droid~\cite{khazatsky2024droid}.
FAST~\cite{pertsch2025fast} proposes discrete cosine transforms for fast and scalable training.
Instead of using discrete action tokens, Octo~\cite{team2024octo} and RDT-1B~\cite{liu2024rdt} adopt a diffusion head for flexible action prediction.
Despite progress, current datasets remain limited to small-scale lab collections --- particularly challenging for dexterous hand manipulation~\cite{zhong2025dexgraspvla} due to higher costs than simple grippers, leading to no real-world dexterous hands' data at scale available.
While simulation offers a scalable alternative~\cite{ye2025dex1b,deng2025graspvla,he2025dexvlg,huang2025efficient}, the sim-to-real gap remains substantial, causing difficulty especially for real-world deployments.

To address the data bottleneck, human videos offer a promising solution, with prior works extracting transferable representations like visual features~\cite{nair2022r3m}, 3D perceptual priors~\cite{xu2025egodtm}, and interaction knowledge (e.g., affordances, contacts, and grasps)~\cite{gavryushin2025maple,chen2025vidbot,ma2025glover++}.
However, these attempts fail to explicitly map human-to-robotic motions due to structural differences.
Some approaches address this at the visual level through data editing techniques, like image inpainting and rendering~\cite{lepert2025phantom}, visual masking~\cite{kareer2024egomimic}, to align observations across embodiments. 
Others focus on action-space alignment via unified human-centric state-action spaces~\cite{niu2025human2locoman, kareer2024egomimic,singh2024hand,yuan2025cross,qiu2025humanoid} or trajectory refinement using RL/physical simulation~\cite{li2025maniptrans, chen2024vividex, zhou2024learning, zhou2025you} to produce physically plausible and smoother actions.
Despite progress, existing works are limited to simplified grippers, as they neglect fine-grained finger motion alignment.
In this paper, we treat human hands as a universal standard for downstream manipulation, enabling robots to learn diverse skills from internet videos.
To our knowledge, we are the first to pretrain a scalable, generalizable VLA via explicit motion modeling from large-scale human videos.

\begin{figure*}[t]
\centering
    \includegraphics[width=1.0\linewidth]{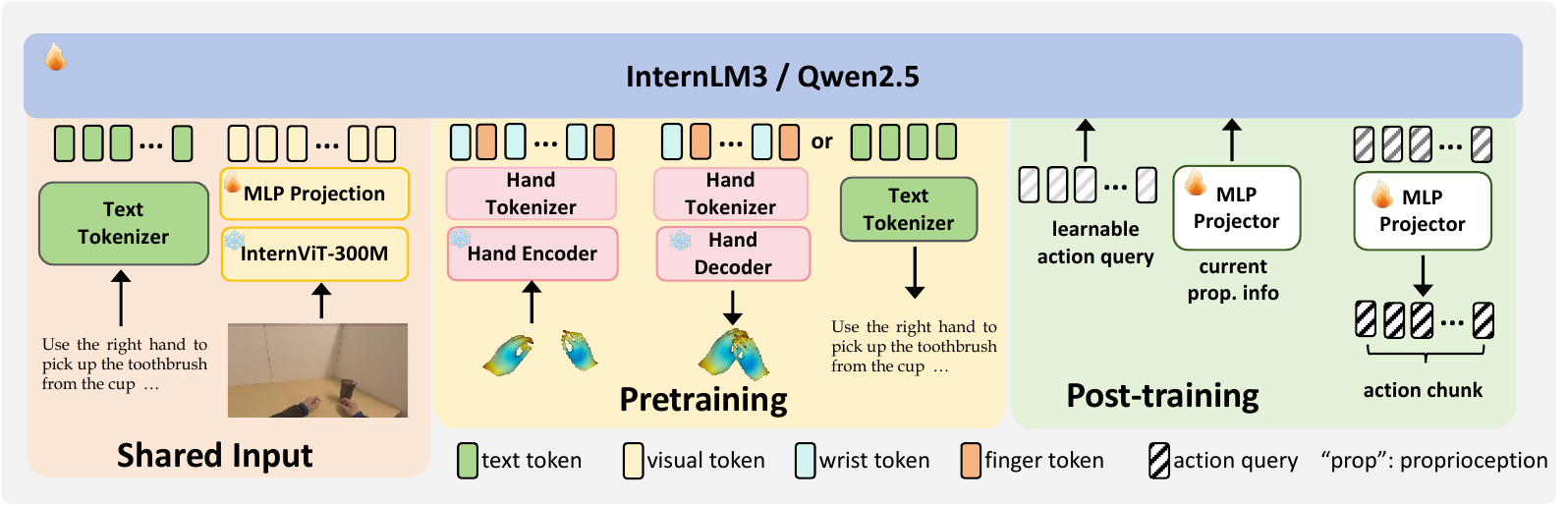}
\caption{\textbf{Overview of \texttt{Being-H0}.} 
The \textcolor{mygreen}{text tokenizer} and \textcolor{mywrite}{visual encoder} are shared by both pretraining and post-training. 
For pretraining and hand motion/translation tasks, \texttt{Being-H0} generates outputs in an autoregressive manner. For post-training and downstream manipulation tasks, \texttt{Being-H0} incorporates a set of learnable queries as the action chunk for prediction.}
\label{fig:being-h0}
\end{figure*}

\section{Overview of Being-H0}

\subsection{Motivation}

While current Vision-Language-Action models (VLAs) have shown remarkable progress, their performance significantly declines in novel scenarios and complex object interactions, falling short of LMMS like LLaVA~\cite{liu2024improved} in instruction following.
This limitation arises from reliance on either synthetic data with inherent sim-to-real gaps~\cite{rajeswaran2017learning,wang2022dexgraspnet} or limited teleoperated lab demonstrations that lack diversity~\cite{kim2024openvla}, particularly for dexterous manipulation tasks~\cite{kim2024openvla}.
Human activity videos offer a promising alternative but introduce new challenges (see Section~\ref{sec:challenge}).
To address these challenges, we analyze the success factors of visual instruction tuning and propose \textbf{\texttt{physical instruction tuning}}, a novel paradigm for training our dexterous VLA --- \textbf{\texttt{Being-H0}}, which includes three key components as shown in Figure~\ref{fig:being-h0}

\noindent\textbf{$\bigstar$ Pretraining. }
Existing LMMs excel at multimodal reasoning but underperform when adapted to Vision-Language-Action models (VLAs) for manipulation tasks. 
As discussed, we attribute this to the fundamental pretraining-downstream data mismatch. 
To bridge this gap, we leverage the anatomical similarity between human and robotic manipulators by introducing hand motion generation pretraining.
Our pretraining treats the human hand as the ideal manipulator, with robotic equivalents as simplified versions of its dexterity.
We train a foundation VLA to predict hand motions from vision and language using a multimodal dataset $\mathcal{D} = \{(\mathbf{v}_i, \mathbf{t}_i, \mathbf{m}_i)\}$, where $\mathbf{v}$ is visual input, $\mathbf{t}$ is language instruction, and $\mathbf{m} =\{\theta, \mathbf{r}_{rot}, \tau, \beta\}$ represents MANO-parameterized motions~\cite{romero2017embodied} (joint angles $\theta$, wrist rotation $\mathbf{r}_{rot}$, translation $\tau$, and shape $\beta$).
Each sample is treated as an instruction-following pair $\{\mathcal{X}_Q, \mathcal{X}_A\}$ optimized via:

\begin{equation}
\theta^* = \arg\min_\theta \sum_{i=1}^N \mathcal{L}(\Theta) = -\sum_{j=1}^{L} \log P_{\Theta}(y_j \mid \mathcal{X}_Q, \hat{y}_{1:j-1}),
\end{equation}

where $\Theta$ denotes the foundational model, $\mathcal{X}_A = \{y_i\}$ contains target tokens from text and motion modalities. 
This unified formulation supports flexible task specifications, including vision-to-motion generation, motion captioning, and multimodal conditioning for diverse hand-object interaction scenarios.

\noindent\textbf{$\bigstar$ Physical Space Alignment. }
Our pretraining bridges the vision-action gap to create a foundation VLA, but faces unique alignment challenges beyond standard visual instruction tuning.
The key difficulties arise from three aspects:
(1) The visual inputs from multiple sources vary in camera intrinsics and are captured under dynamic world coordinates. 
(2) The model's backbone is initialized with 2D vision-text pretraining, leaving it without crucial 3D spatial priors.
(3) Essential physical properties, like force and friction, which humans intuitively understand, are inherently missing in video data.
Unlike biological vision systems that organically develop 3D perception through embodied experience, we explicitly align these disparate data sources via physical space alignment --- unifying observations in a consistent coordinate system to instill 3D reasoning and physical understanding.

\noindent\textbf{$\bigstar$ Post-training. }
After pretraining with physical space alignment, we adapt our foundation VLA to downstream manipulation tasks.
In this work, we employ a straightforward MLP-based projection strategy and plan to explore more sophisticated approaches in the future.

\subsection{Challenges \& Solution}
\label{sec:challenge}

In Section~\ref{sec:introduction}, we raise two central questions:
(1) Can large-scale human videos enable pretraining of dexterous VLAs to explicitly understand and imitate human actions --- akin to how GPT-3 learns language through large-scale pretraining?
(2) Can such a pretrained model be effectively transferred to robotic manipulation via post-training adaptation?
To address these questions, we must overcome several critical challenges. 
Below, we examine these difficulties and outline our corresponding solutions:

\textbf{Pretraining Data Curation. } 
Current VLAs suffer from severe data scarcity compared to the NLP and CV domains.
While datasets like Open X-Embodiment~\cite{o2024open} and AgiBot~\cite{bu2025agibot} exist, they remain orders of magnitude smaller than existing multimodal benchmarks and primarily focus on end-effector control, neglecting fine-grained finger coordination due to hardware costs.
Human videos could potentially help this, but remain underutilized as most approaches primarily focus on implicit alignment (e.g., GR00T N1.5's latent action optimization~\cite{bjorck2025gr00t}) without proven benefits.
Recently, some works have started to explore text-to-motion based on lab-collected datasets~\cite{zhan2024oakink2} with precise labels.
However, these data are limited by their small scale (\textless 30K), thus lacking diversity and generalization.
Instead, in-the-wild datasets (e.g., Ego4D~\cite{grauman2024ego}) can provide scale, but these datasets suffer from camera inconsistencies and motion granularity issues.
Our solution systematically integrates heterogeneous sources through MANO parameter standardization~\cite{romero2017embodied} and weak-perspective alignment, creating a unified dataset spanning over 1,000 hours across more than 150 tasks.

\textbf{Precise Hand Quantization. } 
Our work treats hand motions as a foreign language, raising a key question:
``Can discrete motion tokens maintain sufficient precision for action prediction?''
While prior works suggest quantization disrupts pose continuity and loses precision, our VQ-based tokenizer~\cite{van2017neural} achieves millimeter-level reconstruction accuracy with careful design.
We discretize continuous MANO sequences $\mathcal{M} \in \mathbb{R}^{T \times D}$ with a 1D-Conv encoder which produces feature maps $z \in \mathbb{R}^{\lceil T/\alpha \rceil \times d}$ as follows:

\begin{equation}
\mathcal{M} \xrightarrow{{\rm Encoder}} z \xrightarrow{{\rm VQ}} \{m_1, \ldots, m_n\} \xrightarrow{{\rm Decoder}} \hat{\mathcal{M}},
\end{equation}

where $T$ denotes the frame number and $\alpha$ is the temporal downsampling rate.
Motion tokens $m_i \in \{\texttt{<motion\_id\_0>}, \ldots,$ $\texttt{<motion\_id\_K>}\}$ are delimited by $\texttt{<MOT>}$ and $\texttt{</MOT>}$ to form coherent motion blocks, ensuring seamless integration with text in our unified LMM model.

\textbf{Unified Cross-Modal Reasoning. } 
To model the intricate relationships between visual observations, language instructions, and hand motions, we process all modalities into a unified token sequence $\mathbf{S} = \{s_i\}$, where each token $s_i$ can represent text, vision, or motion. 
Visual tokens replace $\texttt{<IMG\_CONTEXT>}$ placeholders while motion tokens are structured into coherent blocks within the sequence.
Cross-modal interactions emerge via shared attention where the query $\mathbf{Q}_{v,t,m}$, key $\mathbf{K}_{v,t,m}$, and value $\mathbf{V}_{v,t,m}$ are computed from the concatenated states $\mathbf{H}_{v,t,m} = [\mathbf{H}_v; \mathbf{H}_t; \mathbf{H}_m]$.
This enables the model to learn rich multimodal dependencies: mapping visual scenes to manipulation strategies, grounding language instructions in precise finger motions, and aligning temporal motion patterns with task objectives.

\textbf{Adaptive Robot Control Transfer. }
Although the pretrained foundation VLA enables continuous motion generation while preserving broad capabilities, direct transfer of human hand motions to downstream manipulators remains challenging due to kinematic mismatches, varying degrees of freedom, and physical constraints.
To validate the effectiveness of large-scale learning from human videos, we adopt a simple MLP-based projection method that employs a fixed set of learnable queries as action chunks for downstream manipulators.
In future work, we aim to investigate more efficient and adaptive control transfer strategies to further bridge the gap between human motion and robotic action.

The following section elaborates on our technical solutions to train \textbf{\texttt{Being-H0}}, including the \textbf{\texttt{physical instruction tuning}} pipeline and our million-level hand motion dataset \textbf{\texttt{UniHand}}.

\section{Physical Instruction Tuning}

\subsection{\texorpdfstring{$\bigstar$}{} Pretraining}

Based on large-scale human videos, we pretrain a foundation Vision-Language-Action model (VLA) for diverse downstream manipulators from grippers to dexterous hands.
Human hands, with their high degrees of freedom, subsume the capabilities of existing devices; thus, \ModelName's human-centric training explicitly generalizes to a broad spectrum of manipulation tasks through kinematic mapping.
In this section, we first give a brief overview of model architecture (Section~\ref{sec:model_overview}), followed by how to precisely tokenize human hand motions (Section~\ref{sec:hand-motion-tokenization}).
Then, we detail the process of multimodal integration (Section~\ref{sec:mm_token_process}) and describe its training process (Section~\ref{sec:model_training}) and decoding modes (Section~\ref{sec:next_token_pred}).

\begin{figure*}[!ht]
\centering
    \includegraphics[width=1.0\linewidth]{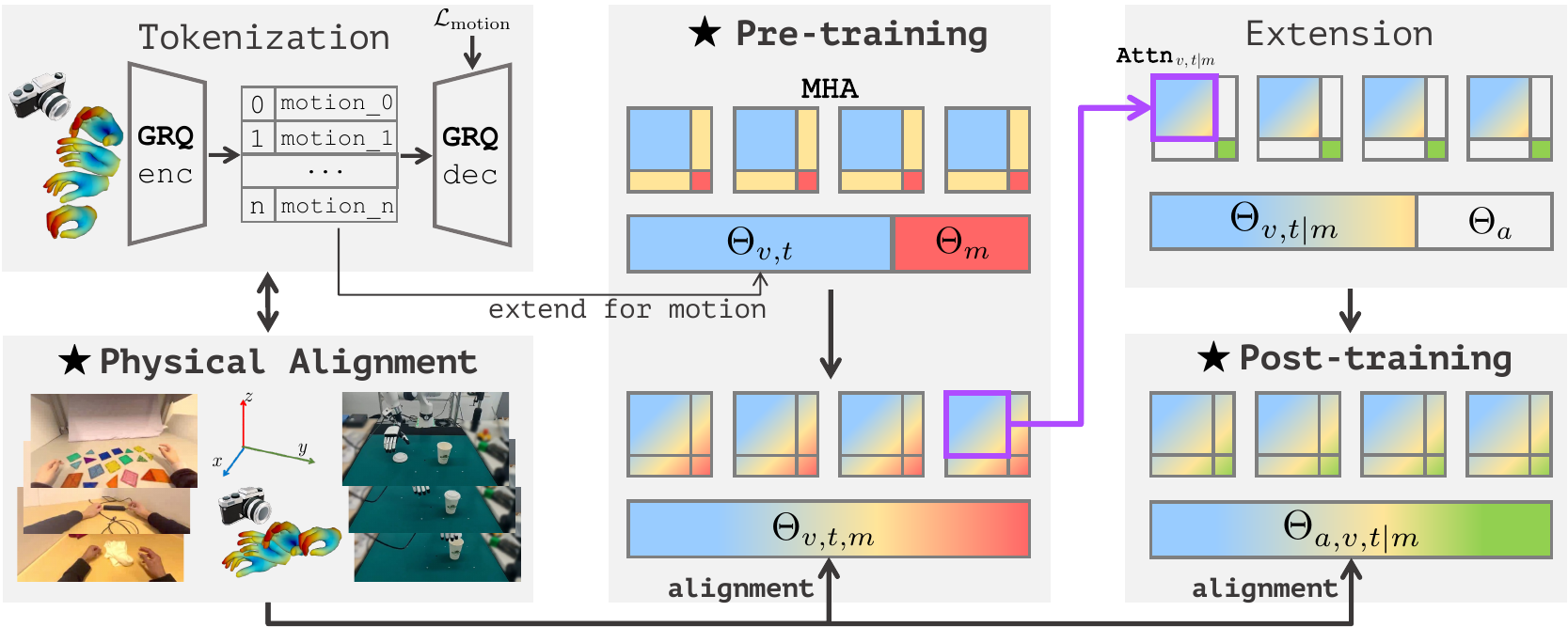}
\caption{\textbf{Physical Instruction Tuning.} 
Our training paradigm bridges human video datasets and robotic manipulation through a novel unified physical instruction tuning. 
\textbf{Left}: Part-level motion tokenization converts continuous hand motions into discrete tokens. Physical space alignment unifies heterogeneous data sources — from human hand demonstrations in the videos (dataset) to real-robot data — through coordinate system alignment and MANO parameterization, creating consistent representations for both pretraining and post-training supervision.
\textbf{Mid}: During pretraining, we extend vision-text parameters $\Theta_{v,t}$ to include motion parameters $\Theta_m$, enabling multi-head attention across vision, text, and motion tokens within a unified sequence. We use \textcolor[HTML]{5E9EFF}{blue} to denote visual and text attention, \textcolor[HTML]{FF0F00}{red} for motion attention, and \textcolor[HTML]{FFB934}{yellow} for cross-modal attention.
\textbf{Right}: The extension phase shows how attention mechanisms adapt to pretrained cross-modal dependencies ($\text{Attn}_{v,t|m}$), followed by post-training where action parameters $\Theta_a$ are incorporated to produce the final VLA with parameters $\Theta_{a,v,t|m}$ for downstream robotic tasks. The \textcolor[HTML]{5D8535}{green} part represents action attention.}
\label{fig:physical_instruction_tunning}
\end{figure*}

\subsubsection{Model Architecture}
\label{sec:model_overview}
Our model \ModelName is built on a pretrained LMM, specifically adopting the InternVL3 architecture~\cite{chen2024expanding}.
This backbone comprises a pre-trained InternViT-300M as the visual encoder and a 2-layer MLP as the projector.
At each timestep, the model processes an image-text pair input to predict hand motion sequences.
Following~\cite{chen2024far}, we implement a dynamic high-resolution strategy that tiles input images into patches while maintaining aspect ratios to minimize distortion, thereby preserving fine-grained visual details.
Inspired by~\cite{wangscaling}, we treat hand motion as a foreign language to facilitate seamless integration with the LMM.
During pre-training, a hand motion tokenizer quantizes continuous motion features into discrete embeddings.
To integrate motion tokens into the LMM backbone, we extend its vocabulary with $K$ discrete codes from a motion codebook.
We also introduce two special tokens, \texttt{<MOT>} and \texttt{</MOT>}, to mark the boundaries of the motion blocks. 

\subsubsection{Hand Motion Tokenization}
\label{sec:hand-motion-tokenization}
The motion tokenizer aims to encode $T$-frame hand features $\mathcal{M}=\{m_1,m_2,...,m_T\}$ of raw motion sequence into $\lceil T/\alpha \rceil$ token embeddings with dimensionality of $d$, where $\alpha$ denotes the temporal downsampling rate.

\noindent\textbf{Motion Feature. }
We use the 3D model MANO~\cite{romero2017embodied} to represent hand pose, which is parameterized as $m=\{\theta, \mathbf{r}_{rot}, \tau, \beta\}$.
However, it is critical to represent hand motion both efficiently and effectively. 
Therefore, selecting an appropriate feature space is crucial.
In this paper, we explore five alternative representations:

\begin{itemize}[leftmargin=1.5em]
    \item \textbf{MANO-D51}: Hand motion in each frame is encoded as $m \in \mathbb{R}^{51}$, consisting of $\theta \in \mathbb{R}^{15\times 3}$, $\mathbf{r}_{rot} \in \mathbb{R}^3$ and ${\tau} \in \mathbb{R}^3$, where $\theta$ and $\mathbf{r}_{rot}$ are represented as axis-angle form.
    \item \textbf{MANO-D99}: Hand motion in each frame is encoded as $m \in \mathbb{R}^{99}$. Unlike MANO-D51, this feature uses 6D rotations $\theta \in \mathbb{R}^{15\times 6}$ and $\mathbf{r}_{rot}\in \mathbb{R}^6$ instead of axis-angle. 
     \item \textbf{MANO-D109}: It extends MANO-D99 by additionally incorporating shape parameters $\beta \in \mathbb{R}^{10}$.
     \item \textbf{MANO-D114}: It extends MANO-D51 by adding joint positions $j\in \mathbb{R}^{21\times 3}$. Note that the joint positions only serve as auxiliary features during the reconstruction training, while in evaluation and inference, we solely utilize the 51 dimensional parameters.
     \item \textbf{MANO-D162}: Similar to MANO-D114, it extends MANO-D99 by adding joint positions $j\in \mathbb{R}^{21\times 3}$. 
\end{itemize}

In our experiments, we observe that the 6D rotation features achieve better reconstruction quality for finger joint rotations, while the axis-angle feature performs better for wrist poses. 
We attribute this phenomenon to the structural characteristics of different hand parts. The wrist, which typically exhibits larger but simpler rotations, benefits from the compactness and computational efficiency of the axis-angle formulations~\cite{bukschat2020efficientpose,mahendran20173d}. In contrast, finger rotations involve finer details that are better captured by the continuous nature and numerical stability of the 6D rotation representation.

Although the overall reconstruction error is lower when using axis-angle due to the dominant influence of wrist pose errors, we ultimately choose the 6D rotation feature for our hand motion tokenizer, as it yields better performance in \ModelName training. 
A possible explanation is that wrist pose patterns are relatively easier for LMMs to learn, whereas modeling fine-grained finger movements poses a greater challenge. Therefore, in this paper, we choose MANO-D162 as the feature for hand motion. We will explore the combination of axis-angle feature for wrist and 6D rotation for fingers in future work.

\begin{wrapfigure}{r}{0.5\textwidth}
    \centering
    \includegraphics[width=\linewidth]{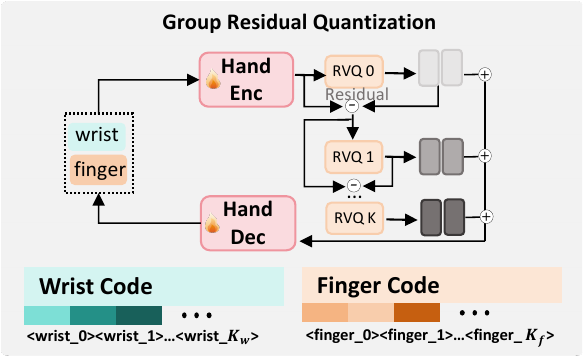}
    \caption{Architecture of part-level hand motion tokenization based on GRQ.}
    \label{fig:grq}
   \vspace{-10pt}  
\end{wrapfigure}

\noindent\textbf{Grouped Residual Quantization. } 
The motion tokenizer's precision critically influences both the quality of generated hand motions and the transferability of learned motion priors for downstream manipulation tasks. 
To ensure optimal performance, we carefully design a dedicated tokenizer for hand motion. 
Our architecture builds upon the Grouped Residual Quantizer Variational Autoencoder (GRQ-VAE)~\cite{yang2023hifi} as shown in Figure~\ref{fig:grq}. 
Given a motion sequence $\mathcal{M} \in \mathbb{R}^{T\times D}$, an encoder transforms it into a feature map $z\in \mathbb{R}^{\lceil T/\alpha \rceil \times d}$.
The tokenizer then discretizes the feature map $z$ through a multi-stage residual quantization process. 
First, the feature dimension $d$ is partitioned into $n$ groups along the channel axis, and each group is independently quantized using a residual vector quantizer (RQ)~\cite{lee2022autoregressive}. 
For each group $z^{(g)} \in \mathbb{R}^{\lceil T/\alpha \rceil \times d/n}$, a stack of $L$ codes from codebook $\mathcal{C}^{(g)}$ is applied, where $L$ represents the number of RQ layers. 
Therefore, the quantization for each feature vector $z^{(g)}_i$ in group $g$ proceeds as follows:

\begin{equation}
r_0 = z^{(g)}_i,\quad
q_l = \mathop{\arg\min}_{c \in \mathcal{C}^{(g)}} \|r_{l-1} - c\|_2,\quad
r_l = r_{l-1} - q_l,
\end{equation}

where $i \in \lceil T/\alpha \rceil$, $q_l$ and $r_l$ denote the selected code and residual at layer $l$.
Then, the final quantized representation is computed as:

\begin{equation}
\hat{z}^{(g)}_i = \sum_{l=1}^L q_l.
\end{equation}

By grouping feature channels and quantizing each group through residual stages, GRQ achieves a more expressive representation of motion inputs.
We observe that the reconstructing error of wrist parameters $\mathbf{r}_{rot}$ and $\tau$ primarily arises from their broader distribution spanning 3D space, despite their critical role in motion tokenizing precision. 
To address this, we introduce a wrist-specific reconstruction loss term to the overall quantization objective, which is defined as:

\begin{equation}
\mathcal{L}_{\text{wrist}} =||\mathbf{w} - \hat{\mathbf{w}}||_2^2,\quad
\mathbf{w}=[\mathbf{r}_{rot},\tau],
\end{equation}

where $\hat{\mathbf{w}}$ denotes the reconstructed wrist parameters.
The final objective combines reconstruction and commitment losses~\cite{lee2022autoregressive}, as well as the wrist term ($\lambda_{\{1,2\}}$ denotes balancing weights):

\begin{equation}
\mathcal{L}=\mathcal{L}_{\text{recon}} + \lambda_{1}\mathcal{L}_{\text{commit}} + \lambda_{2}\mathcal{L}_{\text{wrist}},
\end{equation}

and we employ an exponential moving average (EMA) update strategy for the codebook.

\noindent\textbf{Part-Level Motion Tokenizer. } 
Given the higher complexity of wrist parameter reconstruction compared to finger motions, we propose separate tokenizers for wrist and finger parameters inspired by~\cite{wangscaling}, such that each tokenizer can better model the part-level features. 
Specifically, the hand motion feature $m=\{\theta, \mathbf{r}_{rot}, \tau, \beta\}$ is decomposed into the wrist motion $\{\mathbf{r}_{rot}, \tau\}$ for global pose and accurate positioning, and the finger motion $\{\theta, \beta\}$ for fine-grained manipulation.
This part-level tokenization not only improves feature modeling but also provides explicit token semantics, enabling the LMM backbone to better capture structured hand dynamics.
When using the part-level tokenizer, the wrist loss $\mathcal{L}_{\text{wrist}}$ is omitted.

\subsubsection{Multimodal Integration}
\label{sec:mm_token_process}

Like traditional LLMs, it employs next token prediction to generate outputs.
\ModelName processes three modalities --- RGB vision, text, and hand motion --- uniformly by tokenizing them into discrete tokens.
While text is processed following LLMs, we elaborate on the process of the other two modalities in detail below.

\textbf{Vision Token. }
Visual inputs undergo processing to handle variable-resolution images and dynamic content complexity. 
Given an input image, we first employ dynamic patching, generating $N$ patches based on image content complexity. 
Following InternVL~\cite{zhu2025internvl3}, the patching strategy includes thumbnail generation to preserve global context: a downsampled version $I_{\rm thumb}$ with a pixel shuffle ratio of 0.5 is consistently retained alongside the detailed patches. 
The visual process first extracts features from these patches using a vision encoder, which are then projected into a unified embedding space via an MLP layer.
The vision tokens are structured with boundary markers \texttt{<IMG>} and \texttt{</IMG>}, while \texttt{<IMG\_CONTEXT>} serves as a placeholder token that is dynamically replaced by actual visual embeddings during processing.

\textbf{Motion Token. }
Motion data is first quantized before integration into the token stream.
Given the motion feature sequence represented as $\mathcal{M}$, the motion tokenizer quantizes it into a sequence of discrete tokens $\{m_i\}$. 
Motion sequences are structured with boundary tokens \texttt{<MOT>} and \texttt{</MOT>}, creating a motion block of $128$ tokens per second.
This structured representation ensures that motion information is clearly delineated within the token stream while maintaining compatibility with the transformer architecture.

\textbf{Multimodal Fusion. }
The model processes all modalities through a unified token space, utilizing shared embedding layers and attention mechanisms.
During fusion, vision tokens replace the \texttt{<IMG\_CONTEXT>} placeholders while motion tokens are inserted as structured blocks within the text sequence.
This generates a combined token sequence $\mathbf{S} = \{s_i\}$ where each element $s_i$ may represent text, visual, or motion content.
The attention mechanisms operate across modalities simultaneously.
For the concatenated multimodal hidden states $\mathbf{H}_{v,t,m} = [\mathbf{H}_v; \mathbf{H}_t; \mathbf{H}_m]$ (representing vision, text, and motion embeddings), we compute query, key and value through shared projections:

\begin{equation}
    \mathbf{Q}_{v,t,m} = \mathbf{W}_Q \mathbf{H}_{v,t,m}, \quad
    \mathbf{K}_{v,t,m} = \mathbf{W}_K \mathbf{H}_{v,t,m}, \quad
    \mathbf{V}_{v,t,m} = \mathbf{W}_V \mathbf{H}_{v,t,m}
\end{equation}

where $\mathbf{W}_{\{Q,K,V\}}$ denotes the weight matrices.
This architecture enables direct cross-modal attention, allowing the model to capture rich interdependencies between modalities, such as correlating visual observations with specific hand movements, or grounding language instructions in motion sequences.

Figure~\ref{fig:physical_instruction_tunning} illustrates this multimodal integration process, where our pretraining extends the original vision-text parameters $\Theta_{v,t}$ to incorporate motion parameters $\Theta_m$, enabling unified processing of all three modalities through shared attention mechanisms.
The model learns to generate coherent motion sequences by predicting discrete motion tokens within the broader context of visual observations and language instructions.

\subsubsection{Training Details}
\label{sec:model_training}

\noindent\textbf{Hand Motion Tokenizer. }
We sample hand motion sequences at 15 frames per second (FPS) and tokenize them using fixed one-second windows.
Noting that camera shift may occur within the temporal window, and \texttt{Being-H0} does not predict camera motion during inference.
Considering this, we transform each sequence into the coordinate system of its first frame.
To enable coherent generation of longer sequences, where each one-second segment within a multi-second output should be represented relative to the first frame of the entire sequence, we implement a specialized training strategy.
For each one-second sample, we randomly select a reference frame from a larger temporal window (10 seconds in this paper) and transform the motion sequence relative to this frame.
This approach allows the learned motion tokens to effectively represent movements relative to varying world coordinate systems while preserving long-term consistency.

\noindent\textbf{Foundation VLA. }
The model is trained using standard next-token prediction.
To optimize the integrated motion codes, we introduce a dual-level masking strategy to constrain the optimization at both the vocabulary and token-wise levels for more effective training.

\begin{itemize}[leftmargin=1.5em]
\item \textbf{Vocabulary-Level Logit Masking. }
Since motion codes $\mathcal{V}_{\rm motion}$ occupy only a small portion of the foundation VLA's vocabulary $\mathcal{V}$, we selectively mask non-motion logits on motion labels with a probability of $\mathcal{P}$.
This would help to focus gradient updates on motion-related embedding space and avoid being diluted by trivial non-motion generations.
Given predicted logits $\mathbf{z} \in \mathbb{R}^{| \mathcal{V}|}$, we apply masking as:

\begin{equation}
\tilde{\mathbf{z}}_i = 
\begin{cases}
\mathbf{z}_i & i \in  \mathcal{V}_\text{motion} \\
-\infty & \text{otherwise}.
\end{cases}
\quad \text{(with probability } \mathcal{P} \text{)}.
\end{equation}

\item \textbf{Token-level Loss Masking. }  The token-wise cross-entropy losses are computed using the masked logits $\tilde{\mathbf{z}}$.  
To address the natural variation in hand motion complexity (from static poses to unpredictable jitters), we filter extreme loss values, focusing learning on moderately challenging tokens.
For per-token losses $L = \{\ell_1, \ldots, \ell_N\}$, we define the filtered loss set as:

\begin{equation}
\tilde{L} = \big\{ \ell_i \in L \mid Q_\text{low} \leq \ell_i \leq Q_\text{high} \big\}.
\end{equation}

where $Q_\text{low}$, $Q_\text{high}$ denote the lower and upper percentiles of preset percentages.
\end{itemize}
\noindent  The final motion loss is computed as the mean over the masked losses:

\begin{equation}
\mathcal{L}_{\rm motion} = \frac{1}{|\tilde{L}|} \sum_{\ell_i \in \tilde{L}} \ell_i.
\end{equation}

\subsubsection{Decoding Modes}  
\label{sec:next_token_pred}

Our pretrained foundation VLA is able to generate both textual and motion outputs through unified next-token prediction, with valid motion blocks decoded into MANO parameters via the motion tokenizer.
In this work, we provide three decoding modes to balance generation flexibility and the structural consistency required to recover valid hand motion:   
\textbf{(1) Free-format Mode.} 
This mode allows fully flexible autoregressive sampling without any explicit structural constraints.
However, it risks producing invalid motion blocks that fail to decode.
\textbf{(2) Block-formatted Mode.} 
This mode enforces structural consistency to the expected motion block by restricting sampling to the motion token vocabulary only between \texttt{<MOT>} and \texttt{</MOT>} delimiters.
For evaluation with ground-truth motion block counts, we avoid premature generation of the \texttt{<EOS>} token until generating the target number of motion blocks.
\textbf{(3) Soft-formatted Mode.}
To evaluate local motion generation quality while accounting for the one-to-many nature between motion instruction and sequences, we employ a soft constraint strategy.
After generating each motion block, we blend the predicted and ground-truth MANO parameters via their mean, then re-quantize this hybrid through the motion tokenizer.
This enforces the generation anchored in a plausible neighborhood of the ground-truth, providing a more reliable estimate of the model's ability to produce high-quality motion in the vicinity of real trajectories. 

\subsection{\texorpdfstring{$\bigstar$}{} Physical Space Alignment}

To build a sufficiently large-scale dataset of dexterous human videos, we collect samples from diverse datasets and public sources.
However, this approach introduces variability in camera systems, posing challenges for effective pretraining.
Furthermore, existing LMMs exhibit limited 3D perception capabilities --- a well-documented limitation in prior research~\cite{yuan2024robopoint,ma20243dsrbench,yuan2025seeing,yang2025thinking}.
To alleviate this, we introduce physical space alignment, a unified toolkit that maps videos recorded by different cameras into a consistent physical space, meanwhile incorporating 3D spatial reasoning and physical attributes (if available) to enhance geometric and perceptual consistency across the datasets.
As illustrated in Figure~\ref{fig:physical_instruction_tunning}, this alignment process serves as the critical bridge between human demonstrations from videos and robotic manipulations, enabling consistent MANO parameterization across both domains for effective learning.
We introduce two strategies below, with more to be explored in future work.

\subsubsection{Weak-Perspective Projection Alignment}

Inherent differences in camera systems across data sources introduce inconsistent projection of 3D space. 
Although humans can intuitively perceive depth and estimate distances between hands and objects for grasping, models trained on such multi-sourced datasets frequently struggle with accurately mapping the image projections to the actual 3D context, leading to errors in 3D spatial reasoning. To alleviate this, we establish a unified weak-perspective camera space, which ensures consistent alignment from 2D visual content to a shared 3D reference frame.
This approach maintains uniform pixel scales for objects at similar depths, mitigating the inconsistencies caused by differing camera intrinsics.
Specifically, given a source image with camera intrinsics $K = \{f_x, f_y, c_x, c_y\}$ and target camera intrinsics $K' = \{f_x', f_y', c_x', c_y'\}$, we compute scale factors and translation offsets as:

\begin{equation}
s_x = \frac{f_x'}{f_x}, \quad 
s_y = \frac{f_y'}{f_y}, \quad 
\Delta x = c_x' - s_x \cdot c_x, \quad 
\Delta y = c_y' - s_y \cdot c_y.
\end{equation}

Each pixel $(u, v)$ in the source image is then transformed to $(u', v')$ via:

\begin{equation}
u' = s_x \cdot u + \Delta x, \quad 
v' = s_y \cdot v + \Delta y,   
\end{equation}

The source image is further cropped or padded to the same target resolution $(\mathcal{W}, \mathcal{H})$.
For videos with severe lens distortion (e.g., fisheye cameras), we first normalize the field of view (FoV) to $90^\circ$ to minimize projection artifacts before alignment.

\subsubsection{View-Invariant Motion Distribution Balancing}
\label{sec:view-balance}
Developing robust instruction-following capabilities requires careful pre-processing of instructional tuning data to ensure a balanced data distribution, particularly for physical instruction tuning.
If a single camera configuration dominates the dataset, it may introduce bias into the 3D perception system, ultimately limiting the model’s generalization ability in unseen camera settings.
To mitigate this issue, we propose a novel distribution-balancing strategy to augment video-motion pairs from small-scale data sources, preventing them from being overshadowed by larger ones. 
During balancing, we vary hand pose distribution without introducing camera viewpoint and position changes.
Importantly, our approach preserves weak-perspective consistency across motions from different sources, ensuring coherent 3D understanding.
Unlike conventional image-only augmentations (e.g., random cropping or flipping) that may disrupt weak-perspective alignment between hand motion and visual observations, our strategy employs two complementary components:

\noindent\textbf{Depth Scaling.}  
For a hand pose in camera coordinate $m_c = \{\beta, \theta, R_c, \tau_c\}$, where $\tau_c = (\tau_c^x, \tau_c^y, \tau_c^z)$ denotes the wrist’s 3D position and $R_c\in\mathbb{R}^{3\times3}$ is the rotation matrix form of $\mathbf{r}_{rot}$, we perturb human hand’s depth by randomly sampling a scaling factor $\lambda_s$:

\begin{equation}
\tau_c^{z'} = \lambda_s \cdot \tau_c^z
\end{equation}

To maintain weak-perspective consistency, the paired image is rescaled by $1/\lambda_s$, yielding an intermediate resolution $(\mathcal{W} / \lambda_s, \mathcal{H} / \lambda_s)$.
We constrain $\lambda_s$ to plausible ranges to avoid unrealistic perspective distortions caused by the non-negligible physical size of the hand.

\noindent\textbf{In-Plane Rotation.} 
To diversify hand positions in the image plane without introducing geometric inconsistencies, we apply a rotation around the camera's optical axis (Z-axis) with a uniformly sampled angle $\varphi$. 
This updates both wrist position and global rotation:

\begin{equation}
\tau_c' = R_z(\varphi) \cdot \tau_c,\quad
{R_c}' = R_z(\varphi) \cdot R_c.
\end{equation}

Here, $R_z(\varphi)$ is the rotation transform matrix. The corresponding image is rotated synchronously by $\varphi$ to maintain consistency.
All transformed frames are then adjusted to the target resolution $(\mathcal{W}, \mathcal{H})$, ensuring weak-perspective projection integrity while increasing hand diversity.

\subsubsection{Further Discussion}
Beyond the two strategies proposed above, we believe that integrating richer physical cues can further improve the model's understanding of spatial and physical environments. For instance, incorporating visual depth information, tactile feedback, or other multi-sensory signals may provide more grounded and embodied representations of human activities. These modalities offer complementary perspectives on physical interactions and 3D structure, which are often ambiguous or underspecified in 2D visual inputs alone.

Such multi-sensory integration could address fundamental limitations inherent in vision-only approaches. Depth information from RGB-D sensors could resolve spatial ambiguities that arise from weak-perspective projection, while tactile feedback could capture crucial contact dynamics, grip forces, and material properties that are invisible in visual observations but essential for successful manipulation. Audio signals from object interactions could further disambiguate manipulation strategies that appear visually similar but involve different physical processes, such as distinguishing between gentle placement and firm pressing actions.

These enhanced alignment strategies could create more robust representations that better capture the rich physical understanding humans naturally possess during manipulation tasks. As we scale our approach to larger and more diverse datasets, incorporating such multi-modal physical cues will become increasingly important for bridging the gap between human demonstration data and reliable robotic deployment across varied real-world scenarios.




\subsection{\texorpdfstring{$\bigstar$}{} Post-Training for Dexterous Manipulation}

Following pretraining and physical space alignment, our foundation VLA possesses comprehensive vision-language-motion understanding but requires adaptation to specific robotic manipulation tasks. As illustrated in Figure~\ref{fig:physical_instruction_tunning}, this post-training stage extends our model parameters from $\Theta_{v,t,m}$ to $\Theta_{a,v,t|m}$, incorporating action parameters $\Theta_a$ that enable direct robotic control while leveraging the rich multimodal representations learned during pretraining.
In this paper, we validate our approach using dexterous hands as the downstream manipulator.
Even though our paradigm remains adaptable to various robotic hand configurations, including grippers.
For instance, wrist motion can serve as a prior to guide end-effector movement, which has previously been explored~\cite{kareer2024egomimic}.
Therefore, we do not elaborate on further.

The kinematic differences between human and robotic hands prevent direct transfer of the foundation VLA and its motion tokens.
Since our major focus is dexterous VLA learning from large-scale human videos, we bridge this gap with a non-autoregressive MLP-based projection method as shown in Figure~\ref{fig:being-h0}, leaving further alternatives (e.g., discrete action tokens or diffusion policies) for future work.
Specifically, we leverage the VLA backbone as a pretrained encoder, where a lightweight MLP $f_p$ projects the dexterous hand's proprioceptive states into its embedding space.
The proprioceptive embedding is combined with visual-text tokens to form a unified context (\texttt{ctx}), enabling joint reasoning about sensory inputs, language, and current physical configuration.
For action generation, we employ a set of learnable query tokens $\{\mathbf{q}_1, \ldots, \mathbf{q}_{N_{\rm a}}\}$ that attend to these contexts within the pretrained encoder, with a regression policy head MLP $f_r$ transforming the pretrained encoder's outputs into executable dexterous poses.
The objective of post-training is to replicate expert demonstrations through imitation learning. 
Given an expert action sequence $\mathbf{a}^* = \{\mathbf{a}^*_i\}$, we optimize the model by minimizing the L1 loss between predicted actions $\mathbf{a}$ and the expert data:

\begin{equation}
\mathbf{a}_i = f_r\Bigl(\Theta\bigl(\mathbf{q}_i, \texttt{ctx} \oplus f_p(\mathbf{p}_t)\bigr)\Bigr), 
\quad 
\mathcal{L}(\Theta_{\mathrm{a}}) = \frac{1}{N_a} \sum_{i=1}^{N_a} \|\mathbf{a}_i - \mathbf{a}^*_i\|_1,
\end{equation}

where $N_a$ represents the action chunk size and $\Theta_{\rm a}$ denotes all trainable parameters during post-training, including the foundation VLA $\Theta$, the action queries $\mathbf{q}_i$, the proprioceptive projector $f_p$, and the policy head $f_r$. 
The term $\texttt{ctx}$ represents the visual and textual context, and $\oplus$ denotes the concatenation operation.
The term $\mathbf{a}^*_i \in \mathbb{R}^{D_{\rm a}}$ denotes the actions from expert demonstrations, and $D_a$ is the dimension of the action space.
This approach effectively evolves the pretrained foundation VLA to produce robot-executable control while maintaining cross-modal reasoning, with supportive tasks like: motion generation from visual-textual inputs, text-based captioning of observed motions, and robot control adaptation via domain-specific fine-tuning.

\begin{figure*}
\centering
\includegraphics[width=1\textwidth]{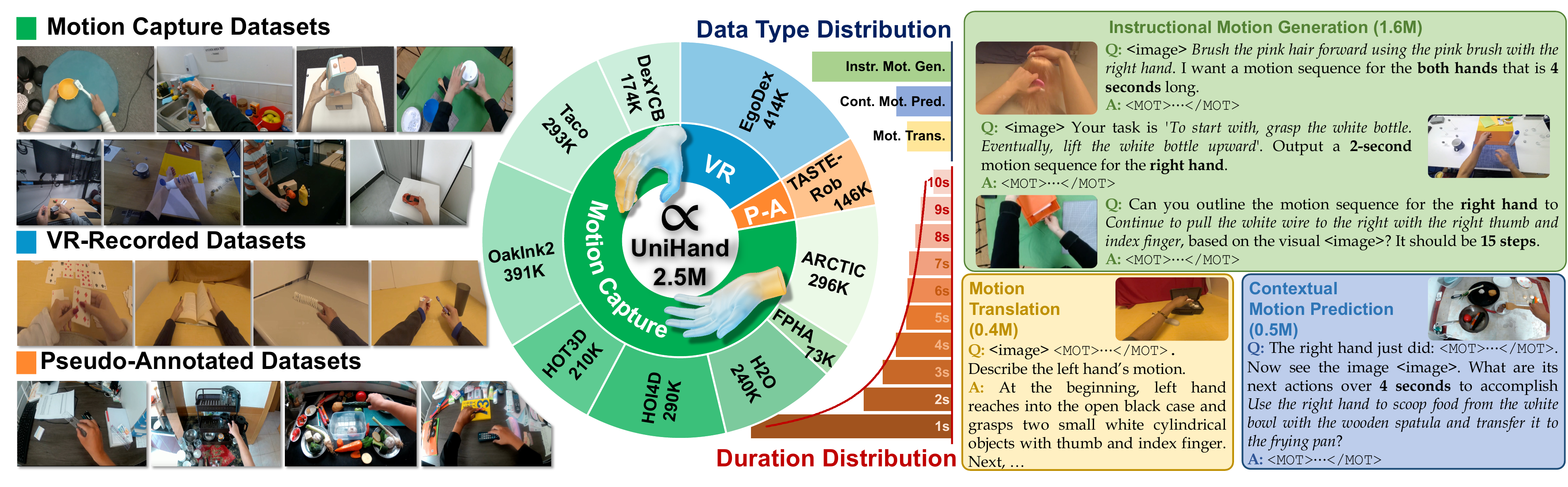}
\caption{ The overview of our \texttt{UniHand-2.5M}. \textbf{Left}: The scenes and tasks from different data source types.
\textbf{Mid}: The distribution of different data sources, data types, and durations. \textbf{Right}: Samples from different data types.
}
\label{tab:unihand2-5} 
\end{figure*}

\section{\DataName: Scaling up Hand Motion Instructional Data}

Our pretrained foundation VLA requires massive human videos to learn diverse dexterous skills.
In this section, we introduce the data curation pipeline in detail.

\subsection{Data Sources and Statistics}

We curate our dataset from three primary sources, each with distinct advantages and trade-offs: 
\textbf{(1) Motion capture datasets}~\cite{fan2023arctic,garcia2018first,kwon2021h2o,liu2022hoi4d}, which incorporate high-precision 3D annotations from multi-view motion capture systems in controlled environments (e.g., studio, lab), though their diversity is often limited.
For example, OAKINK2~\cite{zhan2024oakink2} offers multi-view, object-centric recordings of real-world bimanual manipulation, including complex tasks.
\textbf{(2) VR-recorded datasets}, which leverage VR devices (e.g., Apple Vision Pro) with calibrated cameras and SLAM-based tracking to capture natural hand-object interactions in less constrained settings while maintaining reliable 3D ground truth.
A notable example is EgoDex~\cite{hoque2025egodex}, which includes up to 194 household manipulation tasks such as tying shoelaces and folding laundry.
\textbf{(3) Pseudo-annotated datasets}~\cite{perrett2025hd}, which we utilize off-the-shelf hand motion predictors~\cite{perrett2025hd} to generate pseudo 3D labels from in-the-wild videos. 
While noisier, these datasets excel in scalability and diversity, as seen in prior large-scale applications~\cite{cai2023smpler}.
For instance, Taste-Rob~\cite{zhao2025taste} includes around 100K egocentric videos with aligned language instructions recorded from a fixed viewpoint.
We exclude samples with frequent hand occlusion, out-of-frame (e.g., Ego4D~\cite{grauman2022ego4d}), or dynamic camera viewpoints despite cleaner visuals (e.g., EPIC-Kitchen~\cite{damen2018scaling}), leaving the process of these datasets in our future work.

The recipe of our dataset \DataName is shown in Table~\ref{tab:dataset}.
Our dataset is aggregated from 11 sources with detailed hand motion annotations along with corresponding RGB observations. In total, \DataName contains more than 440K task trajectories, covering over 130 million frames and over 1,100 hours of video, thus offering high diversity and strong coverage of real-world scenarios. Based on this rich data foundation, our instructional data labeling pipeline (detailed in the following sections) produces over 165 million motion-instruction pairs, offering large-scale supervision for dexterous VLA learning.
Moreover, since our data curation pipeline supports both VR-recorded datasets and pseudo-annotated datasets, while the former is relatively easy to collect at scale, and the latter allows us to leverage the vast pool of in-the-wild human videos, more diverse data is on the way to be incorporated in the future.
Due to computational cost constraints, we sample 2.5 million instruction data points from \DataName for pretraining. This subset is selected based on a balanced sampling strategy to ensure diversity in task types and sources, which we refer to as \texttt{UniHand-2.5M}, as shown in Figure~\ref{tab:unihand2-5}.

\begin{table*}[ht]
\centering
\caption{Statistics of \DataName. Our dataset aggregates hand motion data from 11 benchmarks across three types: motion capture, VR-recording, and pseudo-annotation. 
To the best of our knowledge, this represents the largest egocentric hand motion dataset to date. 
By regarding human hands as the golden standard for dexterous hands or grippers, we anticipate that \DataName will establish foundational references for VLA learning. \textbf{\#Inst} refers to the number of our generated instructional samples.}
\vspace{-2mm}
\setlength{\tabcolsep}{3pt}{
\scalebox{0.9}{
\begin{tabular}{l|c|c|c|c|c|c|c|c}
\toprule
 \textbf{Dataset} & \textbf{\#Inst} & \textbf{\#Seq} & \textbf{\#Avg len} & \textbf{\#Frames} & \textbf{\#Hours} & \textbf{Hand joint} & \textbf{Hand Pose}  & \textbf{Ann Granularity}  \\
\midrule
ARCTIC~\cite{fan2023arctic}                & 17.9M & 0.3K & 725.3 & 245K & 2.3 & \textcolor{blue}{\Checkmark} & \textcolor{blue}{\Checkmark} & Action   \\
FPHA~\cite{garcia2018first}        & 798K & 1.2K & 89.8 & 105K & 1.0 & \textcolor{blue}{\Checkmark} & \textcolor{magenta}{\XSolidBrush} & Action   \\
HoloAssist~\cite{wang2023holoassist} & 8.0M & 2.2K & 8081.7 & 17.1M & 166 & \textcolor{blue}{\Checkmark} & \textcolor{magenta}{\XSolidBrush} & Segment  \\
H2O~\cite{kwon2021h2o} & 3.7M & 0.9K & 121.5 & 115K & 1.1 & \textcolor{blue}{\Checkmark} & \textcolor{blue}{\Checkmark} & Action   \\
HOI4D~\cite{liu2022hoi4d}  & 21.2M & 3.0K & 273.0 & 825K & 7.6 & \textcolor{blue}{\Checkmark} & \textcolor{blue}{\Checkmark} & Action \\
HOT3D~\cite{banerjee2025hot3d}  & 8.7M & 2.8K & 150.0 & 420K & 3.9 & \textcolor{blue}{\Checkmark} & \textcolor{blue}{\Checkmark} & N/A  \\
OAKINK2~\cite{zhan2024oakink2} & 18.5M & 2.8K & 244.4 & 695K & 6.5 & \textcolor{blue}{\Checkmark} & \textcolor{blue}{\Checkmark} & Action \\
TACO~\cite{liu2024taco}   & 11.5M & 2.2K & 154.0 & 340K & 3.2 & \textcolor{blue}{\Checkmark} & \textcolor{blue}{\Checkmark} & Action \\
DexYCB~\cite{chao2021dexycb} & 3.6M & 5.6K & 72.8 & 410K & 3.8 &  \textcolor{blue}{\Checkmark} & \textcolor{blue}{\Checkmark} & N/A \\
Taste-Rob~\cite{zhao2025taste} & 1.9M & 85K & 164.1 & 14M& 130 & \textcolor{magenta}{\XSolidBrush} &\textcolor{magenta}{\XSolidBrush}  & Trajectory\\
EgoDex~\cite{hoque2025egodex} & 70.6M & 338K & 264.8 & 89.6M & 829.4 & \textcolor{blue}{\Checkmark} & \textcolor{magenta}{\XSolidBrush} & Trajectory \\
\midrule
Total & 166.5M & 444.1K & - & 130M & 1155 & \textcolor{blue}{\Checkmark} & \textcolor{blue}{\Checkmark} & Fine-Grained \\
\bottomrule
\end{tabular}}}
\label{tab:dataset}
\end{table*}

\subsection{Data Curation Pipeline}
\label{sec:data_curation}

\subsubsection{Hand Pose Standardization}
Our model learns an explicit mapping from 2D visual observations to 3D spatial coordinates by treating hand motions as 3D signals, ensuring both geometric precision and visual-semantic consistency.
To address the heterogeneity in motion labels across datasets, we combine different data sources through hand pose standardization. 
For datasets with motion capture or SLAM-tracked labels, we directly extract their annotations in the form of MANO parameters~\cite{romero2017embodied}.
When only 3D hand joint positions are available, we derive the corresponding MANO parameters via gradient-based optimization.
In cases where datasets lack 3D hand pose or joint annotations entirely, we leverage HaMer~\cite{pavlakos2024reconstructing} for frame-wise pose estimation to maintain consistent motion semantics. 
To enhance the reliability of HaMer's output, we detect and correct left-right hand mismatches by identifying pose discontinuities, followed by applying temporal interpolation to fill minor gaps.
Additionally, the fitting process incorporates joint angle constraints and temporal smoothness regularization to ensure physically plausible and coherent hand motions.

\subsubsection{Task Description Labeling}

To establish strong semantic grounding between vision, language, and motion, we introduce a structured hierarchical labeling framework that enriches motion semantics beyond the sparse or imprecise textual labels in existing datasets. 
This framework provides detailed and consistent textual descriptions, enabling our VLA to effectively align visual inputs, natural language instructions, and quantized hand motion representations.
For structured coverage, we segment each video into non-overlapping chunks with a maximum length of 10 seconds, ensuring each captures a distinct phase of the task. 
We then samples frames at 2FPS and leverage \texttt{Gemini-2.5-Flash-Lite}~\cite{comanici2025gemini} to generate annotations at two temporal levels: 
At the  \textbf{chunk level}, we produce imperative instructions and concise summaries that describe overarching hand activities and object interactions. 
At the more granular \textbf{per-second level}, we further divide each chunk into overlapping 1-second windows, annotating them with precise instructions and captions that detail contact states, object attributes, hand parts, and motion trajectories relative to the camera perspective. 
To ensure clarity and completeness, we separately annotate global two-handed actions and individual hand actions, capturing both bilateral and unilateral descriptions.
This multi-scale labeling strategy guarantees comprehensive and consistent coverage, bridging high-level task objectives with fine-grained hand-object interactions in a unified framework.


\subsubsection{Instructional Data Generation. } 

Building on our systematic annotations, we construct instruction-following training data to explicitly establish rich vision-language-motion alignment for the foundation VLA. 
To achieve this, the designed instructional tasks focus on multiple grounding aspects for hand motion understanding, including spatial-temporal alignment of hand trajectories with visual context, precise object attributes and contact states, clear action intentions, and consistency between high-level instructions and fine-grained motion steps. 
Guided by these principles, we develop training data for three complementary task types: 
\textbf{(1) Instructional motion generation}, where the model learns to generate step-by-step motion sequences conditioned on scene images and task instructions; 
\textbf{(2) Motion translation}, where the model is required to convert the motion sequence and visual cues into language descriptions of hand-object interactions; and 
\textbf{(3) Contextual motion prediction}, where the model is asked to anticipate subsequent motion sequences based on prior motion history, the current scene observation, and optional instructions or task goals.

For implementation, we design approximately 20 base templates per task type and employ \texttt{Gemini-2.5-Pro} to generate diverse instructional variants.
Each template explicitly incorporates target duration specifications, enabling the model to handle varying temporal granularities and sequence lengths.
Through rule-based instantiation, we populate these templates with grounded instructions, motion tokens, and explicit length constraints.

\begin{wrapfigure}{r}{0.48\textwidth}
    \centering
    \vspace{-15pt}      
    \includegraphics[width=0.46\textwidth]{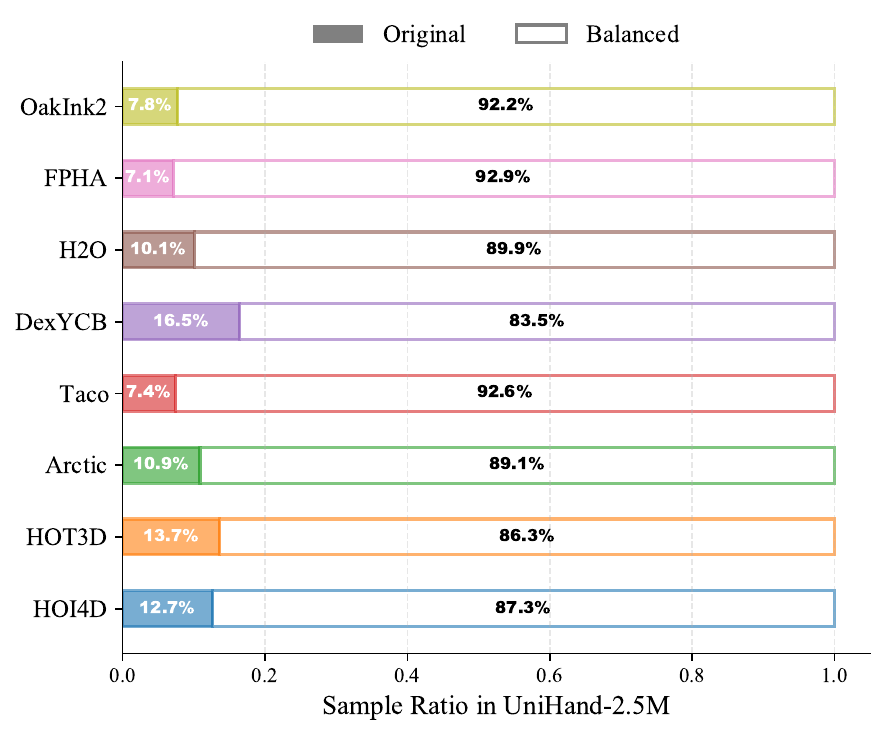}
    \vspace{-5pt}
    \caption{Comparison of the proportion of instruction samples derived from original data vs. view-balanced data in \texttt{UniHand-2.5M}.}
    \label{fig:data_balance_ratio}
   \vspace{-10pt}  
\end{wrapfigure}

To ensure a balanced visual perspective distribution in the training set, we apply the view-invariant motion distribution balancing method described in Section~\ref{sec:view-balance} to augment the data. Based on the balanced dataset, our pipeline produces over 165 million high-quality instructional pairs spanning multiple temporal scales, handedness configurations, and manipulation scenarios, with systematic quality checks for semantic coherence. To furthermore balance the distribution of data sources and task types in the training data, we sample a subset with 2.5 million instances from the full dataset that achieves a more even coverage of both task categories and data origins. The distribution of different tasks and data sources within \texttt{UniHand-2.5M} is shown in Figure~\ref{tab:unihand2-5}, while the proportion of samples generated from the view-balanced data is illustrated in Figure~\ref{fig:data_balance_ratio}.
This unified design provides robust supervision for the model to learn consistent mappings between vision, language, and structured motion.
 of bilateral and unilateral hand-object interactions. Together, this structured multi-scale annotation framework ensures comprehensive and consistent coverage of both high-level task objectives and fine-grained hand-object interactions, providing rich motion data for downstream modeling and analysis.

\section{Experiments}

In this section, we first introduce our experimental setups in Section~\ref{sec:exp_setup}, including details of the implementation and evaluation.
To validate \texttt{Being-H0}, we demonstrate our principal comparison on hand motion generation and translation tasks in Section~\ref{sec:comp_hand_gen_translate} and on long-range motion generation in Section~\ref{sec:exp_long_motion}, and carry out ablation studies in Section \ref{sec:exp_ablation} and robot experiments in Section~\ref{sec:exp_robot}.

\subsection{Experimental Setup}
\label{sec:exp_setup}

\subsubsection{Implementation Details}  
We encode motion sequences using a temporal downsampling ratio $\alpha=4$, padding them with zeros to ensure that the length is a multiple of $\alpha$ to prevent information loss.
Our part-level motion tokenizer employs an 8-layer Grouped Residual Quantization (GRQ) architecture with a default group size $n=2$ by default, converting each one-second motion sequence into $2 \times n \times L \times \lceil T/\alpha \rceil = 128$ tokens. 
We use codebook sizes $K_w = K_f = 4096$ per part, shared across all RQ layers, with a code dimension of $d=512$.
The tokenizer is optimized with a batch size of 2048 and a learning rate of $2 \times 10^{-4}$. The loss function is weighted as $\lambda_1=0.02, \lambda_2=1.0$.
For multimodal sequence modeling, we consider three scales of \texttt{InternVL3} (1B, 8B, and 14B) as the backbone, trained on \texttt{UniHand-2.5M}.
Each training instance includes a scene image scaled as $448 \times 448$ and a corresponding hand motion sequence aligned with the image under camera coordinates. 
Hand poses are represented using MANO-D162, sampled at 15 FPS and discretized into 128 motion tokens per hand per second via our tokenizer. 
During pertaining, the vocabulary-level logit masking is conducted with probability $\mathcal{P} = 50\%$, and the token-level loss is filtered within $[15\%,95\%]$ percentile.
We use AdamW with a learning rate of $1 \times 10^{-5}$, a batch size of 128, and training on 32 NVIDIA A800-80G GPUs, jointly fine-tuning both the ViT adapter and LLM backbone.

\subsubsection{Evaluation on Hand Motion Generation}
\label{sec:eval_setup_hand_gen}

We conduct experiments on this benchmark to validate the effectiveness of our pretrained foundation VLA.
Below, we describe the datasets, benchmarking tasks, and evaluation metrics used in our study.

\noindent\textbf{Dataset Setups.}
For evaluation, we reserve 5\% of videos along with their paired hand motion and textual annotations from sources in \DataName. 
Notably, the distributions of wrist translation vary across sources, with EgoDex contributing the majority of samples.
To balance the overall distribution, we augment the other sources to expand their coverage in the translation space. 
As a result, EgoDex remains the central mode of the distribution, while the other sources form a broader yet sparser long-tail region.  
To reflect this structure, we evaluate the model on two distinct splits: the held-out EgoDex samples (referred to as the ``\textbf{head split}'') and the combined samples from TACO, HOI4D, H2O, and OakInk2 (referred to as the ``\textbf{tail split}''). 
This enables a systematic assessment of the model's ability to capture dominant motion patterns while generalizing to less frequent motion contexts in our multi-source data.

\noindent\textbf{Task Definition.}
To comprehensively evaluate our foundation VLA in grounding 2D visual cues to 3D hand motion, maintaining temporal motion coherence, and aligning vision-language-motion semantics, we design three complementary tasks.

\begin{itemize}[leftmargin=1.5em]
\item \textbf{Visual-Grounded Hand Motion Generation.}  
    The model takes as input a static scene image, a textual instruction, and a specified duration, then directly generates a semantically aligned hand motion sequence in absolute 3D coordinates that matches the image. 
    To accomplish this, the model must accurately establish an absolute spatial mapping from 2D to 3D while converting the instruction semantics into temporally coherent motion dynamics.
\item \textbf{Contextualized Hand Motion Generation.} 
    The model receives a scene image, a short segment of ground-truth hand motion as context, and a follow-up textual instruction to predict the subsequent hand motion sequence. 
    By incorporating the motion context, this task eliminates ambiguity in the initial 2D-to-3D mapping, allowing the evaluation to focus on whether the generated motion remains semantically aligned with the instruction while preserving coherence with the provided context.
\item \textbf{Hand Motion Translation.}  
    The model is prompted with a scene image and a hand motion sequence to generate a textual description of the motion. 
    This inverse task evaluates the model’s ability to interpret hand motion semantics and accurately translate 3D motion back into linguistically grounded descriptions.
\end{itemize}

These tasks together constitute a holistic benchmark for assessing vision-language-motion grounding from both generation and understanding perspectives.

\noindent\textbf{Evaluation Protocol.}
To comprehensively evaluate the quality of generated motions, we analyze three key aspects: spatial accuracy, temporal coherence, and semantic consistency. 
For hand generation tasks, we employ five primary metrics to jointly assess spatial accuracy and semantic alignment of the generated hand motions:
\begin{itemize}[leftmargin=1.5em]
\item \textbf{MPJPE (Mean Per Joint Position Error)} 
    measures overall spatial accuracy by computing the mean Euclidean distance between each generated joint and its ground-truth position in absolute 3D space. 
\item \textbf{MWTE (Mean Wrist Translation Error)} 
    evaluates global trajectory fidelity through the mean Euclidean distance between predicted and ground-truth wrist positions across the sequence.
\item \textbf{PA-MPJPE (Procrustes Aligned MPJPE)}  
    isolates relative pose accuracy by aligning predicted joints to the ground truth via rigid transformation (including scaling, rotation, and translation).
\item \textbf{M2T R@3 (Motion-to-Text Retrieval Top-3 Accuracy)}   
    assesses semantic alignment by embedding generated motion into a shared representation space and retrieving the top-3 matching textual descriptions using a dataset-specific text-motion retrieval model (TMR~\cite{petrovich23tmr}). 
\item \textbf{FID (Fréchet Inception Distance)} quantifies the distribution similarity by comparing the generated and real motion embeddings in the dataset-specific latent space of the TMR model, measuring how well synthesized motions match the true data distribution.
\end{itemize}
For the hand motion translation task, we evaluate using a single retrieval metric: 
\textbf{T2M R@3 (Text-to-Motion Retrieval Top-3 Accuracy)}, which reports how effectively the generated textual description retrieves its corresponding motion sequence from a database. 
This metric directly verifies whether the model’s text output accurately captures the semantic content of the motion.
We complement this with the valid generation rate for free-form generation, which reflects how consistently the model produces motion sequences adhering to the required structural format. 
Together, these metrics offer a holistic assessment of the model’s capabilities to generate physically plausible, temporally coherent hand motions that remain faithful to the given instructions.
 
\subsubsection{Evaluation on Dexterous Manipulation}
As intended, our \texttt{Being-H0} is able to execute dexterous manipulation tasks after physical instruction tuning.
To assess the transferability of hand motion knowledge acquired from \DataName to downstream tasks, we conduct real-robot experiments for quantitative evaluation.

\begin{wrapfigure}{r}{.36\textwidth}
    \centering
    \vspace{-8pt}      
    \includegraphics[width=.3\textwidth]{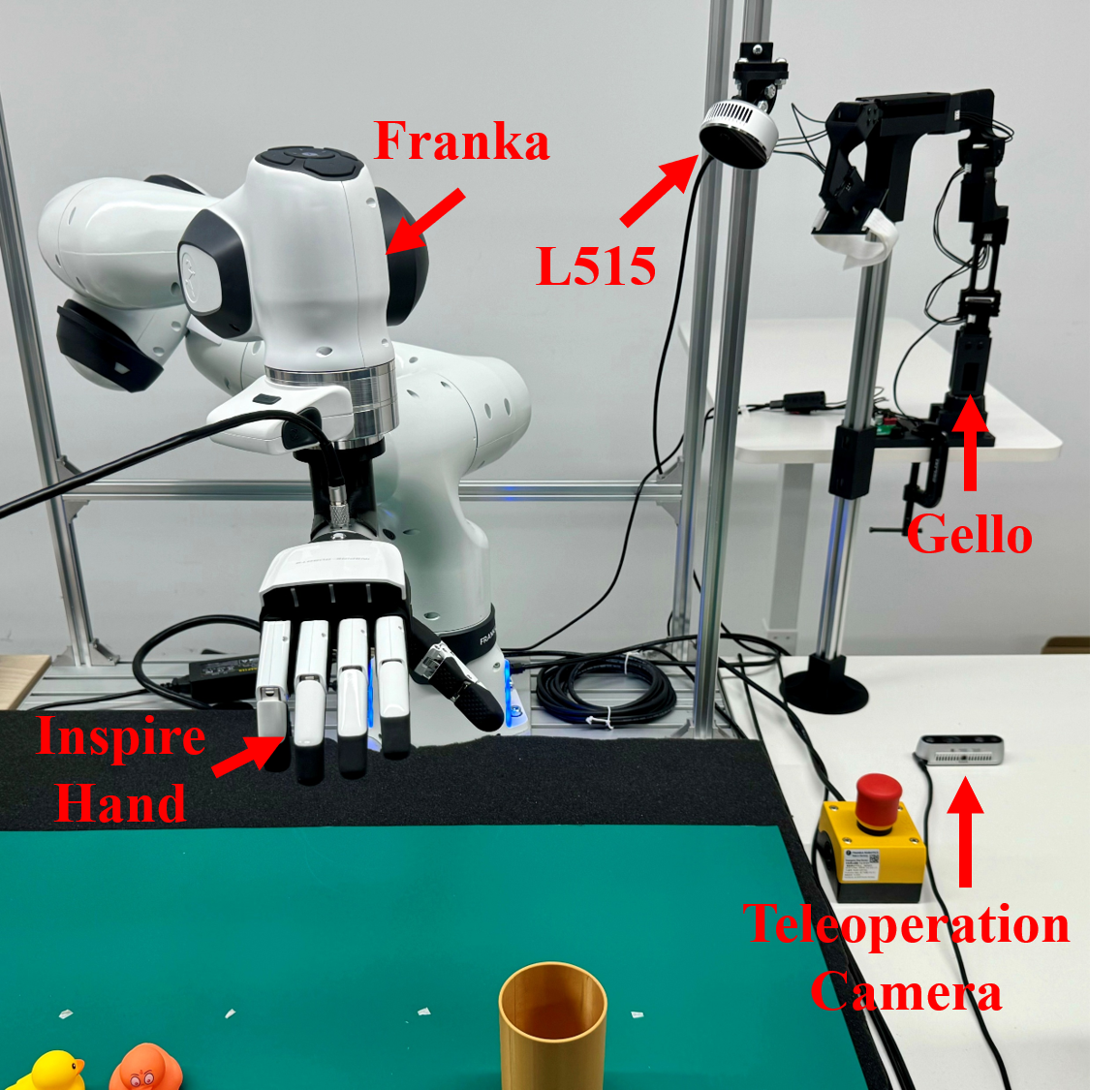}
    \caption{The hardware system.}
    \label{fig:hardware}
    \vspace{-10pt} 
\end{wrapfigure}

\noindent\textbf{Robot System.}
 Our experiments are conducted using a hardware setup including a 7-DoF Franka Research 3 arm, a 6-DoF Inspire hand, and a RealSense L515 camera for RGB streaming. 
To collect demonstrations for imitation learning, we introduce an improved teleoperation system that integrates the Gello exoskeleton~\cite{wu2024gello} for arm control with a RealSense D435i camera for hand pose estimation and retargeting~\cite{qin2023anyteleop,yuan2025being}, as shown in Figure~\ref{fig:hardware}.
The evaluation covers a diverse set of manipulation tasks, including grasp-and-place skills and interactions with both articulated and deformable objects. 
For each task, we collect 50–100 teleoperation trajectories to post-train our foundation VLA. 
At each step, the policy takes as input an egocentric RGB image and robot proprioception to generate an action chunk consisting of end-effector poses and dexterous hand joint positions.

\noindent\textbf{Evaluation Tasks.}
We design a suite of real-world manipulation tasks that assess fundamental skills while also challenging its generalization and precision in complex scenarios.

\begin{itemize}[leftmargin=1.5em]
\item \textbf{Grasping and Placing (\texttt{Pick-Place-Toy}).} 
    The task includes three sub-scenarios. 
    The \textbf{Seen} scenario assesses basic capabilities by requiring the robot to pick up a toy seen during post-training. 
    To test object generalization, the \textbf{Unseen} scenario introduces a novel toy with visual properties (e.g., color).
    Finally, the \textbf{Clutter} scenario evaluates advanced scene understanding by challenging the robot to identify and retrieve the target toy from among multiple distractors in a cluttered environment. 
\item \textbf{Articulated Object Manipulation (\texttt{Close-Toolbox}, \texttt{Close-Lid}).} 
    The robot must perform delicate closure actions on a toolbox and a cup lid. 
    These tasks rigorously assess the model's capability for accurate end-effector positioning and orientation, as well as stable interaction with object mechanics.
\item \textbf{Deformable Object Manipulation (\texttt{Unfold-Clothes}).} 
    The robot is required to unfold a piece of cloth, testing the model's ability to perform multi-finger fine-grained manipulation and dynamically alter the state of non-rigid objects.
\item \textbf{Precise Motion Control (\texttt{Pour-Cup}).} 
    In this task, the robot is asked to pour liquid from one cup to another, requiring the generation of smooth, stable motion trajectories to maintain temporal action coherence throughout the actions.
\end{itemize}

\noindent\textbf{Evaluation Protocol.}
We evaluate dexterous manipulation performance using success rate as our primary metric, where task success follows strict binary criteria (e.g., complete lid closure, accurate toy placement). 
Each task undergoes 20 randomized trials with varied initial object positions to ensure statistical robustness.
Our \texttt{Being-H0} is benchmarked against two baselines: GR00T N1.5~\cite{bjorck2025gr00t} and InternVL3~\cite{zhu2025internvl3}.
We select GR00T because it is the only large-scale VLA model pre-trained on egocentric human videos with a focus on dexterous manipulation. 
This stands in contrast to most other models, such as OpenVLA~\cite{kim2024openvla}, which are designed for grippers.
Meanwhile, InternVL3 shares Being-H0's architecture and scale, but it lacks hand motion pretraining and physical alignment. 
All models receive identical post-training on the same teleoperation datasets to assess the benefits of pre-training on egocentric hand datasets.
To better understand model capabilities and failure modes, we conduct qualitative analysis to perform model behavior analysis across three dimensions:  motion precision (e.g., \texttt{Close-Lid}), semantic understanding of instructions (e.g., \texttt{Pick the white duck}), and robustness in complex tasks (e.g., \texttt{Unfold-Clothes}).

\subsection{Comparisons on Hand Motion Generation and Translation}
\label{sec:comp_hand_gen_translate}

\begin{wraptable}{r}{0.45\textwidth}
\vspace{-1em}
\centering
\caption{Comparison results on two tasks: visual-grounded motion generation and hand motion translation, where we use valid rate (\%) and T2M R@3 (\%) as the metrics.}
\label{tab:valid_rate}
\vspace{-1mm}
\setlength{\tabcolsep}{4pt}
\scalebox{.9}{
\begin{tabular}{m{2.2cm} >{\centering\arraybackslash}m{2cm} *{2}{>{\centering\arraybackslash}m{1cm}}}
\toprule
\multirow{2}{*}{\textbf{Model}} & \multirow{2}{*}{\textbf{Valid Rate} $\uparrow$} & \multicolumn{2}{c}{\textbf{T2M R@3} $\uparrow$}   \\
\cmidrule(lr){3-4}
 &  & head & tail \\
\midrule
\addlinespace
    \rowcolor{BlockC}
ground truth   & - & 33.5 & 42.7 \\
\addlinespace
    \rowcolor{BlockA!30}
\ModelName-1B  & 64.8 & 12.5 & 14.3  \\
\rowcolor{BlockA!30}
\ModelName-8B  & 99.8 & 18.4 & 19.7 \\
\rowcolor{BlockA!30}
\ModelName-14B & 100.0 & 19.0 & 22.1 \\
\bottomrule
\end{tabular}}
\end{wraptable}

To begin with, we set the prediction horizon as one second for all experiments in this section.
Our evaluation first examines the model's capability to produce motion sequences that adhere to the required structural format (\texttt{<MOT>}$......$\texttt{</MOT>}).
As shown in Table~\ref{tab:valid_rate}, experiments on visual-grounded motion generation tasks are carried out based on free-format mode (Section~\ref{sec:next_token_pred}), revealing substantial differences in valid generation rates across model scales.
While \ModelName-1B achieves only modest success in preserving motion block structure, the more capable \ModelName-8B and \ModelName-14B respectively reach 99.8\% and 100\% validity, demonstrating that increased model scale significantly improves the learning of structural motion format.

Table~\ref{tab:valid_rate} also evaluates models' motion understanding capabilities through the hand motion translation task.  
We employ T2M R@3 metrics on both head and tail splits to measure how accurately the generated text descriptions retrieve corresponding motion sequences, verifying the semantic fidelity of the model's text outputs.
Results show that larger models consistently achieve higher T2M R@3 scores, confirming their stronger bidirectional alignment between motion and language modalities.

We report principal quantitative results on the visual-grounded and contextualized hand motion generation tasks in Table~\ref{table:dex_traj_gen}.
To provide a competitive and generalizable baseline, we adopt GR00T N1.5~\cite{bjorck2025gr00t} as the baseline. Originally trained on multimodal robotics and human datasets, we adapt it to our setting following a similar procedure as in MEgoHand~\cite{zhou2025megohand}. Specifically, we redefine its action representation as a dual-hand motion sequence. And when a single-hand motion sequence is required, the other hand is padded with zeros. The initial hand pose is used as the state for the action head in the contextualized hand motion generation, while a fixed state is employed in the visual-grounded hand motion generation.
To maintain consistent evaluation across different model sizes and baselines, we adopt the block-formatted mode (Section~\ref{sec:next_token_pred}), enforcing each generated sequence to be decoded into the correct format.
Our analysis reveals three key findings.
First, larger models consistently demonstrate superior performance with lower MPJPE, MWTE, and PA-MPJPE scores, reflecting enhanced spatial grounding and more plausible pose generation.
Second, they achieve significantly better results on M2T R@3 and FID metrics, indicating stronger semantic consistency between generated motions and input instructions. 
Notably, the performance advantage is particularly pronounced on the tail split, suggesting that model scaling substantially improves generalization capabilities across diverse motion distributions.

\begin{table}[t!]
\centering
\caption{
Comparison results of visual-grounded and contextualized hand motion generation tasks on both head and tail splits, whose samples come from $\{$EgoDex$\}$ and $\{$TACO, HOI4D, H2O, OakInk2$\}$, respectively. MPJPE, MWTE, and PA-MPJPE are reported in centimeters (cm), while the M2T R@3 is reported in percentage (\%), which stays consistent in the rest of the article.
}
\label{table:dex_traj_gen}
\vspace{-2mm}
\setlength{\tabcolsep}{2pt}{
\scalebox{0.9}{
\begin{tabular}{m{3cm} *{10}{>{\centering\arraybackslash}m{1.2cm}}}
\toprule
\multirow{2}{*}{\textbf{Model}}
& \multicolumn{2}{c}{\textbf{MPJPE} $\downarrow$} 
& \multicolumn{2}{c}{\textbf{MWTE} $\downarrow$} 
& \multicolumn{2}{c}{\textbf{PA-MPJPE} $\downarrow$}
& \multicolumn{2}{c}{\textbf{M2T R@3} $\uparrow$}
& \multicolumn{2}{c}{\textbf{FID} $\downarrow$}
\\
\cmidrule(lr){2-3}
\cmidrule(lr){4-5}
\cmidrule(lr){6-7}
\cmidrule(lr){8-9}
\cmidrule(lr){10-11}
& head & tail
& head & tail
& head & tail
& head & tail
& head & tail
\\
 \midrule \addlinespace 
\rowcolor{BlockA!30}
\multicolumn{11}{l}{\textbf{\# Visual-Grounded Hand Motion Generation}} \\
\rowcolor{BlockA!30}
GR00T N1.5~\cite{bjorck2025gr00t} & 9.82 & 15.35 & 8.51 & 11.20 & 1.33 & 1.41 & 13.1 & 14.8 & 11.7 & 14.4  \\
\rowcolor{BlockA!30}
\ModelName-1B   & 9.71 & 17.21 & 8.25 & 12.04 & 1.50 & 1.55 & 12.1 & 15.3 & 12.2 & 13.1  \\
\rowcolor{BlockA!30}
\ModelName-8B   & 7.20 & 9.02 & 5.69 & 8.11 & 1.09 & 1.32 & 15.9 & 18.7 & 11.5 & 13.4  \\
\rowcolor{BlockA!30}
\ModelName-14B  & 6.87 & 8.11 & 5.19 & 7.41 & 1.03 & 1.20 & 17.2 & 20.5 & 10.3 & 11.8  \\
\addlinespace
\rowcolor{BlockB!30}
\multicolumn{11}{l}{\textbf{\# Contextualized Hand Motion Generation}} \\
\rowcolor{BlockB!30}
GR00T N1.5~\cite{bjorck2025gr00t} & 7.14 & 8.55 & 6.65 & 7.93 & 1.11 & 1.25 & 16.1 & 20.5 &  11.2 &  13.3  \\
\rowcolor{BlockB!30}
\ModelName-1B  & 8.73 & 11.34 & 7.88 & 10.67 & 1.17 & 1.38 & 15.4 & 20.6 & 14.7 & 15.6  \\
\rowcolor{BlockB!30}
\ModelName-8B  & 6.67 &  7.98 & 5.03 &  6.93 & 0.90 & 1.03 & 19.7 & 21.4 & 10.1 &  11.7 \\
\rowcolor{BlockB!30}
\ModelName-14B & 6.21 & 7.33 & 4.89 & 6.52 & 0.92 & 1.04 & 20.1 & 23.5 & 9.8 & 10.1 \\

\bottomrule
\end{tabular}
}}
\end{table}

\subsection{Comparisons on Long-range Motion Generation}
\label{sec:exp_long_motion}

We evaluate the long-term motion generation capabilities of different \ModelName variants through systematic comparisons, as presented in Table~\ref{tab:longseq_results}. 
Longer motion sequences inherently suffer from accumulated prediction errors that may cause trajectory drift or degraded pose quality. 
To mitigate this, we implement the soft-formatted mode (Section~\ref{sec:next_token_pred}) during inference, which constrains generated motions within plausible ranges relative to ground-truth distributions.
For comprehensive analysis, we categorize results into \textbf{short-term} (2–5 seconds) and \textbf{long-term} (6–10 seconds) ranges, allowing precise examination of quality degradation with increasing sequence length.  
Our evaluation employs MPJPE, MWTE, and PA-MPJPE as the core metrics for spatial accuracy and trajectory stability. 
The results demonstrate that generation quality naturally deteriorates with longer sequences, evidenced by elevated MPJPE and MWTE values in the long-term range due to error accumulation.
However, larger models maintain more stable spatial accuracy under soft constraints, as they benefit from partial ground-truth context to anchor the trajectory.

\begin{table}[ht]
\centering
\caption{
Comparison results of hand motion generation tasks upon long-range sequences.
We adopt the soft-formatted mode (Section~\ref{sec:next_token_pred}) and report short-term (2–5s) and long-term (6–10s) results.
}
\label{tab:longseq_results}
\vspace{-2mm}
\setlength{\tabcolsep}{1.5pt}
\scalebox{0.9}{
\begin{tabular}{m{2.2cm} *{12}{>{\centering\arraybackslash}m{1.2cm}}}
\toprule
\multirow{3}{*}{\textbf{Model}} 
& \multicolumn{6}{c}{\textbf{Short-Term (2–5s)}} 
& \multicolumn{6}{c}{\textbf{Long-Term (6–10s)}} \\
\cmidrule(lr){2-7} \cmidrule(lr){8-13}
& \multicolumn{2}{c}{MPJPE  $\downarrow$} & \multicolumn{2}{c}{MWTE  $\downarrow$} & \multicolumn{2}{c}{PA-MPJPE  $\downarrow$}
& \multicolumn{2}{c}{MPJPE  $\downarrow$} & \multicolumn{2}{c}{MWTE  $\downarrow$} & \multicolumn{2}{c}{PA-MPJPE  $\downarrow$} \\
\cmidrule(lr){2-3} \cmidrule(lr){4-5} \cmidrule(lr){6-7}
\cmidrule(lr){8-9} \cmidrule(lr){10-11} \cmidrule(lr){12-13}
& head & tail & head & tail & head & tail
& head & tail & head & tail & head & tail \\
\midrule
\addlinespace 
\rowcolor{BlockA!30}
\multicolumn{13}{l}{\textbf{\# Visual-Grounded Hand Motion Generation}} \\
\rowcolor{BlockA!30}
\ModelName-1B  & 8.97 & 9.96 & 7.01 & 8.75 & 1.43 & 1.67 & 9.12 & 11.24 & 7.13 & 9.91 & 1.60 & 1.81 \\ 
\rowcolor{BlockA!30}
\ModelName-8B   & 7.55 & 8.45 & 5.78 & 7.51 & 1.10 & 1.30 & 8.21 & 9.98 & 6.12 & 8.34 & 1.22 & 1.36 \\
\rowcolor{BlockA!30}
\ModelName-14B  & 7.43 & 8.39 & 5.65 & 7.39 & 1.11 & 1.28 & 7.98 & 9.72 & 5.88 & 8.01 & 1.18 & 1.32 \\
\addlinespace
\rowcolor{BlockB!30}
\multicolumn{13}{l}{ \textbf{\# Contextualized Hand Motion Generation}} \\
\rowcolor{BlockB!30}
\ModelName-1B   & 8.44 & 9.52 & 6.71 & 7.99 & 1.20 & 1.45 & 9.01 & 10.98 & 6.98 & 8.75 & 1.35 & 1.50 \\
\rowcolor{BlockB!30}
\ModelName-8B  & 7.67 & 8.20 & 5.81 & 7.13 & 1.01 & 1.22 & 8.23 & 9.67 & 6.23 & 7.83 & 1.14 & 1.27 \\ 
\rowcolor{BlockB!30}
\ModelName-14B  & 7.39 & 8.51 & 5.77 & 7.21 & 1.05 & 1.25 & 8.01 & 9.45 & 6.02 & 7.67 & 1.18 & 1.30 \\
\bottomrule
\end{tabular}
}
\end{table}

\subsection{Ablation Study}
\label{sec:exp_ablation}

We perform systematic ablation studies to examine the impact of architectural decisions and training protocols on the model's ability to generate spatially plausible and semantically coherent hand motions.
These investigations focus on two critical tasks: (1) hand reconstruction for motion tokenizer, which evaluates the model's capacity to accurately reproduce input motion sequences, and (2) visual-grounded motion generation for foundation VLA.

\subsubsection{Optimal Practice for Hand Motion Tokenization}

\noindent\textbf{Part-Level vs. Others.}
Our comparative evaluation in Table~\ref{tab:ablate-motion-feature} demonstrates the advantages of part-level tokenization over approaches that quantize the entire hand uniformly.
We benchmark against two GRQ variants: a 4-group configuration ($n=4$) and a 16-layer residual quantization architecture ($L=16$), both employing an identical codebook size ($K=K_w+K_f=8192$)
To maintain experimental rigor, all variants adhere to a fixed token count per second and consistent codebook dimensionality.
All evaluations are conducted on \DataName's held-out test set, exclusively containing motion sequences absent from training data.
The part-level tokenizer achieves superior performance, demonstrating the effectiveness of separately tokenizing the wrist and fingers for modeling hand motions.

\noindent\textbf{MANO-D162 vs. Others.}
Building on our tokenization framework (Section~\ref{sec:hand-motion-tokenization}), we analyze different motion features through GRQ reconstruction.
The axis-angle formulation yields superior overall reconstruction accuracy (MANO-D51 vs. MANO-D99; MANO-D114 vs. MANO-D162), while 6D rotation achieves better PA-MPJPE scores, indicating its potential advantage at modeling the finger actions.
As demonstrated in Table~\ref{tab:abl-data-cfg}, the 6D rotation-based MANO-D162 feature proves most effective for \texttt{Being-H0} training.
We further observe that incorporating auxiliary joint positions ($j$) enhances performance, whereas modeling hand shape parameters ($\beta$) degrades results. 
Consequently, we maintain constant shape parameters from each sequence's initial frame, allowing the tokenizer to focus exclusively on motion dynamics.

\begin{table*}[h]
\centering
\caption{Performance of different motion tokenization practices on the hand reconstruction task, including motion features and part-level tokenizing. Results are reported in centimeters (cm).}
\label{tab:ablate-motion-feature}
\vspace{-2mm}
\setlength{\tabcolsep}{6pt}
\scalebox{0.9}{
\begin{tabular}{lcccccc}
\toprule
\multirow{2}{*}{\textbf{Feature}} 
& \multicolumn{2}{c}{\textbf{Part-Level}} & \multicolumn{2}{c}{\textbf{4-Groups}} 
& \multicolumn{2}{c}{\textbf{16-Layers}}  \\
\cmidrule(r){2-3} \cmidrule(r){4-5} \cmidrule(r){6-7} 
& MPJPE~$\downarrow$ & PA-MPJPE~$\downarrow$ & MPJPE~$\downarrow$ & PA-MPJPE~$\downarrow$ & MPJPE~$\downarrow$ & PA-MPJPE~$\downarrow$\\
\midrule
\rowcolor{BlockC}
MANO-D51 & 0.556 & 0.209 & 1.165 & 0.184 & 1.466 & 0.243 \\
\rowcolor{BlockC}
MANO-D99  & 0.584 & 0.149 & 1.093 & 0.148 & 1.510 & 0.170 \\
\addlinespace
\rowcolor{BlockA!30}
\multicolumn{7}{l}{\textbf{\# with shape parameters $\beta$}} \\
\rowcolor{BlockA!30}
MANO-D109 & 0.592 & 0.160 & 1.107 & 0.140 & 1.602 & 0.201 \\
\addlinespace
\rowcolor{BlockB!30}
\multicolumn{7}{l}{\textbf{\# with auxiliary joint positions $j$}} \\
\rowcolor{BlockB!30}
MANO-D114 & \textbf{0.523} & 0.167 & 0.810 & 0.202 & 0.996 & 0.253 \\
\rowcolor{BlockB!30}
MANO-D162 & 0.573 & \textbf{0.129} & 0.704 & 0.138 & 1.054 & 0.226 \\
\bottomrule
\end{tabular}
}
\end{table*} 

\noindent\textbf{Impact on Hand Motion Generation.} We adopt the MANO-D162 + Part-Level configuration as our default motion tokenizer for training \ModelName. 
To verify its effectiveness, we additionally train models using three alternative configurations that show lower GRQ reconstruction errors, namely MANO-D114 + 4-Groups, MANO-D162 + 4-Groups, and MANO-D114 + Part-Level.
As summarized in Table~\ref{tab:abl-data-cfg}, despite MANO-D162 exhibiting slightly higher MPJPE in reconstruction (Table~\ref{tab:ablate-motion-feature}), it consistently outperforms others in hand motion generation tasks.
We attribute this to the 6D rotation representation and the part-level decomposition, which likely facilitate better temporal modeling and autoregressive generation of fine-grained hand motions.

\begin{table}[ht]
  \centering
  \caption{Ablation on different motion tokenizer variants and data recipes for visual-grounded hand motion generation. Here ``Trans'' and ``Context'' denote the task types of motion translation, while ``Balance'' represents view-invariant motion distribution balancing.}
  \label{tab:abl-data-cfg}
\vspace{-2mm}
\setlength{\tabcolsep}{2pt}
\scalebox{0.9}{
  \begin{tabular}{m{4cm} *{10}{>{\centering\arraybackslash}m{1.2cm}}}
    \toprule
      \multirow{2}{*}{\textbf{Variants}} 
& \multicolumn{2}{c}{\textbf{MPJPE} $\downarrow$} 
& \multicolumn{2}{c}{\textbf{MWTE} $\downarrow$} 
& \multicolumn{2}{c}{\textbf{PA-MPJPE} $\downarrow$}
& \multicolumn{2}{c}{\textbf{M2T R@3} $\uparrow$}
& \multicolumn{2}{c}{\textbf{FID} $\downarrow$}
\\
\cmidrule(lr){2-3} 
\cmidrule(lr){4-5} 
\cmidrule(lr){6-7} 
\cmidrule(lr){8-9} 
\cmidrule(lr){10-11}
& head & tail 
& head & tail 
& head & tail 
& head & tail 
& head & tail \\
    \midrule
    \rowcolor{BlockC}
    \multicolumn{11}{l}{\textcolor{black}{\textbf{\# Base}}}\\
    \rowcolor{BlockC}
     \ModelName‐8B  & \textbf{7.20} & \textbf{9.02}  & 5.69 & \textbf{8.11 } & \textbf{1.09} & \textbf{1.32} & \textbf{15.9} & \textbf{18.7} & 11.5 & \textbf{13.4} \\
    \addlinespace
    \rowcolor{BlockA!30}
    \multicolumn{11}{l}{\textcolor{black}{\textbf{\# Tokenizer Variants}}}
    \\
    \rowcolor{BlockA!30}
    MANO-D114 + 4‐Groups    & 8.31 & 10.35 & 6.52 & 9.11  & 1.14 & 1.35 & 13.1 & 14.7 & 13.7 & 15.6 \\
       \rowcolor{BlockA!30}
       MANO‐D162 + 4‐Groups    & 7.98 & 9.71  & 5.58 & 8.98  & 1.09 & 1.38 & 15.4 & 17.1 & 11.9 & 14.3 \\
       \rowcolor{BlockA!30}
       MANO‐D114 + Part‐Level  & 7.74 & 9.92  & 6.11 & 8.83  & 1.16 & 1.41 & 12.3 & 16.1 & 13.1 & 12.3 \\
    \addlinespace
    \rowcolor{BlockB!30}
    \multicolumn{11}{l}{\textbf{\# Data Recipe}}
    \\
    \rowcolor{BlockB!30}
     w/o Translate            & 7.22 & 9.11  & \textbf{5.51} & 8.12  & 1.27 & 1.46 & 13.2 & 11.6 & 13.1 & 15.7 \\
     \rowcolor{BlockB!30}
       w/o Context       & 8.01 & 10.97 & 6.34 & 9.03  & 1.11 & 1.52 & 13.7 & 15.2 & \textbf{11.1} & 14.9 \\
       \rowcolor{BlockB!30}
       w/o Balance                  & 8.54 & 12.13 & 7.74 & 10.04 & 1.24 & 1.57 & 11.3 & 10.3 & 15.8 & 16.3 \\
    \bottomrule
  \end{tabular}}
\end{table}

\subsubsection{Impact of Data Configuration}

To optimize the performance of \ModelName, we construct \DataName to include not only standard hand motion generation data but also diverse instruction types and a view-invariant balanced distribution (Section~\ref{sec:data_curation}). 
To understand how different components of our \DataName influence the performance of \ModelName, we systematically examine the effects of instructional data types, distribution balancing, and data scale.

\begin{wrapfigure}{r}{0.48\textwidth}
    \centering
    \vspace{-14pt}      
    \includegraphics[width=0.46\textwidth]{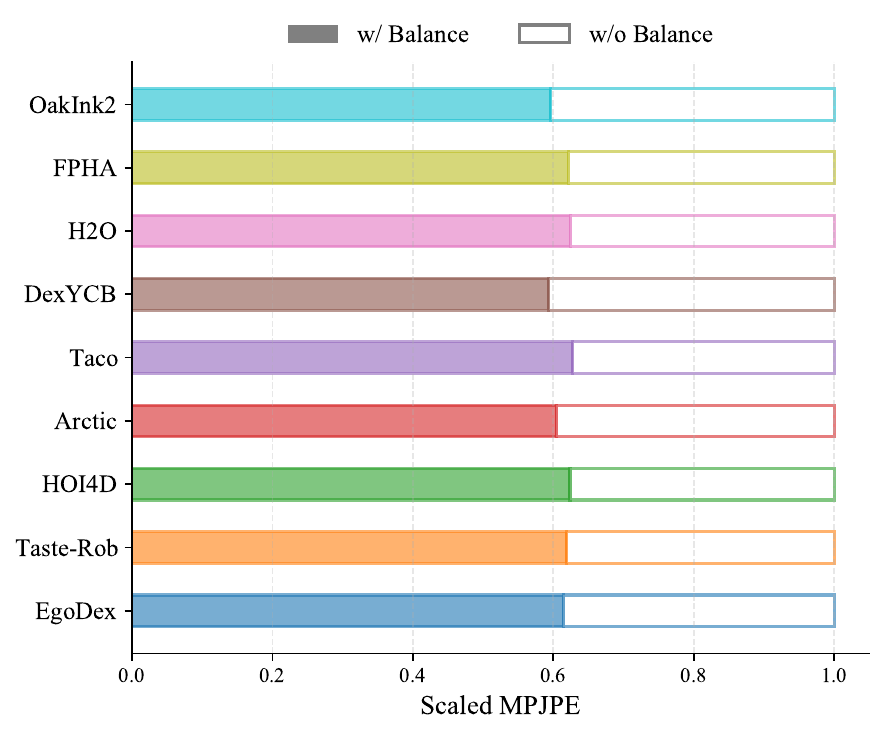}
    \vspace{-8pt}
    \caption{Ablation of view-invariant motion distribution balancing (``Balance'') on motion reconstruction.}
    \label{fig:ablate-view-aug}
   \vspace{-10pt} 
\end{wrapfigure}

\noindent\textbf{How does view-invariant motion distribution balancing improve generalization?}
As described in Section~\ref{sec:next_token_pred}, our data balancing strategy aims to equalize motion-view coverage across data sources. This is achieved by augmenting hand poses while preserving consistent weak-perspective projection. 
We assess its effectiveness from two perspectives. On the one hand, we evaluate its influence on the motion tokenizer learning. Specifically, we compare GRQ reconstruction performance with and without this strategy on the held-out test set of \DataName, comprising unseen motion sequences during training with augmented poses.
As shown in Figure~\ref{fig:ablate-view-aug}, applying balancing significantly reduces GRQ reconstruction errors across all datasets, even for sources without explicit augmentation (e.g., EgoDex and Taste-Rob). This indicates that the tokenizer benefits from more evenly distributed motion-view coverage, leading to more precise encoding of both global wrist motion and fine-grained finger articulation. 
On the other hand,  we also evaluate its impact on the learning of foundation VLA. We compare \ModelName with a variant trained on the dataset without data balancing. As shown in Table~\ref{tab:abl-data-cfg}, removing this strategy results in substantial performance degradation on the tail split. Without balancing, the model tends to overfit to dominant camera configurations, limiting its generalization to underrepresented motion-view combinations. These results underscore the dual role of view-invariant balancing in improving both the tokenizer’s representational robustness and the motion model’s ability to generate accurate and semantically grounded motions across diverse perspectives.

\noindent\textbf{Do auxiliary supervision tasks benefit the visual-grounded hand motion generation?} Beyond the basic instructional motion generation data, our dataset incorporates two additional types of supervision data: hand motion translation and contextual motion prediction. These correspond respectively to the evaluation tasks of \textit{Hand Motion Translation} and \textit{Contextualized Hand Motion Generation}. In addition to enabling these capabilities, we perform ablation experiments to assess how these data types influence performance on the core task of \textit{Visual-grounded Hand Motion Generation}. As shown in Table~\ref{tab:abl-data-cfg}, removing translation supervision yields only marginal changes in global wrist metrics (MPJPE, MWTE), but leads to clear degradation in PA-MPJPE, M2T R@3, and FID. This suggests that translation data benefits in capturing fine-grained hand articulation aligned with semantic goals. In contrast, removing contextual motion prediction leads to uniform drops across all metrics. This highlights its central role in generating temporally coherent and context-aware motion. These findings indicate that auxiliary supervision not only enables task-specific capabilities but also strengthens the core motion generation quality through richer semantic and temporal grounding.

\begin{figure}[t]
    \centering    
    \includegraphics[width=0.75\textwidth]{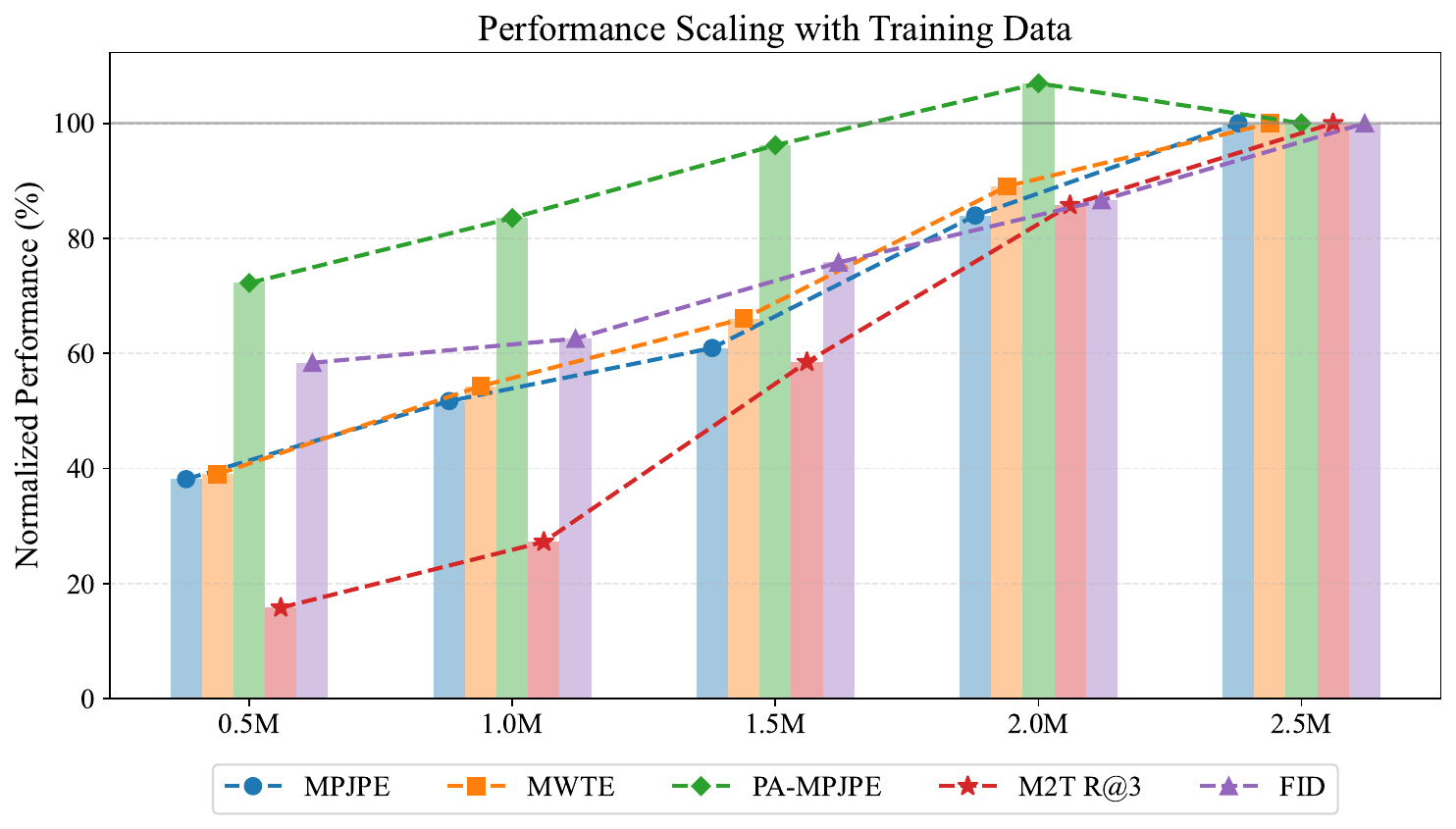}
    \caption{The performance of \ModelName-8B alongside the increasing training data scale for visual-grounded hand motion generation. All metrics are normalized to make the last checkpoint (2.5M training samples) represent 100\%, with higher values indicating better performance. The metrics include MPJPE, MWTE, PA-MPJPE, M2T R@3, and FID and are averaged on the core split and the tail split. }
    \label{fig:abl-data-scale}
\end{figure}

\noindent\textbf{Does model performance benefit from increased training scale?}
We further investigate how performance scales with the number of training samples. As shown in Figure~\ref{fig:abl-data-scale}, model performance improves steadily with increasing training size up to 2.5M samples, demonstrating the value of scaling diverse motion-language data. Notably, while PA-MPJPE slightly drops at the largest scale, semantic alignment metrics such as M2T R@3 continue to rise. One possible explanation is that larger data volumes increase diversity in task-object combinations and motion semantics, encouraging the model to prioritize semantic plausibility over exact replication of finger pose detail. This trade-off reflects the model’s growing emphasis on functional and contextual correctness as training data becomes more abundant.

\begin{figure*}[t]
\centering
\includegraphics[width=1\textwidth]{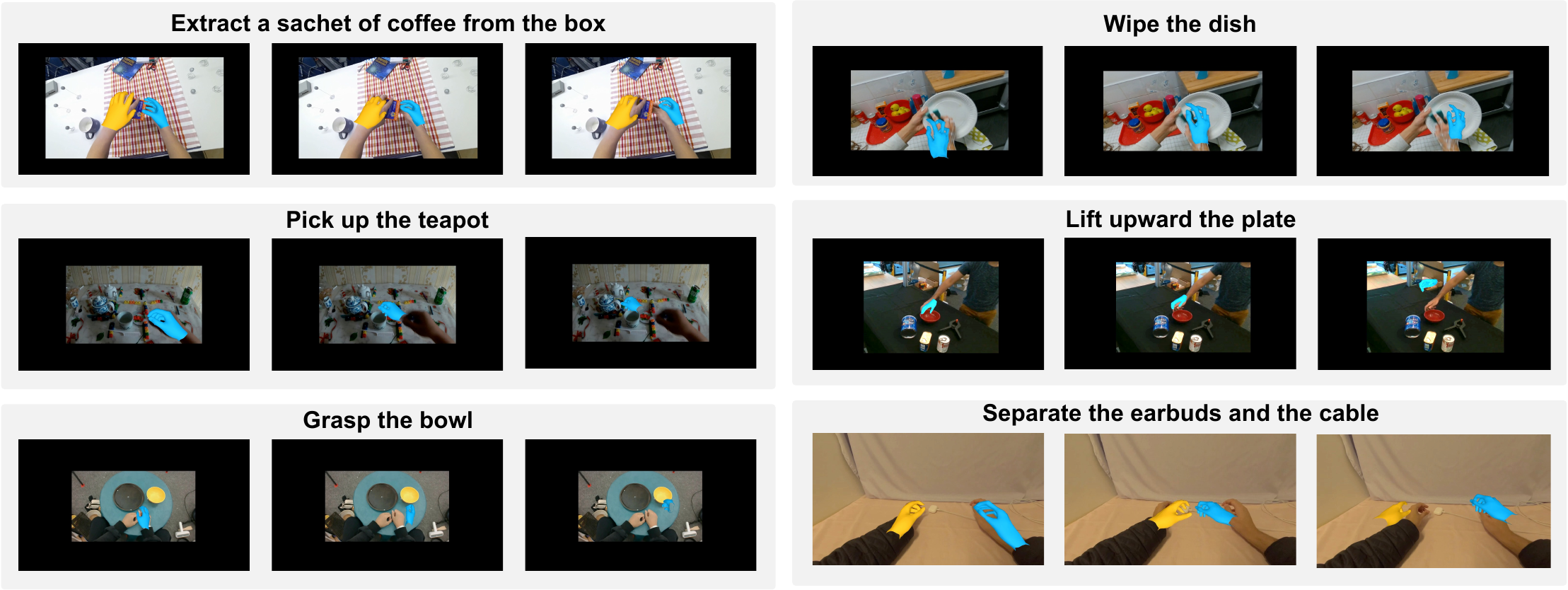}
\caption{Hand Motion Generation Samples from \ModelName-8B across various tasks, scenes, and viewpoints. A simplified task instruction is given for each block. The three frames illustrate the generated hand motion over time, rendered in the first-frame camera coordinate system and overlaid on the RGB image. Black padding around each image is introduced to enforce consistent weak-perspective projection. }
\label{fig:gen_samples} 
\end{figure*}

\begin{figure*}[ht]
\centering
\includegraphics[width=1\textwidth, trim={0cm, 8cm, 9.5cm, 0cm}, clip]{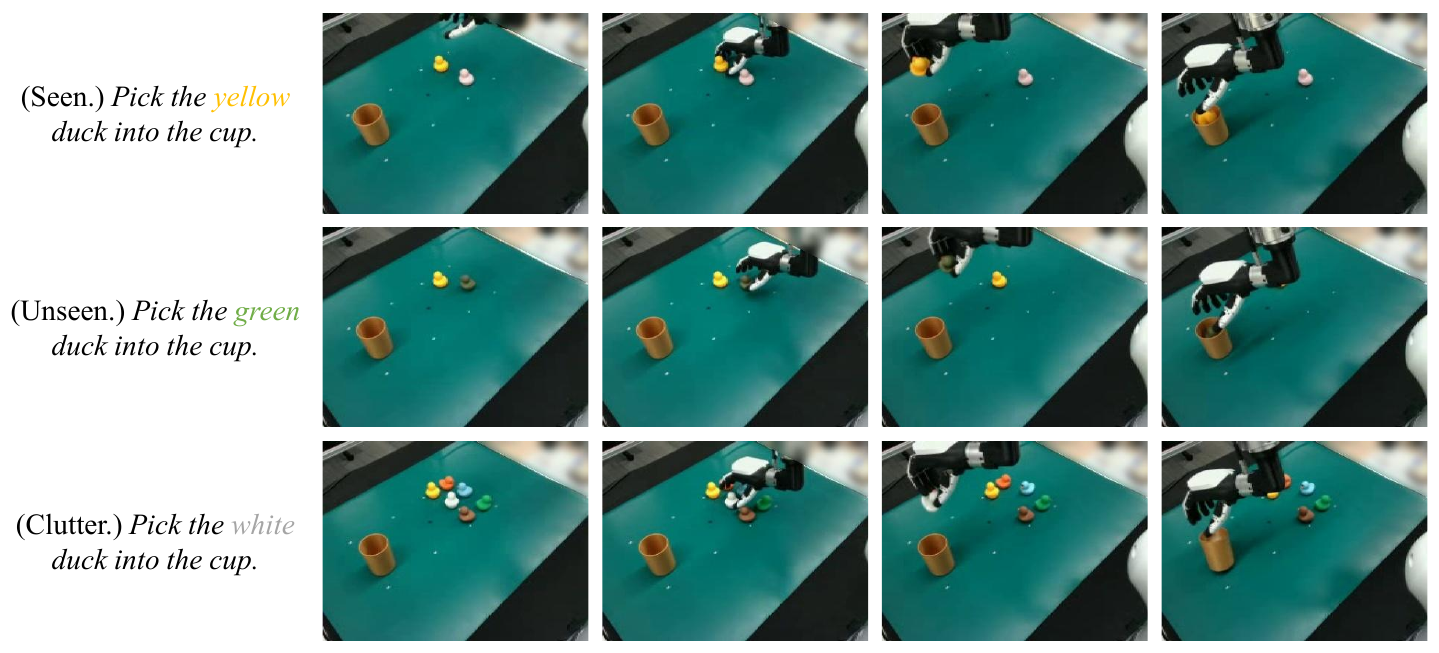}
\caption{\texttt{Being-H0} performing Pick-Place-Toy on seen objects, unseen objects, and in cluttered scenes.}
\label{fig:real-duck} 
\end{figure*}

\begin{figure*}[ht]
\centering
\includegraphics[width=1\textwidth, trim={0cm, 3.5cm, 2.5cm, 0cm}, clip]{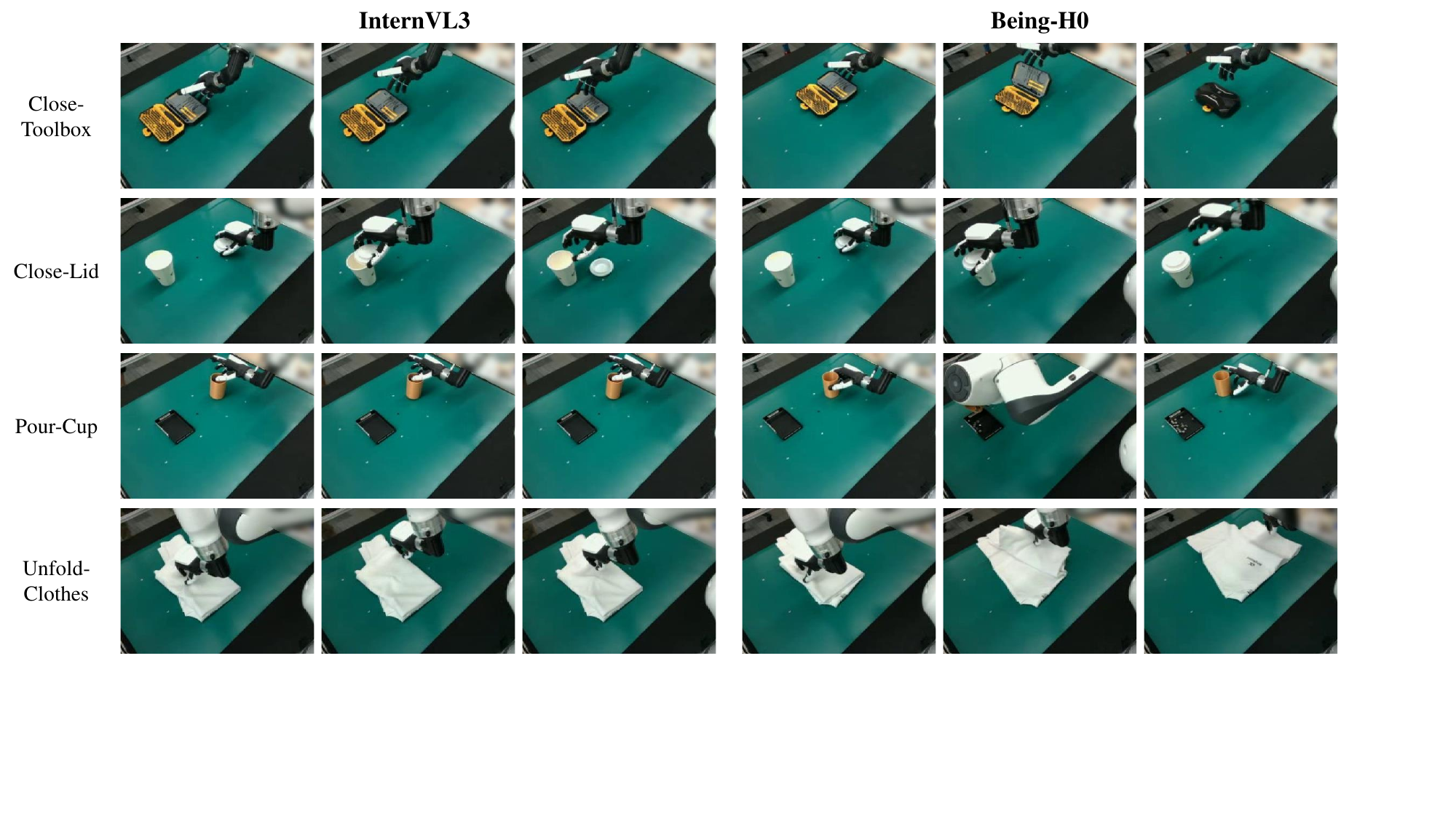}
\caption{Qualitative comparisons of \texttt{Being-H0} and the baseline (InternVL3).}
\label{fig:real-compare} 
\end{figure*}

\subsection{Hand Motion Generation Samples}
To qualitatively demonstrate the capability of \ModelName in generating physically plausible hand motions, we present representative samples in Figure~\ref{fig:gen_samples}. Each row corresponds to a different task instruction, spanning diverse scenes and camera viewpoints. Each block shows a sequence of generated hand motions rendered over the first video frame in the corresponding camera coordinate system. The rendered hands are color-coded (yellow and blue) to denote left and right hands, respectively.

To standardize visualization under a unified weak-perspective projection, we apply a transformation strategy to the frames, resulting in black borders around the images. The effective image region (excluding the black padding) thus reflects differences in hand-object interaction depth, as closer interactions appear larger. Our model successfully handles both single-hand and dual-hand motions across a wide range of manipulation tasks, demonstrating its generalization ability under varied viewpoints and physical contexts.

\subsection{Real-World Experiments}
\label{sec:exp_robot}

To validate the real-world benefits of human video pretraining, we conduct experiments on dexterous hand manipulation tasks as shown in Table~\ref{tab:real-robot-exp}.
Here, ``\texttt{InternVL3}'' denotes using the same VLA architecture and parameters but without our physical instruction tuning.
\texttt{Being-H0} consistently achieves the highest success rates across all tasks, with each evaluated over 20 trials using randomized initial object positions.
In the \texttt{Pick-Place-Toy} task, while the finetuned GR00T N1.5~\cite{bjorck2025gr00t} performs comparably on in-domain, seen objects, its generalization degrades significantly when faced with unseen objects and cluttered scenes. 
Notably, GR00T employs an implicit learning approach to predict latent actions, whereas our \texttt{Being-H0}, through explicit motion tokenization, exhibits stronger generalization even with far less post-training data.
For example, Figure~\ref{fig:real-duck} demonstrates that \texttt{Being-H0} not only succeeds in picking the seen \textcolor{myyellow}{yellow} duck but also generalizes to the unseen \textcolor{mygreen}{green} duck.
More notably, in a cluttered environment with multiple distractors, \texttt{Being-H0} accurately follows the instruction “Pick the \textcolor{gray}{white} duck”, correctly identifying and retrieving the target \textcolor{gray}{white} duck. 
This highlights its robust integration of vision, language, and action understanding.

\begin{table*}[h]
\centering
\caption{Performance comparison: Success rates (\%) of \texttt{Being-H0} versus baseline models on real-world dexterous manipulation tasks.}
\label{tab:real-robot-exp}
\vspace{-2mm}
\setlength{\tabcolsep}{6pt}
\scalebox{0.9}{
\begin{tabular}{l|ccc|c|c|c|c}
\toprule
\multirow{2}{*}{Task} & \multicolumn{3}{c|}{Pick-Place-Toy} & \multirow{2}{*}{Close-Toolbox} & \multirow{2}{*}{Close-Lid} & \multirow{2}{*}{Pour-Cup} & \multirow{2}{*}{Unfold-Clothes} \\
\cline{2-4}
& \textit{Seen.} & \textit{Unseen.} & \textit{Clutter.} & & & \\
\midrule
GR00T N1.5~\cite{bjorck2025gr00t}  & \textbf{0.75} & 0.40 & 0.50 & 0.80 & 0.50 & 0.90 & 0.60 \\
InternVL3~\cite{zhu2025internvl3} & 0.55 & 0.55 & 0.50 & 0.50 & 0.25 & 0.55 & 0.45\\
Being-H0 & \textbf{0.75} & \textbf{0.65} & \textbf{0.60} & \textbf{0.85} & \textbf{0.60} & \textbf{1.00} & \textbf{0.75} \\
\bottomrule
\end{tabular}
}
\end{table*}

The advantages of \texttt{Being-H0} are particularly pronounced in tasks that demand fine-grained manipulation. 
The baseline model \texttt{InternVL3}, without physical instruction tuning and thus lacking prior knowledge of hand motion, exhibits significantly weaker performance. 
A qualitative comparison in Figure~\ref{fig:real-compare} clearly reveals the failure modes of \texttt{InternVL3}:
\begin{itemize}[leftmargin=1.5em]
    \item \texttt{Close-Toolbox}: Its motion trajectory lacks precision, often failing to contact the edge of the toolbox lid, which prevents it from applying sufficient force to close it.
    \item \texttt{Close-Lid}: It shows positional deviation, frequently misaligning the lid beside the cup's rim instead of seating it correctly.
    \item \texttt{Pour-Cup}: Its grasp is unstable, sometimes failing to hold the cup securely, which compromises the subsequent pouring motion's stability.
    \item \texttt{Unfold-Clothes}: It misjudges the operational height, causing its fingers to close at an improper elevation and miss the edge of the cloth, resulting in a failed unfolding action.
\end{itemize}

In sharp contrast, \texttt{Being-H0} demonstrates robust and precise behaviors across these tasks. 
It accurately positions and closes the lid, pinches the cloth's edge to unfold it, and maintains a stable grasp on the cup for smooth pouring. 
These successful outcomes highlight \texttt{Being-H0}'s ability to effectively transfer the knowledge of accurate hand motion generation from physical instruction tuning to downstream robotic tasks, enabling it to excel at complex and fine-grained dexterous manipulation.

\begin{figure*}[ht]
\centering
\includegraphics[width=1\textwidth, trim={0cm, 0cm, 0cm, 0cm}, clip]{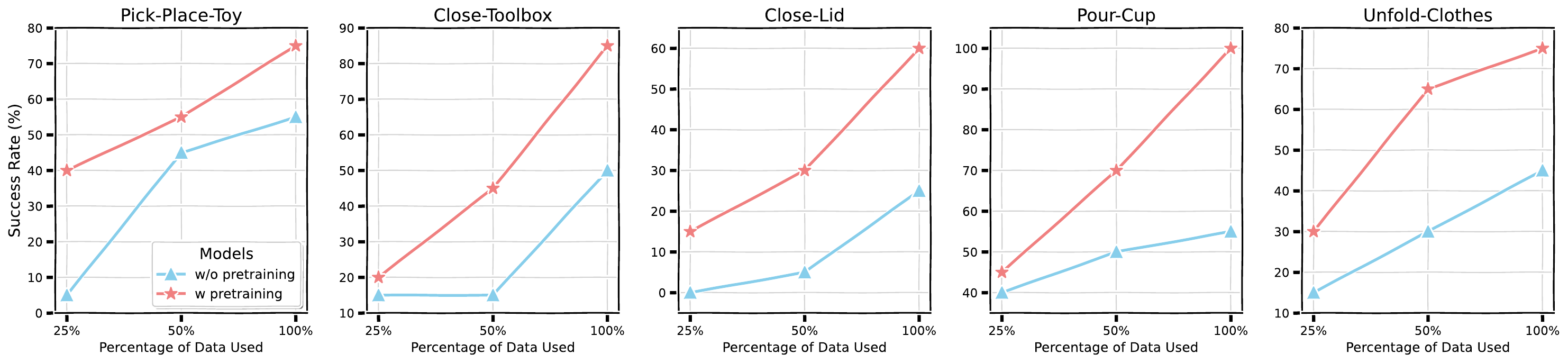}
\caption{Comparison of data efficiency between \texttt{Being-H0} and the baseline without human hand pre-training (InternVL3). For each task, the horizontal axis shows the percentage of teleoperation data used for finetuning, and the vertical axis reports the task success rate of the learned policy.}
\label{fig:sample_efficiency} 
\end{figure*}

\noindent\textbf{Data Efficiency. }
To assess the data efficiency of our \texttt{Being-H0}, we examine its downstream task performance when trained with different quantities of demonstration data.
We compare our pretrained model (\texttt{Being-H0}) against a non-pretrained baseline (InternVL3), evaluating both models with 25\%, 50\%, and 100\% of the available demonstration data across multiple tasks.
As shown in Figure~\ref{fig:sample_efficiency}, the results reveal that \texttt{Being-H0} maintains consistent and substantial performance advantages over the baseline at all data scales.
This result strongly demonstrates the significant benefits of our physical instruction tuning. 
The general hand motion priors acquired during pretraining provide the model with a strong starting point, allowing it to adapt more quickly to downstream dexterous manipulation tasks.

\texttt{Being-H0} can match or even exceed the baseline's performance while using only a fraction of the data. 
For instance, in the \texttt{Pick-Place-Toy} task, \texttt{Being-H0} trained on just 25\% of the data performs comparably to the baseline trained on 100\% of the data, while in the \texttt{Close-Toolbox} and \texttt{Unfold-Clothes} tasks, \texttt{Being-H0} trained on just 25\% of the data performs comparably to the baseline trained on 50\% of the data. 
In the more challenging \texttt{Close-Lid} task, the baseline fails completely (0\% success rate) with only 25\% of the data, whereas \texttt{Being-H0} already achieves a 15\% success rate, highlighting the difficulty of learning such a task from scratch. 
This superior data efficiency is crucial for real-world robotics. 
It greatly reduces the need for large-scale, high-quality teleoperated demonstrations.
This, in turn, lowers the cost and time of data collection and accelerates the deployment of dexterous robots in complex, real-world scenarios.

\section{Conclusion and Future Work}

We present \ModelName, an advanced, scalable and sample-efficient dexterous VLA trained on large-scale human videos through our novel \texttt{physical instruction tuning} paradigm comprising pretraining, physical space alignment, and post-training.
By using the human hand as the foundation manipulator and transferring dexterity from human videos to robotic learning, we resolve the pretraining-downstream data mismatch when adapting LMMs into VLAs.
Our unified million-level dataset \DataName systematically integrates diverse sources (motion capture, VR recordings, and RGB-only videos) to overcome real-robot data scarcity.

Our approach tackles four key challenges in learning dexterous manipulation from human videos.
For pretraining data curation, we systematically integrate heterogeneous data sources through MANO parameter standardization and projection alignment.
For hand quantization, our grouped residual quantization achieves millimeter-level reconstruction accuracy while enabling seamless integration with language models, effectively treating motion as another language.
For cross-modal reasoning, we unify all modalities within an autoregressive sequence to build sophisticated cross-modal dependencies linking visual scenes to manipulation strategies and grounding language instructions in precise finger movements. 
For robot control transfer, our physical instruction tuning leverages pretrained multimodal representations despite kinematic differences between human hands and robotic manipulators.

This work lays a foundation for learning robotic manipulation from human videos at scale, with several promising future directions.
First, further developing physical space alignment benefits transferring from human demonstrations to robotic control (e.g., incorporating depth and tactile feedback to enhance manipulation capabilities and physical plausibility).
Second, scaling \texttt{Being-H0} to complex scenarios involving tool use, multi-object interactions, and long-horizon reasoning presents an exciting frontier. 
Lastly, integration with simulation environments and reinforcement learning could enable more robust policy learning and safer real-world deployment.
We hope to pave the way for research in this increasingly important area and enable the development of more capable robotic systems that can operate in diverse real-world environments.

\vspace{3cm}
\section*{Credit Authorship}

\begin{multicols}{2} 
\noindent\textbf{Model Training}
\begin{itemize}
    \item \textsc{Hao Luo}
    \item \textsc{Yicheng Feng}
    \item \textsc{Wanpeng Zhang}
    \item \textsc{Sipeng Zheng}
\end{itemize}

\noindent\textbf{Motion Tokenization}
\begin{itemize}
    \item \textsc{Yicheng Feng}
    \item \textsc{Sipeng Zheng}
\end{itemize}

\noindent\textbf{Architecture \& Infrastructure}
\begin{itemize}
    \item \textsc{Wanpeng Zhang}
    \item \textsc{Hao Luo}
    \item \textsc{Sipeng Zheng}
\end{itemize}

\noindent\textbf{Data Source \& Alignment}
\begin{itemize}
    \item \textsc{Hao Luo}
    \item \textsc{Ye Wang}
    \item \textsc{Wanpeng Zhang}
    \item \textsc{Yicheng Feng}
    \item \textsc{Jiazheng Liu}
    \item \textsc{Sipeng Zheng}
\end{itemize}

\columnbreak

\noindent\textbf{Real-World Experiments}
\begin{itemize}
    \item \textsc{Ye Wang}
    \item \textsc{Haoqi Yuan}
\end{itemize}
\noindent\textbf{Robotic System}
\begin{itemize}
    \item \textsc{Haoqi Yuan}
    \item \textsc{Chaoyi Xu}
\end{itemize}

\noindent\textbf{Project Lead}
\begin{itemize}
    \item \textsc{Zongqing Lu}
\end{itemize}

\end{multicols} 

\clearpage

\bibliographystyle{unsrt}
\bibliography{ref}

\begin{thebibliography}{100}

\bibitem{sampath2023review}
Suhas~Kadalagere Sampath, Ning Wang, Hao Wu, and Chenguang Yang.
\newblock Review on human-like robot manipulation using dexterous hands.
\newblock {\em Cogn. Comput. Syst.}, 5(1):14--29, 2023.

\bibitem{huang2025human}
Yayu Huang, Dongxuan Fan, Haonan Duan, Dashun Yan, Wen Qi, Jia Sun, Qian Liu, and Peng Wang.
\newblock Human-like dexterous manipulation for anthropomorphic five-fingered hands: A review.
\newblock {\em Biomimetic Intelligence and Robotics}, page 100212, 2025.

\bibitem{brohan2022rt}
Anthony Brohan, Noah Brown, Justice Carbajal, Yevgen Chebotar, Joseph Dabis, Chelsea Finn, Keerthana Gopalakrishnan, Karol Hausman, Alex Herzog, Jasmine Hsu, et~al.
\newblock Rt-1: Robotics transformer for real-world control at scale.
\newblock {\em arXiv preprint arXiv:2212.06817}, 2022.

\bibitem{brohan2023rt}
Anthony Brohan, Noah Brown, Justice Carbajal, Yevgen Chebotar, Xi~Chen, Krzysztof Choromanski, Tianli Ding, Danny Driess, Avinava Dubey, Chelsea Finn, et~al.
\newblock Rt-2: Vision-language-action models transfer web knowledge to robotic control.
\newblock {\em arXiv preprint arXiv:2307.15818}, 2023.

\bibitem{kim2024openvla}
Moo~Jin Kim, Karl Pertsch, Siddharth Karamcheti, Ted Xiao, Ashwin Balakrishna, Suraj Nair, Rafael Rafailov, Ethan Foster, Grace Lam, Pannag Sanketi, et~al.
\newblock Openvla: An open-source vision-language-action model.
\newblock {\em arXiv preprint arXiv:2406.09246}, 2024.

\bibitem{black2410pi0}
Kevin Black, Noah Brown, Danny Driess, Adnan Esmail, Michael Equi, Chelsea Finn, Niccolo Fusai, Lachy Groom, Karol Hausman, Brian Ichter, et~al.
\newblock $\pi$0: A vision-language-action flow model for general robot control. corr, abs/2410.24164, 2024. doi: 10.48550.
\newblock {\em arXiv preprint ARXIV.2410.24164}, 2024.

\bibitem{ma2024survey}
Yueen Ma, Zixing Song, Yuzheng Zhuang, Jianye Hao, and Irwin King.
\newblock A survey on vision-language-action models for embodied ai.
\newblock {\em arXiv preprint arXiv:2405.14093}, 2024.

\bibitem{o2024open}
Abby O’Neill, Abdul Rehman, Abhiram Maddukuri, Abhishek Gupta, Abhishek Padalkar, Abraham Lee, Acorn Pooley, Agrim Gupta, Ajay Mandlekar, Ajinkya Jain, et~al.
\newblock Open x-embodiment: Robotic learning datasets and rt-x models: Open x-embodiment collaboration 0.
\newblock In {\em 2024 IEEE International Conference on Robotics and Automation (ICRA)}, pages 6892--6903. IEEE, 2024.

\bibitem{khazatsky2024droid}
Alexander Khazatsky, Karl Pertsch, Suraj Nair, Ashwin Balakrishna, Sudeep Dasari, Siddharth Karamcheti, Soroush Nasiriany, Mohan~Kumar Srirama, Lawrence~Yunliang Chen, Kirsty Ellis, et~al.
\newblock Droid: A large-scale in-the-wild robot manipulation dataset.
\newblock {\em arXiv preprint arXiv:2403.12945}, 2024.

\bibitem{bu2025agibot}
Qingwen Bu, Jisong Cai, Li~Chen, Xiuqi Cui, Yan Ding, Siyuan Feng, Shenyuan Gao, Xindong He, Xu~Huang, Shu Jiang, et~al.
\newblock Agibot world colosseo: A large-scale manipulation platform for scalable and intelligent embodied systems.
\newblock {\em arXiv preprint arXiv:2503.06669}, 2025.

\bibitem{team2024octo}
Octo~Model Team, Dibya Ghosh, Homer Walke, Karl Pertsch, Kevin Black, Oier Mees, Sudeep Dasari, Joey Hejna, Tobias Kreiman, Charles Xu, et~al.
\newblock Octo: An open-source generalist robot policy.
\newblock {\em arXiv preprint arXiv:2405.12213}, 2024.

\bibitem{cutler2024benchmarking}
Elizabeth Cutler, Yuning Xing, Tony Cui, Brendan Zhou, Koen van Rijnsoever, Ben Hart, David Valencia, Lee Violet~C Ong, Trevor Gee, Minas Liarokapis, et~al.
\newblock Benchmarking reinforcement learning methods for dexterous robotic manipulation with a three-fingered gripper.
\newblock {\em arXiv preprint arXiv:2408.14747}, 2024.

\bibitem{an2025dexterous}
Shan An, Ziyu Meng, Chao Tang, Yuning Zhou, Tengyu Liu, Fangqiang Ding, Shufang Zhang, Yao Mu, Ran Song, Wei Zhang, et~al.
\newblock Dexterous manipulation through imitation learning: A survey.
\newblock {\em arXiv preprint arXiv:2504.03515}, 2025.

\bibitem{guruprasad2024benchmarking}
Pranav Guruprasad, Harshvardhan Sikka, Jaewoo Song, Yangyue Wang, and Paul~Pu Liang.
\newblock Benchmarking vision, language, \& action models on robotic learning tasks.
\newblock {\em arXiv preprint arXiv:2411.05821}, 2024.

\bibitem{de2025scaffolding}
Vincent de~Bakker, Joey Hejna, Tyler Ga~Wei Lum, Onur Celik, Aleksandar Taranovic, Denis Blessing, Gerhard Neumann, Jeannette Bohg, and Dorsa Sadigh.
\newblock Scaffolding dexterous manipulation with vision-language models.
\newblock {\em arXiv preprint arXiv:2506.19212}, 2025.

\bibitem{xu2023unidexgrasp}
Yinzhen Xu, Weikang Wan, Jialiang Zhang, Haoran Liu, Zikang Shan, Hao Shen, Ruicheng Wang, Haoran Geng, Yijia Weng, Jiayi Chen, et~al.
\newblock Unidexgrasp: Universal robotic dexterous grasping via learning diverse proposal generation and goal-conditioned policy.
\newblock In {\em Proceedings of the IEEE/CVF Conference on Computer Vision and Pattern Recognition}, pages 4737--4746, 2023.

\bibitem{wan2023unidexgrasp++}
Weikang Wan, Haoran Geng, Yun Liu, Zikang Shan, Yaodong Yang, Li~Yi, and He~Wang.
\newblock Unidexgrasp++: Improving dexterous grasping policy learning via geometry-aware curriculum and iterative generalist-specialist learning.
\newblock In {\em Proceedings of the IEEE/CVF International Conference on Computer Vision}, pages 3891--3902, 2023.

\bibitem{zhong2025dexgraspvla}
Yifan Zhong, Xuchuan Huang, Ruochong Li, Ceyao Zhang, Yitao Liang, Yaodong Yang, and Yuanpei Chen.
\newblock Dexgraspvla: A vision-language-action framework towards general dexterous grasping.
\newblock {\em arXiv preprint arXiv:2502.20900}, 2025.

\bibitem{he2025dexvlg}
Jiawei He, Danshi Li, Xinqiang Yu, Zekun Qi, Wenyao Zhang, Jiayi Chen, Zhaoxiang Zhang, Zhizheng Zhang, Li~Yi, and He~Wang.
\newblock Dexvlg: Dexterous vision-language-grasp model at scale.
\newblock {\em arXiv preprint arXiv:2507.02747}, 2025.

\bibitem{huang2025efficient}
Ziye Huang, Haoqi Yuan, Yuhui Fu, and Zongqing Lu.
\newblock Efficient residual learning with mixture-of-experts for universal dexterous grasping.
\newblock In {\em The Thirteenth International Conference on Learning Representations}, 2025.

\bibitem{nair2022r3m}
Suraj Nair, Aravind Rajeswaran, Vikash Kumar, Chelsea Finn, and Abhinav Gupta.
\newblock R3m: A universal visual representation for robot manipulation.
\newblock {\em arXiv preprint arXiv:2203.12601}, 2022.

\bibitem{radosavovic2023real}
Ilija Radosavovic, Tete Xiao, Stephen James, Pieter Abbeel, Jitendra Malik, and Trevor Darrell.
\newblock Real-world robot learning with masked visual pre-training.
\newblock In {\em Conference on Robot Learning}, pages 416--426. PMLR, 2023.

\bibitem{bjorck2025gr00t}
Johan Bjorck, Fernando Casta{\~n}eda, Nikita Cherniadev, Xingye Da, Runyu Ding, Linxi Fan, Yu~Fang, Dieter Fox, Fengyuan Hu, Spencer Huang, et~al.
\newblock Gr00t n1: An open foundation model for generalist humanoid robots.
\newblock {\em arXiv preprint arXiv:2503.14734}, 2025.

\bibitem{liu2024improved}
Haotian Liu, Chunyuan Li, Yuheng Li, and Yong~Jae Lee.
\newblock Improved baselines with visual instruction tuning.
\newblock In {\em Proceedings of the IEEE/CVF conference on computer vision and pattern recognition}, pages 26296--26306, 2024.

\bibitem{qiu2025humanoid}
Ri-Zhao Qiu, Shiqi Yang, Xuxin Cheng, Chaitanya Chawla, Jialong Li, Tairan He, Ge~Yan, David~J Yoon, Ryan Hoque, Lars Paulsen, et~al.
\newblock Humanoid policy\~{} human policy.
\newblock {\em arXiv preprint arXiv:2503.13441}, 2025.

\bibitem{kim2021integrated}
Uikyum Kim, Dawoon Jung, Heeyoen Jeong, Jongwoo Park, Hyun-Mok Jung, Joono Cheong, Hyouk~Ryeol Choi, Hyunmin Do, and Chanhun Park.
\newblock Integrated linkage-driven dexterous anthropomorphic robotic hand.
\newblock {\em Nature communications}, 12(1):7177, 2021.

\bibitem{liu2023visual}
Haotian Liu, Chunyuan Li, Qingyang Wu, and Yong~Jae Lee.
\newblock Visual instruction tuning.
\newblock {\em Advances in neural information processing systems}, 36:34892--34916, 2023.

\bibitem{lee2022autoregressive}
Doyup Lee, Chiheon Kim, Saehoon Kim, Minsu Cho, and Wook-Shin Han.
\newblock Autoregressive image generation using residual quantization.
\newblock In {\em Proceedings of the IEEE/CVF Conference on Computer Vision and Pattern Recognition}, pages 11523--11532, 2022.

\bibitem{yang2023hifi}
Dongchao Yang, Songxiang Liu, Rongjie Huang, Jinchuan Tian, Chao Weng, and Yuexian Zou.
\newblock Hifi-codec: Group-residual vector quantization for high fidelity audio codec.
\newblock {\em arXiv preprint arXiv:2305.02765}, 2023.

\bibitem{vaswani2017attention}
Ashish Vaswani, Noam Shazeer, Niki Parmar, Jakob Uszkoreit, Llion Jones, Aidan~N Gomez, {\L}ukasz Kaiser, and Illia Polosukhin.
\newblock Attention is all you need.
\newblock {\em Advances in neural information processing systems}, 30, 2017.

\bibitem{radford2018improving}
Alec Radford, Karthik Narasimhan, Tim Salimans, Ilya Sutskever, et~al.
\newblock Improving language understanding by generative pre-training.
\newblock In {\em arxiv}. San Francisco, CA, USA, 2018.

\bibitem{radford2019language}
Alec Radford, Jeffrey Wu, Rewon Child, David Luan, Dario Amodei, Ilya Sutskever, et~al.
\newblock Language models are unsupervised multitask learners.
\newblock {\em OpenAI blog}, 1(8):9, 2019.

\bibitem{brown2020language}
Tom Brown, Benjamin Mann, Nick Ryder, Melanie Subbiah, Jared~D Kaplan, Prafulla Dhariwal, Arvind Neelakantan, Pranav Shyam, Girish Sastry, Amanda Askell, et~al.
\newblock Language models are few-shot learners.
\newblock {\em Advances in neural information processing systems}, 33:1877--1901, 2020.

\bibitem{zhu2023minigpt}
Deyao Zhu, Jun Chen, Xiaoqian Shen, Xiang Li, and Mohamed Elhoseiny.
\newblock Minigpt-4: Enhancing vision-language understanding with advanced large language models.
\newblock {\em arXiv preprint arXiv:2304.10592}, 2023.

\bibitem{wang2024qwen2}
Peng Wang, Shuai Bai, Sinan Tan, Shijie Wang, Zhihao Fan, Jinze Bai, Keqin Chen, Xuejing Liu, Jialin Wang, Wenbin Ge, et~al.
\newblock Qwen2-vl: Enhancing vision-language model's perception of the world at any resolution.
\newblock {\em arXiv preprint arXiv:2409.12191}, 2024.

\bibitem{zheng2024unicode}
Sipeng Zheng, Bohan Zhou, Yicheng Feng, Ye~Wang, and Zongqing Lu.
\newblock Unicode: Learning a unified codebook for multimodal large language models.
\newblock {\em arXiv preprint arXiv:2403.09072}, 2024.

\bibitem{zhang2025beingvl0}
Wanpeng Zhang, Zilong Xie, Yicheng Feng, Yijiang Li, Xingrun Xing, Sipeng Zheng, and Zongqing Lu.
\newblock From pixels to tokens: Byte-pair encoding on quantized visual modalities.
\newblock In {\em The Thirteenth International Conference on Learning Representations}, 2025.

\bibitem{zhang2025beingvl05}
Wanpeng Zhang, Yicheng Feng, Hao Luo, Yijiang Li, Zihao Yue, Sipeng Zheng, and Zongqing Lu.
\newblock Unified multimodal understanding via byte-pair visual encoding.
\newblock In {\em Proceedings of the IEEE/CVF International Conference on Computer Vision}, 2025.

\bibitem{touvron2023llama}
Hugo Touvron, Thibaut Lavril, Gautier Izacard, Xavier Martinet, Marie-Anne Lachaux, Timoth{\'e}e Lacroix, Baptiste Rozi{\`e}re, Naman Goyal, Eric Hambro, Faisal Azhar, et~al.
\newblock Llama: Open and efficient foundation language models.
\newblock {\em arXiv preprint arXiv:2302.13971}, 2023.

\bibitem{touvron2023llama2}
Hugo Touvron, Louis Martin, Kevin Stone, Peter Albert, Amjad Almahairi, Yasmine Babaei, Nikolay Bashlykov, Soumya Batra, Prajjwal Bhargava, Shruti Bhosale, et~al.
\newblock Llama 2: Open foundation and fine-tuned chat models.
\newblock {\em arXiv preprint arXiv:2307.09288}, 2023.

\bibitem{bai2023qwen}
Jinze Bai, Shuai Bai, Yunfei Chu, Zeyu Cui, Kai Dang, Xiaodong Deng, Yang Fan, Wenbin Ge, Yu~Han, Fei Huang, et~al.
\newblock Qwen technical report.
\newblock {\em arXiv preprint arXiv:2309.16609}, 2023.

\bibitem{radford2021learning}
Alec Radford, Jong~Wook Kim, Chris Hallacy, Aditya Ramesh, Gabriel Goh, Sandhini Agarwal, Girish Sastry, Amanda Askell, Pamela Mishkin, Jack Clark, et~al.
\newblock Learning transferable visual models from natural language supervision.
\newblock In {\em International conference on machine learning}, pages 8748--8763. PmLR, 2021.

\bibitem{zhai2023sigmoid}
Xiaohua Zhai, Basil Mustafa, Alexander Kolesnikov, and Lucas Beyer.
\newblock Sigmoid loss for language image pre-training.
\newblock In {\em Proceedings of the IEEE/CVF international conference on computer vision}, pages 11975--11986, 2023.

\bibitem{alayrac2022flamingo}
Jean-Baptiste Alayrac, Jeff Donahue, Pauline Luc, Antoine Miech, Iain Barr, Yana Hasson, Karel Lenc, Arthur Mensch, Katherine Millican, Malcolm Reynolds, et~al.
\newblock Flamingo: a visual language model for few-shot learning.
\newblock {\em Advances in neural information processing systems}, 35:23716--23736, 2022.

\bibitem{lu2023empirical}
Yadong Lu, Chunyuan Li, Haotian Liu, Jianwei Yang, Jianfeng Gao, and Yelong Shen.
\newblock An empirical study of scaling instruct-tuned large multimodal models.
\newblock {\em arXiv preprint arXiv:2309.09958}, 2023.

\bibitem{you2023ferret}
Haoxuan You, Haotian Zhang, Zhe Gan, Xianzhi Du, Bowen Zhang, Zirui Wang, Liangliang Cao, Shih-Fu Chang, and Yinfei Yang.
\newblock Ferret: Refer and ground anything anywhere at any granularity.
\newblock {\em arXiv preprint arXiv:2310.07704}, 2023.

\bibitem{li2025otter}
Bo~Li, Yuanhan Zhang, Liangyu Chen, Jinghao Wang, Fanyi Pu, Joshua~Adrian Cahyono, Jingkang Yang, Chunyuan Li, and Ziwei Liu.
\newblock Otter: A multi-modal model with in-context instruction tuning.
\newblock {\em IEEE Transactions on Pattern Analysis and Machine Intelligence}, 2025.

\bibitem{chen2024allava}
Guiming~Hardy Chen, Shunian Chen, Ruifei Zhang, Junying Chen, Xiangbo Wu, Zhiyi Zhang, Zhihong Chen, Jianquan Li, Xiang Wan, and Benyou Wang.
\newblock Allava: Harnessing gpt4v-synthesized data for lite vision-language models.
\newblock {\em arXiv preprint arXiv:2402.11684}, 2024.

\bibitem{zheng2023steve}
Sipeng Zheng, Jiazheng Liu, Yicheng Feng, and Zongqing Lu.
\newblock Steve-eye: Equipping llm-based embodied agents with visual perception in open worlds.
\newblock {\em arXiv preprint arXiv:2310.13255}, 2023.

\bibitem{feng2024videoorion}
Yicheng Feng, Yijiang Li, Wanpeng Zhang, Sipeng Zheng, and Zongqing Lu.
\newblock Videoorion: Tokenizing object dynamics in videos.
\newblock {\em arXiv preprint arXiv:2411.16156}, 2024.

\bibitem{liu2025taking}
Jiazheng Liu, Sipeng Zheng, B{\"o}rje~F Karlsson, and Zongqing Lu.
\newblock Taking notes brings focus? towards multi-turn multimodal dialogue learning.
\newblock {\em arXiv preprint arXiv:2503.07002}, 2025.

\bibitem{team2023gemini}
Gemini Team, Rohan Anil, Sebastian Borgeaud, Jean-Baptiste Alayrac, Jiahui Yu, Radu Soricut, Johan Schalkwyk, Andrew~M Dai, Anja Hauth, Katie Millican, et~al.
\newblock Gemini: a family of highly capable multimodal models.
\newblock {\em arXiv preprint arXiv:2312.11805}, 2023.

\bibitem{comanici2025gemini}
Gheorghe Comanici, Eric Bieber, Mike Schaekermann, Ice Pasupat, Noveen Sachdeva, Inderjit Dhillon, Marcel Blistein, Ori Ram, Dan Zhang, Evan Rosen, et~al.
\newblock Gemini 2.5: Pushing the frontier with advanced reasoning, multimodality, long context, and next generation agentic capabilities.
\newblock {\em arXiv preprint arXiv:2507.06261}, 2025.

\bibitem{achiam2023gpt}
Josh Achiam, Steven Adler, Sandhini Agarwal, Lama Ahmad, Ilge Akkaya, Florencia~Leoni Aleman, Diogo Almeida, Janko Altenschmidt, Sam Altman, Shyamal Anadkat, et~al.
\newblock Gpt-4 technical report.
\newblock {\em arXiv preprint arXiv:2303.08774}, 2023.

\bibitem{bai2025qwen2}
Shuai Bai, Keqin Chen, Xuejing Liu, Jialin Wang, Wenbin Ge, Sibo Song, Kai Dang, Peng Wang, Shijie Wang, Jun Tang, et~al.
\newblock Qwen2. 5-vl technical report.
\newblock {\em arXiv preprint arXiv:2502.13923}, 2025.

\bibitem{team2024chameleon}
Chameleon Team.
\newblock Chameleon: Mixed-modal early-fusion foundation models.
\newblock {\em arXiv preprint arXiv:2405.09818}, 2024.

\bibitem{zhu2025internvl3}
Jinguo Zhu, Weiyun Wang, Zhe Chen, Zhaoyang Liu, Shenglong Ye, Lixin Gu, Hao Tian, Yuchen Duan, Weijie Su, Jie Shao, et~al.
\newblock Internvl3: Exploring advanced training and test-time recipes for open-source multimodal models.
\newblock {\em arXiv preprint arXiv:2504.10479}, 2025.

\bibitem{deitke2024molmo}
Matt Deitke, Christopher Clark, Sangho Lee, Rohun Tripathi, Yue Yang, Jae~Sung Park, Mohammadreza Salehi, Niklas Muennighoff, Kyle Lo, Luca Soldaini, et~al.
\newblock Molmo and pixmo: Open weights and open data for state-of-the-art multimodal models.
\newblock {\em arXiv e-prints}, pages arXiv--2409, 2024.

\bibitem{plappert2016kit}
Matthias Plappert, Christian Mandery, and Tamim Asfour.
\newblock The kit motion-language dataset.
\newblock {\em Big data}, 4(4):236--252, 2016.

\bibitem{mahmood2019amass}
Naureen Mahmood, Nima Ghorbani, Nikolaus~F Troje, Gerard Pons-Moll, and Michael~J Black.
\newblock Amass: Archive of motion capture as surface shapes.
\newblock In {\em Proceedings of the IEEE/CVF international conference on computer vision}, pages 5442--5451, 2019.

\bibitem{guo2022generating}
Chuan Guo, Shihao Zou, Xinxin Zuo, Sen Wang, Wei Ji, Xingyu Li, and Li~Cheng.
\newblock Generating diverse and natural 3d human motions from text.
\newblock In {\em Proceedings of the IEEE/CVF conference on computer vision and pattern recognition}, pages 5152--5161, 2022.

\bibitem{punnakkal2021babel}
Abhinanda~R Punnakkal, Arjun Chandrasekaran, Nikos Athanasiou, Alejandra Quiros-Ramirez, and Michael~J Black.
\newblock Babel: Bodies, action and behavior with english labels.
\newblock In {\em Proceedings of the IEEE/CVF conference on computer vision and pattern recognition}, pages 722--731, 2021.

\bibitem{lin2023motion}
Jing Lin, Ailing Zeng, Shunlin Lu, Yuanhao Cai, Ruimao Zhang, Haoqian Wang, and Lei Zhang.
\newblock Motion-x: A large-scale 3d expressive whole-body human motion dataset.
\newblock {\em Advances in Neural Information Processing Systems}, 36:25268--25280, 2023.

\bibitem{zhang2022egobody}
Siwei Zhang, Qianli Ma, Yan Zhang, Zhiyin Qian, Taein Kwon, Marc Pollefeys, Federica Bogo, and Siyu Tang.
\newblock Egobody: Human body shape and motion of interacting people from head-mounted devices.
\newblock In {\em European conference on computer vision}, pages 180--200. Springer, 2022.

\bibitem{ma2024nymeria}
Lingni Ma, Yuting Ye, Fangzhou Hong, Vladimir Guzov, Yifeng Jiang, Rowan Postyeni, Luis Pesqueira, Alexander Gamino, Vijay Baiyya, Hyo~Jin Kim, et~al.
\newblock Nymeria: A massive collection of multimodal egocentric daily motion in the wild.
\newblock In {\em European Conference on Computer Vision}, pages 445--465. Springer, 2024.

\bibitem{wangscaling}
Ye~Wang, Sipeng Zheng, Bin Cao, Qianshan Wei, Weishuai Zeng, Qin Jin, and Zongqing Lu.
\newblock Scaling large motion models with million-level human motions.
\newblock In {\em Forty-second International Conference on Machine Learning}, 2024.

\bibitem{loper2023smpl}
Matthew Loper, Naureen Mahmood, Javier Romero, Gerard Pons-Moll, and Michael~J Black.
\newblock Smpl: A skinned multi-person linear model.
\newblock In {\em Seminal Graphics Papers: Pushing the Boundaries, Volume 2}, pages 851--866, 2023.

\bibitem{pavlakos2019expressive}
Georgios Pavlakos, Vasileios Choutas, Nima Ghorbani, Timo Bolkart, Ahmed~AA Osman, Dimitrios Tzionas, and Michael~J Black.
\newblock Expressive body capture: 3d hands, face, and body from a single image.
\newblock In {\em Proceedings of the IEEE/CVF conference on computer vision and pattern recognition}, pages 10975--10985, 2019.

\bibitem{tevet2022human}
Guy Tevet, Sigal Raab, Brian Gordon, Yonatan Shafir, Daniel Cohen-Or, and Amit~H Bermano.
\newblock Human motion diffusion model.
\newblock {\em arXiv preprint arXiv:2209.14916}, 2022.

\bibitem{zhang2024motiondiffuse}
Mingyuan Zhang, Zhongang Cai, Liang Pan, Fangzhou Hong, Xinying Guo, Lei Yang, and Ziwei Liu.
\newblock Motiondiffuse: Text-driven human motion generation with diffusion model.
\newblock {\em IEEE transactions on pattern analysis and machine intelligence}, 46(6):4115--4128, 2024.

\bibitem{chen2023executing}
Xin Chen, Biao Jiang, Wen Liu, Zilong Huang, Bin Fu, Tao Chen, and Gang Yu.
\newblock Executing your commands via motion diffusion in latent space.
\newblock In {\em Proceedings of the IEEE/CVF conference on computer vision and pattern recognition}, pages 18000--18010, 2023.

\bibitem{yuan2023physdiff}
Ye~Yuan, Jiaming Song, Umar Iqbal, Arash Vahdat, and Jan Kautz.
\newblock Physdiff: Physics-guided human motion diffusion model.
\newblock In {\em Proceedings of the IEEE/CVF international conference on computer vision}, pages 16010--16021, 2023.

\bibitem{zhang2023remodiffuse}
Mingyuan Zhang, Xinying Guo, Liang Pan, Zhongang Cai, Fangzhou Hong, Huirong Li, Lei Yang, and Ziwei Liu.
\newblock Remodiffuse: Retrieval-augmented motion diffusion model.
\newblock In {\em Proceedings of the IEEE/CVF International Conference on Computer Vision}, pages 364--373, 2023.

\bibitem{lou2023diversemotion}
Yunhong Lou, Linchao Zhu, Yaxiong Wang, Xiaohan Wang, and Yi~Yang.
\newblock Diversemotion: Towards diverse human motion generation via discrete diffusion.
\newblock {\em arXiv preprint arXiv:2309.01372}, 2023.

\bibitem{zhang2024large}
Mingyuan Zhang, Daisheng Jin, Chenyang Gu, Fangzhou Hong, Zhongang Cai, Jingfang Huang, Chongzhi Zhang, Xinying Guo, Lei Yang, Ying He, et~al.
\newblock Large motion model for unified multi-modal motion generation.
\newblock In {\em European Conference on Computer Vision}, pages 397--421. Springer, 2024.

\bibitem{van2017neural}
Aaron Van Den~Oord, Oriol Vinyals, et~al.
\newblock Neural discrete representation learning.
\newblock {\em Advances in neural information processing systems}, 30, 2017.

\bibitem{zhang2023generating}
Jianrong Zhang, Yangsong Zhang, Xiaodong Cun, Yong Zhang, Hongwei Zhao, Hongtao Lu, Xi~Shen, and Ying Shan.
\newblock Generating human motion from textual descriptions with discrete representations.
\newblock In {\em Proceedings of the IEEE/CVF conference on computer vision and pattern recognition}, pages 14730--14740, 2023.

\bibitem{guo2024momask}
Chuan Guo, Yuxuan Mu, Muhammad~Gohar Javed, Sen Wang, and Li~Cheng.
\newblock Momask: Generative masked modeling of 3d human motions.
\newblock In {\em Proceedings of the IEEE/CVF Conference on Computer Vision and Pattern Recognition}, pages 1900--1910, 2024.

\bibitem{you2022locally}
Tackgeun You, Saehoon Kim, Chiheon Kim, Doyup Lee, and Bohyung Han.
\newblock Locally hierarchical auto-regressive modeling for image generation.
\newblock {\em Advances in Neural Information Processing Systems}, 35:16360--16372, 2022.

\bibitem{lu2023humantomato}
Shunlin Lu, Ling-Hao Chen, Ailing Zeng, Jing Lin, Ruimao Zhang, Lei Zhang, and Heung-Yeung Shum.
\newblock Humantomato: Text-aligned whole-body motion generation.
\newblock {\em arXiv preprint arXiv:2310.12978}, 2023.

\bibitem{mentzer2023finite}
Fabian Mentzer, David Minnen, Eirikur Agustsson, and Michael Tschannen.
\newblock Finite scalar quantization: Vq-vae made simple.
\newblock {\em arXiv preprint arXiv:2309.15505}, 2023.

\bibitem{yu2023language}
Lijun Yu, Jos{\'e} Lezama, Nitesh~B Gundavarapu, Luca Versari, Kihyuk Sohn, David Minnen, Yong Cheng, Vighnesh Birodkar, Agrim Gupta, Xiuye Gu, et~al.
\newblock Language model beats diffusion--tokenizer is key to visual generation.
\newblock {\em arXiv preprint arXiv:2310.05737}, 2023.

\bibitem{jiang2023motiongpt}
Biao Jiang, Xin Chen, Wen Liu, Jingyi Yu, Gang Yu, and Tao Chen.
\newblock Motiongpt: Human motion as a foreign language.
\newblock {\em Advances in Neural Information Processing Systems}, 36:20067--20079, 2023.

\bibitem{wang2024motiongpt}
Yuan Wang, Di~Huang, Yaqi Zhang, Wanli Ouyang, Jile Jiao, Xuetao Feng, Yan Zhou, Pengfei Wan, Shixiang Tang, and Dan Xu.
\newblock Motiongpt-2: A general-purpose motion-language model for motion generation and understanding.
\newblock {\em arXiv preprint arXiv:2410.21747}, 2024.

\bibitem{chen2024motionllm}
Ling-Hao Chen, Shunlin Lu, Ailing Zeng, Hao Zhang, Benyou Wang, Ruimao Zhang, and Lei Zhang.
\newblock Motionllm: Understanding human behaviors from human motions and videos.
\newblock {\em arXiv preprint arXiv:2405.20340}, 2024.

\bibitem{zhou2024avatargpt}
Zixiang Zhou, Yu~Wan, and Baoyuan Wang.
\newblock Avatargpt: All-in-one framework for motion understanding planning generation and beyond.
\newblock In {\em Proceedings of the IEEE/CVF Conference on Computer Vision and Pattern Recognition}, pages 1357--1366, 2024.

\bibitem{jiang2024motionchain}
Biao Jiang, Xin Chen, Chi Zhang, Fukun Yin, Zhuoyuan Li, Gang Yu, and Jiayuan Fan.
\newblock Motionchain: Conversational motion controllers via multimodal prompts.
\newblock In {\em European Conference on Computer Vision}, pages 54--74. Springer, 2024.

\bibitem{li2024lamp}
Zhe Li, Weihao Yuan, Yisheng He, Lingteng Qiu, Shenhao Zhu, Xiaodong Gu, Weichao Shen, Yuan Dong, Zilong Dong, and Laurence~T Yang.
\newblock Lamp: Language-motion pretraining for motion generation, retrieval, and captioning.
\newblock {\em arXiv preprint arXiv:2410.07093}, 2024.

\bibitem{li2025human}
Lei Li, Sen Jia, Jianhao Wang, Zhongyu Jiang, Feng Zhou, Ju~Dai, Tianfang Zhang, Zongkai Wu, and Jenq-Neng Hwang.
\newblock Human motion instruction tuning.
\newblock In {\em Proceedings of the Computer Vision and Pattern Recognition Conference}, pages 17582--17591, 2025.

\bibitem{yue2025rl}
Junpeng Yue, Zepeng Wang, Yuxuan Wang, Weishuai Zeng, Jiangxing Wang, Xinrun Xu, Yu~Zhang, Sipeng Zheng, Ziluo Ding, and Zongqing Lu.
\newblock Rl from physical feedback: Aligning large motion models with humanoid control.
\newblock {\em arXiv preprint arXiv:2506.12769}, 2025.

\bibitem{luo2023perpetual}
Zhengyi Luo, Jinkun Cao, Kris Kitani, Weipeng Xu, et~al.
\newblock Perpetual humanoid control for real-time simulated avatars.
\newblock In {\em Proceedings of the IEEE/CVF International Conference on Computer Vision}, pages 10895--10904, 2023.

\bibitem{ji2024exbody2}
Mazeyu Ji, Xuanbin Peng, Fangchen Liu, Jialong Li, Ge~Yang, Xuxin Cheng, and Xiaolong Wang.
\newblock Exbody2: Advanced expressive humanoid whole-body control.
\newblock {\em arXiv preprint arXiv:2412.13196}, 2024.

\bibitem{han2025reindiffuse}
Gaoge Han, Mingjiang Liang, Jinglei Tang, Yongkang Cheng, Wei Liu, and Shaoli Huang.
\newblock Reindiffuse: Crafting physically plausible motions with reinforced diffusion model.
\newblock In {\em 2025 IEEE/CVF Winter Conference on Applications of Computer Vision (WACV)}, pages 2218--2227. IEEE, 2025.

\bibitem{grauman2024ego}
Kristen Grauman, Andrew Westbury, Lorenzo Torresani, Kris Kitani, Jitendra Malik, Triantafyllos Afouras, Kumar Ashutosh, Vijay Baiyya, Siddhant Bansal, Bikram Boote, et~al.
\newblock Ego-exo4d: Understanding skilled human activity from first-and third-person perspectives.
\newblock In {\em Proceedings of the IEEE/CVF Conference on Computer Vision and Pattern Recognition}, pages 19383--19400, 2024.

\bibitem{jiang2021hand}
Hanwen Jiang, Shaowei Liu, Jiashun Wang, and Xiaolong Wang.
\newblock Hand-object contact consistency reasoning for human grasps generation.
\newblock In {\em Proceedings of the IEEE/CVF international conference on computer vision}, pages 11107--11116, 2021.

\bibitem{liu2022joint}
Shaowei Liu, Subarna Tripathi, Somdeb Majumdar, and Xiaolong Wang.
\newblock Joint hand motion and interaction hotspots prediction from egocentric videos.
\newblock In {\em Proceedings of the IEEE/CVF Conference on Computer Vision and Pattern Recognition}, pages 3282--3292, 2022.

\bibitem{banerjee2025hot3d}
Prithviraj Banerjee, Sindi Shkodrani, Pierre Moulon, Shreyas Hampali, Shangchen Han, Fan Zhang, Linguang Zhang, Jade Fountain, Edward Miller, Selen Basol, et~al.
\newblock Hot3d: Hand and object tracking in 3d from egocentric multi-view videos.
\newblock In {\em Proceedings of the Computer Vision and Pattern Recognition Conference}, pages 7061--7071, 2025.

\bibitem{liu2022hoi4d}
Yunze Liu, Yun Liu, Che Jiang, Kangbo Lyu, Weikang Wan, Hao Shen, Boqiang Liang, Zhoujie Fu, He~Wang, and Li~Yi.
\newblock Hoi4d: A 4d egocentric dataset for category-level human-object interaction.
\newblock In {\em Proceedings of the IEEE/CVF Conference on Computer Vision and Pattern Recognition}, pages 21013--21022, 2022.

\bibitem{zhan2024oakink2}
Xinyu Zhan, Lixin Yang, Yifei Zhao, Kangrui Mao, Hanlin Xu, Zenan Lin, Kailin Li, and Cewu Lu.
\newblock Oakink2: A dataset of bimanual hands-object manipulation in complex task completion.
\newblock In {\em Proceedings of the IEEE/CVF Conference on Computer Vision and Pattern Recognition}, pages 445--456, 2024.

\bibitem{grauman2022ego4d}
Kristen Grauman, Andrew Westbury, Eugene Byrne, Zachary Chavis, Antonino Furnari, Rohit Girdhar, Jackson Hamburger, Hao Jiang, Miao Liu, Xingyu Liu, et~al.
\newblock Ego4d: Around the world in 3,000 hours of egocentric video.
\newblock In {\em Proceedings of the IEEE/CVF conference on computer vision and pattern recognition}, pages 18995--19012, 2022.

\bibitem{hasson2019learning}
Yana Hasson, Gul Varol, Dimitrios Tzionas, Igor Kalevatykh, Michael~J Black, Ivan Laptev, and Cordelia Schmid.
\newblock Learning joint reconstruction of hands and manipulated objects.
\newblock In {\em Proceedings of the IEEE/CVF conference on computer vision and pattern recognition}, pages 11807--11816, 2019.

\bibitem{pavlakos2024reconstructing}
Georgios Pavlakos, Dandan Shan, Ilija Radosavovic, Angjoo Kanazawa, David Fouhey, and Jitendra Malik.
\newblock Reconstructing hands in 3d with transformers.
\newblock In {\em Proceedings of the IEEE/CVF Conference on Computer Vision and Pattern Recognition}, pages 9826--9836, 2024.

\bibitem{dong2024hamba}
Haoye Dong, Aviral Chharia, Wenbo Gou, Francisco Vicente~Carrasco, and Fernando~D De~la Torre.
\newblock Hamba: Single-view 3d hand reconstruction with graph-guided bi-scanning mamba.
\newblock {\em Advances in Neural Information Processing Systems}, 37:2127--2160, 2024.

\bibitem{fan2024hold}
Zicong Fan, Maria Parelli, Maria~Eleni Kadoglou, Xu~Chen, Muhammed Kocabas, Michael~J Black, and Otmar Hilliges.
\newblock Hold: Category-agnostic 3d reconstruction of interacting hands and objects from video.
\newblock In {\em Proceedings of the IEEE/CVF Conference on Computer Vision and Pattern Recognition}, pages 494--504, 2024.

\bibitem{damen2018scaling}
Dima Damen, Hazel Doughty, Giovanni~Maria Farinella, Sanja Fidler, Antonino Furnari, Evangelos Kazakos, Davide Moltisanti, Jonathan Munro, Toby Perrett, Will Price, et~al.
\newblock Scaling egocentric vision: The epic-kitchens dataset.
\newblock In {\em Proceedings of the European conference on computer vision (ECCV)}, pages 720--736, 2018.

\bibitem{yu2025dyn}
Zhengdi Yu, Stefanos Zafeiriou, and Tolga Birdal.
\newblock Dyn-hamr: Recovering 4d interacting hand motion from a dynamic camera.
\newblock In {\em Proceedings of the Computer Vision and Pattern Recognition Conference}, pages 27716--27726, 2025.

\bibitem{gkioxari2015contextual}
Georgia Gkioxari, Ross Girshick, and Jitendra Malik.
\newblock Contextual action recognition with r* cnn.
\newblock In {\em Proceedings of the IEEE international conference on computer vision}, pages 1080--1088, 2015.

\bibitem{mallya2016learning}
Arun Mallya and Svetlana Lazebnik.
\newblock Learning models for actions and person-object interactions with transfer to question answering.
\newblock In {\em European Conference on Computer Vision}, pages 414--428. Springer, 2016.

\bibitem{zheng2023open}
Sipeng Zheng, Boshen Xu, and Qin Jin.
\newblock Open-category human-object interaction pre-training via language modeling framework.
\newblock In {\em Proceedings of the IEEE/CVF Conference on Computer Vision and Pattern Recognition}, pages 19392--19402, 2023.

\bibitem{gkioxari2018detecting}
Georgia Gkioxari, Ross Girshick, Piotr Doll{\'a}r, and Kaiming He.
\newblock Detecting and recognizing human-object interactions.
\newblock In {\em Proceedings of the IEEE conference on computer vision and pattern recognition}, pages 8359--8367, 2018.

\bibitem{qi2018learning}
Siyuan Qi, Wenguan Wang, Baoxiong Jia, Jianbing Shen, and Song-Chun Zhu.
\newblock Learning human-object interactions by graph parsing neural networks.
\newblock In {\em Proceedings of the European conference on computer vision (ECCV)}, pages 401--417, 2018.

\bibitem{christen2024diffh2o}
Sammy Christen, Shreyas Hampali, Fadime Sener, Edoardo Remelli, Tomas Hodan, Eric Sauser, Shugao Ma, and Bugra Tekin.
\newblock Diffh2o: Diffusion-based synthesis of hand-object interactions from textual descriptions.
\newblock In {\em SIGGRAPH Asia 2024 Conference Papers}, pages 1--11, 2024.

\bibitem{cha2024text2hoi}
Junuk Cha, Jihyeon Kim, Jae~Shin Yoon, and Seungryul Baek.
\newblock Text2hoi: Text-guided 3d motion generation for hand-object interaction.
\newblock In {\em Proceedings of the IEEE/CVF Conference on Computer Vision and Pattern Recognition}, pages 1577--1585, 2024.

\bibitem{ghosh2023imos}
Anindita Ghosh, Rishabh Dabral, Vladislav Golyanik, Christian Theobalt, and Philipp Slusallek.
\newblock Imos: Intent-driven full-body motion synthesis for human-object interactions.
\newblock In {\em Computer Graphics Forum}, volume~42, pages 1--12. Wiley Online Library, 2023.

\bibitem{brahmbhatt2019contactdb}
Samarth Brahmbhatt, Cusuh Ham, Charles~C Kemp, and James Hays.
\newblock Contactdb: Analyzing and predicting grasp contact via thermal imaging.
\newblock In {\em Proceedings of the IEEE/CVF conference on computer vision and pattern recognition}, pages 8709--8719, 2019.

\bibitem{ho2020denoising}
Jonathan Ho, Ajay Jain, and Pieter Abbeel.
\newblock Denoising diffusion probabilistic models.
\newblock {\em Advances in neural information processing systems}, 33:6840--6851, 2020.

\bibitem{huang2025hoigpt}
Mingzhen Huang, Fu-Jen Chu, Bugra Tekin, Kevin~J Liang, Haoyu Ma, Weiyao Wang, Xingyu Chen, Pierre Gleize, Hongfei Xue, Siwei Lyu, et~al.
\newblock Hoigpt: Learning long-sequence hand-object interaction with language models.
\newblock In {\em Proceedings of the Computer Vision and Pattern Recognition Conference}, pages 7136--7146, 2025.

\bibitem{zhou2025megohand}
Bohan Zhou, Yi~Zhan, Zhongbin Zhang, and Zongqing Lu.
\newblock Megohand: Multimodal egocentric hand-object interaction motion generation.
\newblock {\em arXiv preprint arXiv:2505.16602}, 2025.

\bibitem{intelligence2025pi_}
Physical Intelligence, Kevin Black, Noah Brown, James Darpinian, Karan Dhabalia, Danny Driess, Adnan Esmail, Michael Equi, Chelsea Finn, Niccolo Fusai, et~al.
\newblock $\pi_{0.5}$: a vision-language-action model with open-world generalization.
\newblock {\em arXiv preprint arXiv:2504.16054}, 2025.

\bibitem{pertsch2025fast}
Karl Pertsch, Kyle Stachowicz, Brian Ichter, Danny Driess, Suraj Nair, Quan Vuong, Oier Mees, Chelsea Finn, and Sergey Levine.
\newblock Fast: Efficient action tokenization for vision-language-action models.
\newblock {\em arXiv preprint arXiv:2501.09747}, 2025.

\bibitem{liu2024rdt}
Songming Liu, Lingxuan Wu, Bangguo Li, Hengkai Tan, Huayu Chen, Zhengyi Wang, Ke~Xu, Hang Su, and Jun Zhu.
\newblock Rdt-1b: a diffusion foundation model for bimanual manipulation.
\newblock {\em arXiv preprint arXiv:2410.07864}, 2024.

\bibitem{ye2025dex1b}
Jianglong Ye, Keyi Wang, Chengjing Yuan, Ruihan Yang, Yiquan Li, Jiyue Zhu, Yuzhe Qin, Xueyan Zou, and Xiaolong Wang.
\newblock Dex1b: Learning with 1b demonstrations for dexterous manipulation.
\newblock {\em arXiv preprint arXiv:2506.17198}, 2025.

\bibitem{deng2025graspvla}
Shengliang Deng, Mi~Yan, Songlin Wei, Haixin Ma, Yuxin Yang, Jiayi Chen, Zhiqi Zhang, Taoyu Yang, Xuheng Zhang, Heming Cui, et~al.
\newblock Graspvla: a grasping foundation model pre-trained on billion-scale synthetic action data.
\newblock {\em arXiv preprint arXiv:2505.03233}, 2025.

\bibitem{xu2025egodtm}
Boshen Xu, Yuting Mei, Xinbi Liu, Sipeng Zheng, and Qin Jin.
\newblock Egodtm: Towards 3d-aware egocentric video-language pretraining.
\newblock {\em arXiv preprint arXiv:2503.15470}, 2025.

\bibitem{gavryushin2025maple}
Alexey Gavryushin, Xi~Wang, Robert~JS Malate, Chenyu Yang, Xiangyi Jia, Shubh Goel, Davide Liconti, Ren{\'e} Zurbr{\"u}gg, Robert~K Katzschmann, and Marc Pollefeys.
\newblock Maple: Encoding dexterous robotic manipulation priors learned from egocentric videos.
\newblock {\em arXiv preprint arXiv:2504.06084}, 2025.

\bibitem{chen2025vidbot}
Hanzhi Chen, Boyang Sun, Anran Zhang, Marc Pollefeys, and Stefan Leutenegger.
\newblock Vidbot: Learning generalizable 3d actions from in-the-wild 2d human videos for zero-shot robotic manipulation.
\newblock In {\em Proceedings of the Computer Vision and Pattern Recognition Conference}, pages 27661--27672, 2025.

\bibitem{ma2025glover++}
Teli Ma, Jia Zheng, Zifan Wang, Ziyao Gao, Jiaming Zhou, and Junwei Liang.
\newblock Glover++: Unleashing the potential of affordance learning from human behaviors for robotic manipulation.
\newblock {\em arXiv preprint arXiv:2505.11865}, 2025.

\bibitem{lepert2025phantom}
Marion Lepert, Jiaying Fang, and Jeannette Bohg.
\newblock Phantom: Training robots without robots using only human videos.
\newblock {\em arXiv preprint arXiv:2503.00779}, 2025.

\bibitem{kareer2024egomimic}
Simar Kareer, Dhruv Patel, Ryan Punamiya, Pranay Mathur, Shuo Cheng, Chen Wang, Judy Hoffman, and Danfei Xu.
\newblock Egomimic: Scaling imitation learning via egocentric video.
\newblock {\em arXiv preprint arXiv:2410.24221}, 2024.

\bibitem{niu2025human2locoman}
Yaru Niu, Yunzhe Zhang, Mingyang Yu, Changyi Lin, Chenhao Li, Yikai Wang, Yuxiang Yang, Wenhao Yu, Tingnan Zhang, Bingqing Chen, et~al.
\newblock Human2locoman: Learning versatile quadrupedal manipulation with human pretraining.
\newblock {\em arXiv preprint arXiv:2506.16475}, 2025.

\bibitem{singh2024hand}
Himanshu~Gaurav Singh, Antonio Loquercio, Carmelo Sferrazza, Jane Wu, Haozhi Qi, Pieter Abbeel, and Jitendra Malik.
\newblock Hand-object interaction pretraining from videos.
\newblock {\em arXiv preprint arXiv:2409.08273}, 2024.

\bibitem{yuan2025cross}
Haoqi Yuan, Bohan Zhou, Yuhui Fu, and Zongqing Lu.
\newblock Cross-embodiment dexterous grasping with reinforcement learning.
\newblock In {\em The Thirteenth International Conference on Learning Representations}, 2025.

\bibitem{li2025maniptrans}
Kailin Li, Puhao Li, Tengyu Liu, Yuyang Li, and Siyuan Huang.
\newblock Maniptrans: Efficient dexterous bimanual manipulation transfer via residual learning.
\newblock In {\em Proceedings of the Computer Vision and Pattern Recognition Conference}, pages 6991--7003, 2025.

\bibitem{chen2024vividex}
Zerui Chen, Shizhe Chen, Etienne Arlaud, Ivan Laptev, and Cordelia Schmid.
\newblock Vividex: Learning vision-based dexterous manipulation from human videos.
\newblock {\em arXiv preprint arXiv:2404.15709}, 2024.

\bibitem{zhou2024learning}
Bohan Zhou, Haoqi Yuan, Yuhui Fu, and Zongqing Lu.
\newblock Learning diverse bimanual dexterous manipulation skills from human demonstrations.
\newblock {\em arXiv preprint arXiv:2410.02477}, 2024.

\bibitem{zhou2025you}
Huayi Zhou, Ruixiang Wang, Yunxin Tai, Yueci Deng, Guiliang Liu, and Kui Jia.
\newblock You only teach once: Learn one-shot bimanual robotic manipulation from video demonstrations.
\newblock {\em arXiv preprint arXiv:2501.14208}, 2025.

\bibitem{rajeswaran2017learning}
Aravind Rajeswaran, Vikash Kumar, Abhishek Gupta, Giulia Vezzani, John Schulman, Emanuel Todorov, and Sergey Levine.
\newblock Learning complex dexterous manipulation with deep reinforcement learning and demonstrations.
\newblock {\em arXiv preprint arXiv:1709.10087}, 2017.

\bibitem{wang2022dexgraspnet}
Ruicheng Wang, Jialiang Zhang, Jiayi Chen, Yinzhen Xu, Puhao Li, Tengyu Liu, and He~Wang.
\newblock Dexgraspnet: A large-scale robotic dexterous grasp dataset for general objects based on simulation.
\newblock {\em arXiv preprint arXiv:2210.02697}, 2022.

\bibitem{romero2017embodied}
Javier Romero, Dimitrios Tzionas, and Michael~J. Black.
\newblock Embodied hands: Modeling and capturing hands and bodies together.
\newblock {\em ACM Transactions on Graphics, (Proc. SIGGRAPH Asia)}, 36(6), November 2017.

\bibitem{chen2024expanding}
Zhe Chen, Weiyun Wang, Yue Cao, Yangzhou Liu, Zhangwei Gao, Erfei Cui, Jinguo Zhu, Shenglong Ye, Hao Tian, Zhaoyang Liu, et~al.
\newblock Expanding performance boundaries of open-source multimodal models with model, data, and test-time scaling.
\newblock {\em arXiv preprint arXiv:2412.05271}, 2024.

\bibitem{chen2024far}
Zhe Chen, Weiyun Wang, Hao Tian, Shenglong Ye, Zhangwei Gao, Erfei Cui, Wenwen Tong, Kongzhi Hu, Jiapeng Luo, Zheng Ma, et~al.
\newblock How far are we to gpt-4v? closing the gap to commercial multimodal models with open-source suites.
\newblock {\em Science China Information Sciences}, 67(12):220101, 2024.

\bibitem{bukschat2020efficientpose}
Yannick Bukschat and Marcus Vetter.
\newblock Efficientpose: An efficient, accurate and scalable end-to-end 6d multi object pose estimation approach.
\newblock {\em arXiv preprint arXiv:2011.04307}, 2020.

\bibitem{mahendran20173d}
Siddharth Mahendran, Haider Ali, and Ren{\'e} Vidal.
\newblock 3d pose regression using convolutional neural networks.
\newblock In {\em Proceedings of the IEEE international conference on computer vision workshops}, pages 2174--2182, 2017.

\bibitem{yuan2024robopoint}
Wentao Yuan, Jiafei Duan, Valts Blukis, Wilbert Pumacay, Ranjay Krishna, Adithyavairavan Murali, Arsalan Mousavian, and Dieter Fox.
\newblock Robopoint: A vision-language model for spatial affordance prediction for robotics.
\newblock {\em arXiv preprint arXiv:2406.10721}, 2024.

\bibitem{ma20243dsrbench}
Wufei Ma, Haoyu Chen, Guofeng Zhang, Yu-Cheng Chou, Celso~M de~Melo, and Alan Yuille.
\newblock 3dsrbench: A comprehensive 3d spatial reasoning benchmark.
\newblock {\em arXiv preprint arXiv:2412.07825}, 2024.

\bibitem{yuan2025seeing}
Yifu Yuan, Haiqin Cui, Yibin Chen, Zibin Dong, Fei Ni, Longxin Kou, Jinyi Liu, Pengyi Li, Yan Zheng, and Jianye Hao.
\newblock From seeing to doing: Bridging reasoning and decision for robotic manipulation.
\newblock {\em arXiv preprint arXiv:2505.08548}, 2025.

\bibitem{yang2025thinking}
Jihan Yang, Shusheng Yang, Anjali~W Gupta, Rilyn Han, Li~Fei-Fei, and Saining Xie.
\newblock Thinking in space: How multimodal large language models see, remember, and recall spaces.
\newblock In {\em Proceedings of the Computer Vision and Pattern Recognition Conference}, pages 10632--10643, 2025.

\bibitem{fan2023arctic}
Zicong Fan, Omid Taheri, Dimitrios Tzionas, Muhammed Kocabas, Manuel Kaufmann, Michael~J Black, and Otmar Hilliges.
\newblock Arctic: A dataset for dexterous bimanual hand-object manipulation.
\newblock In {\em Proceedings of the IEEE/CVF Conference on Computer Vision and Pattern Recognition}, pages 12943--12954, 2023.

\bibitem{garcia2018first}
Guillermo Garcia-Hernando, Shanxin Yuan, Seungryul Baek, and Tae-Kyun Kim.
\newblock First-person hand action benchmark with rgb-d videos and 3d hand pose annotations.
\newblock In {\em Proceedings of the IEEE conference on computer vision and pattern recognition}, pages 409--419, 2018.

\bibitem{kwon2021h2o}
Taein Kwon, Bugra Tekin, Jan St{\"u}hmer, Federica Bogo, and Marc Pollefeys.
\newblock H2o: Two hands manipulating objects for first person interaction recognition.
\newblock In {\em Proceedings of the IEEE/CVF international conference on computer vision}, pages 10138--10148, 2021.

\bibitem{hoque2025egodex}
Ryan Hoque, Peide Huang, David~J Yoon, Mouli Sivapurapu, and Jian Zhang.
\newblock Egodex: Learning dexterous manipulation from large-scale egocentric video.
\newblock {\em arXiv preprint arXiv:2505.11709}, 2025.

\bibitem{perrett2025hd}
Toby Perrett, Ahmad Darkhalil, Saptarshi Sinha, Omar Emara, Sam Pollard, Kranti~Kumar Parida, Kaiting Liu, Prajwal Gatti, Siddhant Bansal, Kevin Flanagan, et~al.
\newblock Hd-epic: A highly-detailed egocentric video dataset.
\newblock In {\em Proceedings of the Computer Vision and Pattern Recognition Conference}, pages 23901--23913, 2025.

\bibitem{cai2023smpler}
Zhongang Cai, Wanqi Yin, Ailing Zeng, Chen Wei, Qingping Sun, Wang Yanjun, Hui~En Pang, Haiyi Mei, Mingyuan Zhang, Lei Zhang, et~al.
\newblock Smpler-x: Scaling up expressive human pose and shape estimation.
\newblock {\em Advances in Neural Information Processing Systems}, 36:11454--11468, 2023.

\bibitem{zhao2025taste}
Hongxiang Zhao, Xingchen Liu, Mutian Xu, Yiming Hao, Weikai Chen, and Xiaoguang Han.
\newblock Taste-rob: Advancing video generation of task-oriented hand-object interaction for generalizable robotic manipulation.
\newblock In {\em Proceedings of the Computer Vision and Pattern Recognition Conference}, pages 27683--27693, 2025.

\bibitem{wang2023holoassist}
Xin Wang, Taein Kwon, Mahdi Rad, Bowen Pan, Ishani Chakraborty, Sean Andrist, Dan Bohus, Ashley Feniello, Bugra Tekin, Felipe~Vieira Frujeri, et~al.
\newblock Holoassist: an egocentric human interaction dataset for interactive ai assistants in the real world.
\newblock In {\em Proceedings of the IEEE/CVF International Conference on Computer Vision}, pages 20270--20281, 2023.

\bibitem{liu2024taco}
Yun Liu, Haolin Yang, Xu~Si, Ling Liu, Zipeng Li, Yuxiang Zhang, Yebin Liu, and Li~Yi.
\newblock Taco: Benchmarking generalizable bimanual tool-action-object understanding.
\newblock In {\em Proceedings of the IEEE/CVF Conference on Computer Vision and Pattern Recognition}, pages 21740--21751, 2024.

\bibitem{chao2021dexycb}
Yu-Wei Chao, Wei Yang, Yu~Xiang, Pavlo Molchanov, Ankur Handa, Jonathan Tremblay, Yashraj~S Narang, Karl Van~Wyk, Umar Iqbal, Stan Birchfield, et~al.
\newblock Dexycb: A benchmark for capturing hand grasping of objects.
\newblock In {\em Proceedings of the IEEE/CVF Conference on Computer Vision and Pattern Recognition}, pages 9044--9053, 2021.

\bibitem{petrovich23tmr}
Mathis Petrovich, Michael~J. Black, and G{\"u}l Varol.
\newblock {TMR}: Text-to-motion retrieval using contrastive {3D} human motion synthesis.
\newblock In {\em International Conference on Computer Vision ({ICCV})}, 2023.

\bibitem{wu2024gello}
Philipp Wu, Yide Shentu, Zhongke Yi, Xingyu Lin, and Pieter Abbeel.
\newblock Gello: A general, low-cost, and intuitive teleoperation framework for robot manipulators.
\newblock In {\em 2024 IEEE/RSJ International Conference on Intelligent Robots and Systems (IROS)}, pages 12156--12163. IEEE, 2024.

\bibitem{qin2023anyteleop}
Yuzhe Qin, Wei Yang, Binghao Huang, Karl Van~Wyk, Hao Su, Xiaolong Wang, Yu-Wei Chao, and Dieter Fox.
\newblock Anyteleop: A general vision-based dexterous robot arm-hand teleoperation system.
\newblock {\em arXiv preprint arXiv:2307.04577}, 2023.

\bibitem{yuan2025being}
Haoqi Yuan, Yu~Bai, Yuhui Fu, Bohan Zhou, Yicheng Feng, Xinrun Xu, Yi~Zhan, B{\"o}rje~F Karlsson, and Zongqing Lu.
\newblock Being-0: A humanoid robotic agent with vision-language models and modular skills.
\newblock {\em arXiv preprint arXiv:2503.12533}, 2025.

\end{thebibliography}





\clearpage

\end{document}